\documentclass[sn-mathphys-num,iicol]{sn-jnl}% Math and Physical Sciences Numbered Reference Style 
\usepackage{graphicx}%
\usepackage{multirow}%
\usepackage{amsmath,amssymb,amsfonts}%
\usepackage{amsthm}%
\usepackage{mathrsfs}%
\usepackage{amssymb}
\usepackage[title]{appendix}%
\usepackage{xcolor}%
\usepackage{textcomp}%
\usepackage{manyfoot}%
\usepackage{booktabs}%
\usepackage{algorithm}%
\usepackage{algorithmicx}%
\usepackage{algpseudocode}%
\usepackage{listings}%
\usepackage{natbib}
\usepackage{makecell}
\usepackage{threeparttable}
\usepackage{multicol}
\usepackage{longtable}
\usepackage{bbding}
\usepackage{svg}
\usepackage{colortbl} % 加载 colortbl 包
\usepackage{arydshln}
\usepackage{bm}
\usepackage{bbm}
\usepackage{diagbox}
\usepackage{textcomp}
\usepackage{stfloats}
\usepackage{url}
\usepackage{verbatim}
\usepackage{graphicx}
\usepackage{array}
\usepackage[small]{caption}
\usepackage[utf8]{inputenc}

\usepackage{orcidlink}
\usepackage{tikz}
\usepackage{forest}
\usetikzlibrary{shapes, arrows.meta, positioning}

\definecolor{root-purple}{HTML}{cfbcf0}
\definecolor{data-learning-green}{HTML}{f2fad0}
\definecolor{tabular-image-text-graph-blue}{HTML}{bdd7ee}
\definecolor{classification-generation-representation-red}{HTML}{f6d4d4}
\definecolor{white-box-grey}{HTML}{fcfcfc}
\definecolor{black-box-black}{HTML}{e0e0e0}

\definecolor{root-blue}{HTML}{deebf7}
\definecolor{model-robust-grey}{HTML}{e0e0e0}
\definecolor{additive-purification-detection-orange}{HTML}{fff2cc}
\definecolor{secure-robust-input-green}{HTML}{e2f0d9}
\definecolor{final-grey}{HTML}{fcfcfc}

\theoremstyle{thmstyleone}%
\theoremstyle{thmstyletwo}%

\theoremstyle{thmstylethree}%

\begin{document}

\title{Privacy Leakage on DNNs: A Survey of Model Inversion Attacks and Defenses}


\author[1]{\fnm{Hao} \sur{Fang} \orcidlink{0009-0004-0271-6579}}\email{ffhibnese@gmail.com}
\equalcont{These authors contributed equally to this work.}

\author[1]{\fnm{Yixiang} \sur{Qiu} \orcidlink{0009-0000-3444-5807}}\email{qiu-yx24@mails.tsinghua.edu.cn}
\equalcont{These authors contributed equally to this work.}

\author[1]{\fnm{Hongyao} \sur{Yu} \orcidlink{0009-0009-8525-1565}}\email{chrisqcwx@gmail.com}
\author[1]{\fnm{Wenbo} \sur{Yu} \orcidlink{0009-0004-8077-9487}}\email{wenbo.research@gmail.com}
\author[1]{\fnm{Jiawei} \sur{Kong} \orcidlink{0009-0001-4879-0668}}\email{kjw25@mails.tsinghua.edu.cn}
\author[2]{\fnm{Baoli} \sur{Chong} \orcidlink{0009-0001-2018-6653}}\email{chongbaoli@stu.hit.edu.cn}

\author*[2]{\fnm{Bin} \sur{Chen} \orcidlink{0000-0002-4798-230X}}\email{chenbin2021@hit.edu.cn}
\author[2]{\fnm{Xuan} \sur{Wang} \orcidlink{0000-0002-3512-0649}}\email{wangxuan@cs.hitsz.edu.cn}
\author[1,3]{\fnm{Shu-Tao} \sur{Xia} \orcidlink{0000-0002-8639-982X}}\email{xiast@sz.tsinghua.edu.cn}
\author[4]{\fnm{Ke} \sur{Xu} \orcidlink{0000-0003-2587-8517}}\email{xuke@tsinghua.edu.cn}

\affil[1]{Tsinghua Shenzhen International Graduate School, Tsinghua University, Shenzhen, Guangdong 518055, China}

\affil[2]{School of Computer Science and Technology, Harbin Institute of Technology, Shenzhen, Guangdong 518055, China}

\affil[3]{Peng Cheng Laboratory, Shenzhen, Guangdong 518055, China}

\affil[4]{Department of Computer Science and Technology, Tsinghua University, Beijing 100084, China}

% \affil[3]{\orgdiv{Department}, \orgname{Organization}, \orgaddress{\street{Street}, \city{City}, \postcode{610101}, \state{State}, \country{Country}}}


\abstract{ % Despite the great success Deep Neural Networks (DNNs) have achieved, their training data leakage has drawn growing concerns since it potentially carries users' privacy-sensitive information.
Deep Neural Networks (DNNs) have revolutionized various domains with their exceptional performance across numerous applications. However, Model Inversion (MI) attacks, which disclose private information about the training dataset by abusing access to the trained models, have emerged as a formidable privacy threat. Given a trained network, these attacks enable adversaries to reconstruct high-fidelity data that closely aligns with the private training samples, posing significant privacy concerns. Despite the rapid advances in the field, we lack a comprehensive and systematic overview of existing MI attacks and defenses. To fill this gap, this paper thoroughly investigates this realm and presents a holistic survey. Firstly, our work briefly reviews early MI studies on traditional machine learning scenarios.
We then elaborately analyze and compare numerous recent attacks and defenses on \textbf{D}eep \textbf{N}eural \textbf{N}etworks (DNNs) across multiple modalities and learning tasks. By meticulously analyzing their distinctive features, we summarize and classify these methods into different categories and provide a novel taxonomy.
% Moreover, our work includes recent training data leakage on Large Language Models (LLMs).
% Moreover, we investigate recent training data leakage on Large Language Models.
% and provide a novel taxonomy. 
% and categorize them as a novel paradigm of MI attacks.
Finally, this paper discusses promising research directions and presents potential solutions to open issues. 
To facilitate further study on MI attacks and defenses, we have implemented an open-source model inversion toolbox on \textcolor{blue}{\href{https://github.com/ffhibnese/Model-Inversion-Attack-ToolBox}{GitHub}}.}

\keywords{Data Leakage, Model Inversion Attack, Model Inversion Defense, Privacy Security}



\maketitle 

\section{Introduction}
\begin{figure}[tbp]
\centerline{\includegraphics[width=0.9\columnwidth]{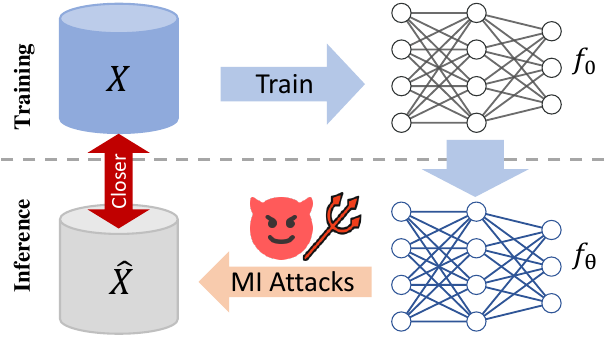}}
\caption{An illustration of MI attacks. The service provider trains the model $f_0(\cdot)$ with private dataset $X$ and obtains $f_{\theta}(\cdot)$ that is subsequently released for public usage. During the inference stage, the adversary inverts the available $f_{\theta}(\cdot)$ to acquire recovered data $\hat{X}$, which highly resembles the inaccessible $X$.}
\label{fig:mi_illstra}
% \vspace{-1em}
\end{figure}

% TODO: declare the difference between membership, property, and embedding inversion from our research topic. 
With the unprecedented development of Deep Learning (DL), Deep Neural Networks (DNNs) have been widely applied in a variety of fields, including medical research \cite{shen2017deep, zhou2023unified}, financial analysis \cite{shah2018comparative, ozbayoglu2020deep}, and personalized recommendations \cite{zhang2019deep}.
While many practical applications of DNNs require training on privacy-sensitive datasets, people may mistakenly expect that these private training samples are securely encoded in the weights of the trained network. However, numerous studies \cite{fredrikson2014privacy, ganju2018property, hu2022membership} have demonstrated that a malicious attacker could utilize the pre-trained models to reveal critical privacy information about the underlying training data. 
Among different types of attacks, Model Inversion (MI), as presented in Fig. \ref{fig:mi_illstra}, stands out as a potent threat that has raised increasing privacy concerns. 
Unlike the \textit{membership inference attacks} or the \textit{property inference attacks} which only reveal narrow aspects of privacy associated with training data \cite{jegorova2022survey}, MI allows an adversary to fully reconstruct the training samples by effectively exploiting the guidance from the given pre-trained models \cite{wang2021variational}.
% cite NIPS VMI

%TODO: make this paragraph shorter and briefly introduce the pipeline of model inversion algorithms.%
Model Inversion attack was first introduced by \cite{fredrikson2014privacy} on tabular data in the context of genomic privacy. They demonstrate that when partial input attributes are available to the attacker, the sensitive genomic markers can be accurately recovered by maximizing the posterior probability corresponding to a specific output of the given linear regression model.
% zhao2021exploiting
A following study \cite{fredrikson2015model} extends MI attacks to decision trees and shadow face recognition networks and reconstructs certain sensitive personal information based on gradient descent optimization. 
Subsequent studies \cite{song2017machine,yang2019neural} further broaden the scope of MI attacks to encompass various Machine Learning (ML) models and more forms of data using diverse optimizing strategies. Additionally, significant advancements have also been made in enhancing the attack efficacy while relaxing the assumptions about the attacker's capabilities and knowledge.
% To further relax the assumption, [] investigate a more general paradigm of MI attacks on different kinds of data and propose algorithms that do not require any knowledge about the target sample. 
Although these methods have exhibited outstanding performance in traditional ML scenarios for shallow networks and simple data, the effectiveness is significantly reduced when attacking deeper and wider DNNs trained on more complex and higher-dimensional data such as RGB images \cite{zhang2020secret}. 

%%%%%%%%%%%%%%%%%%%%%%%%%%%%%%%%%%%%%%%%%%%%%%%%%
% deepinverison
To address this challenge, recent studies \cite{zhang2020secret, struppek2022plug} have made great efforts and achieved remarkable performance improvements. \cite{zhang2020secret} first achieved RGB image recovery from deeper CNNs by exploiting Generative Adversarial Networks (GANs) \cite{goodfellow2016nips} as image priors. Specifically, they propose to train a GAN model with publicly available data to generate high-fidelity images for target classes. Benefiting from the rich prior knowledge encoded in the learned GAN model, the quality and fidelity of reconstructed images are significantly improved. Building upon this generative recovery paradigm, a variety of subsequent studies \cite{yuan2022secretgen, an2022mirror, liu2023diffusion} further propose diverse novel techniques and successfully obtain impressive reconstruction results. 
Moreover, recent explorations in model inversion on text data \cite{carlini2019secret} and graph data \cite{zhang2021graphmi} have shown that language models and graph neural networks are also vulnerable to MI threats. 
Particularly, the serious privacy breach of Large Language Models (e.g., ChatGPT \cite{brown2020language}) has drawn growing attention. Users can generate text-based queries and interact with ChatGPT, raising concerns regarding the inadvertent exposure of sensitive information through the model's responses \cite{nasr2023scalable}. This threat is greatly amplified in this explosion era of AIGC.
% Meanwhile,  that graph data used to train Graph Neural Networks (GNNs) is also vulnerable to MI threats \cite{zhang2021graphmi}. 

% Furthermore, the studies of MI are extended to  (NLP) tasks \cite{carlini2019secret}, especially for the privacy leakage of Large Language Models (LLMs), e.g., ChatGPT. Users can generate text-based queries and interact with ChatGPT, and concerns arise regarding the inadvertent exposure of sensitive information through the model's responses \cite{nasr2023scalable}. This threat is greatly amplified in this explosion era of AIGC.
% Meanwhile, researchers \cite{zhang2021graphmi} reveal that graph data used to train Graph Neural Networks (GNNs) is also vulnerable to MI threats \cite{zhang2021graphmi}. 

To protect the private training data from powerful MI attacks, a series of defense strategies \cite{wen2021defending, wang2021improving, peng2022bilateral} have been proposed to ensure the security of the publicly released models, which can be uniformly divided into \textit{Model Output Processing} and \textit{Robust Model Learning}. 
In light that MI attacks generally rely on the rich information within the model output, \textit{Model Output Processing} aims to mislead the attacker by diminishing the valid information in model output \cite{yang2020defending}. Meanwhile, \textit{Robust Model learning} prefers to incorporate well-designed mechanisms into the model training or fine-tuning process to essentially improve the victim's MI robustness \cite{wang2021improving}.

Despite the privacy threat to training data raised by MI attacks, a comprehensive overview summarizing their advances is currently not available. Recent related research \cite{dibbo2023sok} predominantly focuses on the diverse taxonomies and challenges of MI attacks solely based on some early studies only for the tabular and image data, \cite{song2022survey} analyzes the Python code implementation of several attacks, or \cite{jegorova2022survey} briefly introduces several representative MI attacks in a subsection among various types of privacy inference attacks. In contrast, we present a systematic and elaborate review of the advanced MI attacks and defenses over multiple data modalities and different learning tasks, thoroughly exploring the landscape.

The rest of the paper is organized as follows. Sec. \ref{sec:mi_overview} formally defines model inversion attacks and presents the overall profile of MI attacks with diverse taxonomies. Sec. \ref{sec:mi_ml} briefly reviews early studies on traditional ML scenarios while Sec. \ref{sec:mi_main} details attacks on DNNs along with a general framework of methods on classification tasks, based on which we systematically introduce the differences and relationships of these methods. Sec. \ref{sec:mi_more_modal} includes MI attacks on more modalities. 
Sec. \ref{sec:defense} then introduces and categorizes numerous defense strategies against MI attacks. Sec. \ref{sec:social} discusses the social impact. Finally, Sec. \ref{sec:fur_con} concludes this work and summarizes some remaining challenges and possible future directions.

\textbf{Survey Methodology.} We prioritize papers published in top-tier venues related to deep learning and model inversion, including \{NeurIPS, ICML, ICLR, CVPR, ICCV, ECCV, ACL, EMNLP, NAACL, S\&P, USENIX security, CCS, NDSS, ACM KDD, AAAI, ACM MM, IJCAI, TPAMI, IJCV, TIFS, TDSC\}. 
For preprints, we consider only those that introduce novel technical designs and demonstrate performance improvements over the state-of-the-art model inversion attacks or defenses available at the time, or extend model inversion to novel tasks or data modalities. Citation count is also taken into consideration, with priority given to preprints that have received more than 20 citations.
To ensure comprehensive coverage, we collect candidate papers from multiple sources, \textit{i.e.}, IEEE Xplore, ACM Digital Library, SpringerLink, ACL Anthology, OpenReview, Google Scholar, arXiv, and the official conference proceedings of the venues listed above.
To retrieve relevant literature, we adopt a set of related keywords, including \{model inversion, inversion attack, inversion defense, privacy leakage, privacy protection, data reconstruction, data extraction, data leakage, generative inversion attack, federated inversion, GAN model inversion, diffusion model inversion, text inversion, LLM inversion, language model inversion, graph inversion\}. 
% The retrieved papers are then further filtered through manual inspection of titles, abstracts, and research focus to ensure relevance.
We independently screen the titles and abstracts to exclude papers unrelated to model inversion. The key criterion is whether the primary research topic satisfies the definition of model inversion \cite{fredrikson2014privacy}, as introduced in Section~\ref{subsec:mi_defination}. Full texts were further examined when necessary to determine whether the paper proposes or analyzes model inversion attacks or defenses. This survey covers works starting from 2014, when model inversion was first explicitly formulated, through April 30, 2026.
% This survey covers works starting from 2015, when model inversion was first explicitly formulated, through 2026. 

Overall, we provide a comprehensive survey of the field, covering more than 130 papers related to model inversion attacks and defenses.

\section{Model Inversion Overview}
In this section, we first formally formulate the definition of Model Inversion attacks. Next, we present an overview of these methods from different perspectives based on the taxonomies as summarized in Figure \ref{fig:attack_taxonomy}.
\label{sec:mi_overview}
\subsection{Definition of Model Inversion}
\label{subsec:mi_defination}
% 这里说参考2014年fredrickson的文章给出定义。补充说道在我们的顶一下，攻击者可以用任何手段来挖掘训练好的模型的信息，以重建隐私数据。
% 往往利用模型输出、中间特征信息或者挖掘输入和输出的关系来发动攻击。
The process of transforming training data into machine-learned systems is not one-way, but two \cite{veale2018algorithms}: the learned models can also be inverted to obtain the training data. Canonical model inversion attacks \cite{zhang2020secret, wang2021variational, zhang2021graphmi, parikh2022canary} can be defined as follows: given a pre-trained model $f_{\theta}(\cdot)$ and certain prior knowledge $\mathcal{D}$ (which may be none), the attacker aims to develop an effective algorithm $\mathcal{T}$ that inverts the network $f_{\theta}$ to recover as much as possible of the private information about the training dataset $X$. The inversion algorithm $\mathcal{T}$ generally capitalizes on the confidences or feature representations output by the model to achieve satisfactory reconstruction results. A successful model inversion attack should generate realistic reconstructed results $\hat{X} = \mathcal{T}(f_{\theta}, \mathcal{D})$ that closely resemble the target dataset $X$ without having access to it. 
% Different from the \textit{membership inference attack} or the \textit{property inference attack} that only reveal partial information about the training data \cite{jegorova2022survey}, MI enables an adversary to fully reconstruct private training samples, which has raised growing concern.
% by inverting the given pre-trained models \cite{wang2021variational}.

%%%%%%%%%%%%%%%%%%%%%%%%%%%%%%%%%%%%%%%%%%%%%%%%%
% BY FH, YWB 
% \begin{figure*}[tbp]
% \centerline{\includegraphics[width=\linewidth]{figure/cv_result/Attacks_taxonomy_v7.pdf}}
% \caption{Taxonomy of model inversion attacks across various data modalities and learning tasks.}
% \label{fig:attack_taxonomy}
% \end{figure*}

\begin{figure*}[tbp]
    \centering
    \resizebox{\textwidth}{!}{
        \begin{forest}
            for tree={
                grow'=east,  % 使用标准的 grow
                reversed=true,
                anchor=center,
                parent anchor=east,
                child anchor=west,
                font=\large,
                rectangle,
                draw=black,
                rounded corners,
                align=center,
                minimum width=4em,
                edge={darkgray, line width=1.5pt},
                edge path={
                    \noexpand\path[\forestoption{edge}]
                    (!u.parent anchor) -- ++(5pt,0) |- (.child anchor)\forestoption{edge label};
                },
                s sep=3pt,
                inner xsep=2pt,
                inner ysep=3pt,
                line width=0.8pt,
            },
            where level=0{
                text width=9em,
                rotate=90,           % 文字逆时针旋转90度（竖排）
                anchor=center,       % 节点中心作为锚点
                parent anchor=south,% 父节点中心对齐
                child anchor=north, % 子节点中心对齐
                align=center,        % 文字居中
                minimum height=3em,  % 保证节点高度足够容纳旋转后的文字
            }{},
            where level=1{
                text width=9em,
                rotate=90,           % 文字逆时针旋转90度（竖排）
                anchor=center,       % 节点中心作为锚点
                parent anchor=south,% 父节点中心对齐
                child anchor=north, % 子节点中心对齐
                align=center,        % 文字居中
                minimum height=3em,  % 保证节点高度足够容纳旋转后的文字
            }{},
            where level=2{
                text width=8em,
                parent anchor=east,
                child anchor=west,
            }{},
            where level=3{
                text width=6em,
                parent anchor=east,
                child anchor=west,
            }{},
            where level=4{
                text width=41em,
                align=left,
                anchor=west,
                inner xsep=8pt,
            }{},
            [\ Model Inversion, fill=root-purple
                [\  Data Modalities, fill=data-learning-green
                    [\,\,\,Tabular data, fill=tabular-image-text-graph-blue
                        [\,White-box, fill=white-box-grey
                            [{TreeMIA \cite{fredrikson2015model}}, fill=white-box-grey, edge={white, line width=0pt}]
                        ]
                        [\,Black-box, fill=black-box-black
                            [{Genotype-MIA \cite{fredrikson2014privacy}, TreeMIA \cite{fredrikson2015model}, PPL-Search \cite{carlini2019secret},General-MIA \cite{hidano2018model}}, fill=black-box-black, edge={white, line width=0pt}]
                        ]
                    ]
                    [\,\,\,\,\,Image data, fill=tabular-image-text-graph-blue
                        [\,White-box, fill=white-box-grey
                            [{VMI \cite{wang2021variational}, TreeMIA \cite{fredrikson2015model}, ParaLeak \cite{song2017machine}, GMI \cite{zhang2020secret}, PPA \cite{struppek2022plug}, SecretGen \cite{yuan2022secretgen}, \\ Mirror \cite{an2022mirror}, DiffMI \cite{carlini2023extracting}, PLGMI \cite{yuan2023pseudo}, GAN-Attack \cite{hitaj2017deep}, LOMMA \cite{nguyen2023re}, \\ IF-GMI \cite{qiu2024closer}, KEDMI \cite{chen2021knowledge}, AlphaMI \cite{alphaganmi}, DMMIA \cite{qi2023model}}, fill=white-box-grey, edge={white, line width=0pt}]
                        ]
                        [\,Black-box, fill=black-box-black
                            [{TreeMIA \cite{fredrikson2015model}, ParaLeak \cite{song2017machine}, KL-MIA \cite{yang2019neural}, SecretGen \cite{yuan2022secretgen}, Mirror \cite{an2022mirror}, \\ DiffusionMI \cite{liu2023diffusion}, DiffMI \cite{carlini2023extracting}, BREPMI \cite{kahla2022label}, SparseMI \cite{xu2023sparse}, C2FMI \cite{ye2023c2fmi}, \\
                            XAI-MIA \cite{zhao2021exploiting}, LOKT \cite{nguyen2023labelonly}, SimBA \cite{zhou2023boosting}, Amplified-MIA \cite{zhang2023analysis},  DAMIA \cite{zhang2024aligning}, \\ RLBMI \cite{han2023reinforcement}, P2IMI \cite{liu2024prediction},  }, fill=black-box-black, edge={white, line width=0pt}]
                        ]
                    ]
                    [\,\,\,\,\,\,Text data, fill=tabular-image-text-graph-blue
                        [\,White-box, fill=white-box-grey
                            [{ParaLeak \cite{song2017machine}, CanaryMI \cite{ parikh2022canary},  Ethicist \cite{zhang2023ethicist}, ShadowKI \cite{pan2020privacy}, LayerMap \cite{song2020information}, \\ Mix And Match \cite{elmahdy2024deconstructing}}, fill=white-box-grey, edge={white, line width=0pt}]
                        ]
                        [\,Black-box, fill=black-box-black
                            [{CapLeak \cite{song2017machine}, PPL-Search \cite{carlini2019secret}, DivAttack \cite{nasr2023scalable},  DANN \cite{pan2020privacy},  GEIA \cite{li2023sentence}, \\ LogitInv \cite{morris2023language}, GenRank \cite{carlini2021extracting}, ConfSearch \cite{elmahdy2022privacy}, CtxLeak  \cite{huang2022large}, \\ Output2Prompt \cite{zhang2024extracting}, GenRank+ \cite{yu2023bag}, DiffRank \cite{zanella2020analyzing}, DRAttack \cite{panchendrarajan2021dataset}, \\ Multilingual-EI \cite{chen2024text}, Trans-EI \cite{huang2024transferable}, Ctx-PR \cite{zhangeffective}, PILS \cite{nazir2026better}}, fill=black-box-black, edge={white, line width=0pt}]
                        ]
                    ]
                    [\,\,\,\,\,Graph data, fill=tabular-image-text-graph-blue
                        [\,White-box, fill=white-box-grey
                            [{GraphMI \cite{zhang2021graphmi}, MC-GRA \cite{zhou2023on}, RL-GraphMI \cite{zhang2023model}}, fill=white-box-grey, edge={white, line width=0pt}]
                        ]
                        [\,Black-box, fill=black-box-black
                            [{RL-GraphMI \cite{zhang2023model}, GraphLeaks \cite{duddu2020quantifying}, GNN-Leaks \cite{zhang2022inference}, Link-Stealing \cite{he2021stealing}, \\ LINKTELLER \cite{wu2022LINKTELLER}}, fill=black-box-black, edge={white, line width=0pt}]
                        ]
                    ]
                ]
                [\    Learning Tasks, fill=data-learning-green
                    [\,\,\,Classification, fill=classification-generation-representation-red
                        [\,White-box, fill=white-box-grey
                            [{VMI \cite{wang2021variational}, TreeMIA \cite{fredrikson2015model}, ParaLeak \cite{song2017machine}, SecretGen \cite{yuan2022secretgen}, Mirror \cite{an2022mirror}, \\ GraphMI \cite{zhang2021graphmi}, CanaryMI \cite{parikh2022canary}, PLGMI \cite{yuan2023pseudo}, GAN-Attack \cite{hitaj2017deep}  LOMMA \cite{nguyen2023re}, \\ IF-GMI \cite{qiu2024closer}, KEDMI \cite{chen2021knowledge}, AlphaMI \cite{alphaganmi}, DMMIA \cite{qi2023model}, 
                            Mix And Match \cite{elmahdy2024deconstructing}, \\ MC-GRA \cite{zhou2023on}, RL-GraphMI \cite{zhang2023model}}, fill=white-box-grey, edge={white, line width=0pt}]
                        ]
                        [\,Black-box, fill=black-box-black
                            [{Genotype-MIA \cite{fredrikson2014privacy}, TreeMIA \cite{fredrikson2015model}, ParaLeak \cite{song2017machine}, KL-MIA \cite{yang2019neural}, SecretGen \cite{yuan2022secretgen},\\ Mirror \cite{an2022mirror}, DiffusionMI \cite{liu2023diffusion}, BREPMI \cite{kahla2022label}, SparseMI \cite{xu2023sparse}, C2FMI \cite{ye2023c2fmi},\\ XAI-MIA \cite{zhao2021exploiting}, LOKT \cite{nguyen2023labelonly}, SimBA \cite{zhou2023boosting}, Amplified-MIA \cite{zhang2023analysis}, DAMIA \cite{zhang2024aligning}, \\ RLBMI \cite{han2023reinforcement}  ConfSearch \cite{elmahdy2022privacy}, RL-GraphMI \cite{zhang2023model}, GraphLeaks \cite{duddu2020quantifying}, \\ GNN-Leaks \cite{zhang2022inference}, Link-Stealing \cite{he2021stealing}, LINKTELLER \cite{wu2022LINKTELLER}}, fill=black-box-black, edge={white, line width=0pt}]
                        ]
                    ]
                    [\,\,\,\,\,Generation, fill=classification-generation-representation-red
                        [\,White-box, fill=white-box-grey
                            [{DiffusionMI \cite{liu2023diffusion}, Ethicist \cite{zhang2023ethicist}}, fill=white-box-grey, edge={white, line width=0pt}]
                        ]
                        [\,Black-box, fill=black-box-black
                            [{DivAttack \cite{nasr2023scalable}, LogitInv \cite{morris2023language}, GenRank \cite{carlini2021extracting}, CtxLeak \cite{huang2022large}, Output2Prompt \cite{zhang2024extracting}, \\ GenRank+ \cite{yu2023bag},  DRAttack \cite{panchendrarajan2021dataset}, Ctx-PR \cite{zhangeffective}, PILS \cite{nazir2026better}}, fill=black-box-black, edge={white, line width=0pt}]
                        ]
                    ]
                    [Representation \\ Learning, fill=classification-generation-representation-red
                        [\,White-box, fill=white-box-grey
                            [{ShadowKI \cite{pan2020privacy}, LayerMap \cite{song2020information}}, fill=white-box-grey, edge={white, line width=0pt}]
                        ]
                        [\,Black-box, fill=black-box-black
                            [{PPL-Search \cite{carlini2019secret}, DANN \cite{pan2020privacy}, GEIA \cite{li2023sentence}, DiffRank \cite{zanella2020analyzing}, Multilingua-EI \cite{chen2024text}, \\ Trans-EI \cite{huang2024transferable}}, fill=black-box-black, edge={white, line width=0pt}]
                        ]
                    ]
                ]
            ]
        \end{forest}
    }
    \caption{Taxonomy of model inversion attacks across various data modalities and learning tasks.}
    \label{fig:attack_taxonomy}
\end{figure*}

% \input{table/summary}
%%%%%%%%%%%%%%%%%%%%%%%%%%%%%%%%%%%%%%%%%%%%%%%%%

\subsection{Attacker's Capability}
% To better analyze and compare different reconstruction approaches, we generally categorize these methods of three modalities 
Based on the attacker's capability and knowledge, existing MI methods can be uniformly categorized into \textbf{\emph{white-box}} and \textbf{\emph{black-box}} attacks. Specifically, a white-box scenario implies that the attacker has full access to the weights and outputs of the target model. Benefiting from the available gradient information of the victim model, white-box attacks \cite{zhang2020secret, struppek2022plug} have demonstrated excellent recovery efficacy through advanced gradient-based optimization strategies. Conversely, black-box settings only allow access to the predicted confidence probabilities or hard labels, i.e., an attacker can solely control the input query and obtain the model output. To handle this more challenging situation, researchers employ gradient estimation \cite{kahla2022label, xu2023sparse} or heuristic algorithms \cite{an2022mirror, ye2023c2fmi} to navigate the data recovery.

\subsection{Reconstructed Data Modalities}
% \todo{explain that early studies focus on simple data and networks while recent attacks aim at higher resolution and more complicated images.}
Given that the attack paradigms are largely contingent upon the target data modality, we present a taxonomy of MI attacks based on data modality to provide a concise overview of these attacks. 

\textit{Tabular Data Recovery (Sec. \ref{sec:mi_ml}).} Pioneer MI studies \cite{fredrikson2014privacy, fredrikson2015model} investigate attacks on tabular data that contains sensitive personal information (e.g., addresses, telephones, and medical records) in traditional ML scenarios. An adversary maliciously accesses an ML model and infers sensitive information by picking feature values for each individual from all possible feature combinations to minimize the expected error rates calculated by the individual's label and the model output. 

\begin{table*}[t]
\centering
\caption{Comparison of our survey with existing surveys on model inversion.}
\label{tab:survey_comparison}
\resizebox{\textwidth}{!}{
\begin{tabular}{lccccc}
\toprule
\textbf{Survey} &
\textbf{Attack \& Defense Taxonomy} &
\textbf{Data Modalities} &
\textbf{Reproduced Results} &
\textbf{Toolkit for MI} &
\textbf{Extra Features} \\
\midrule

Dibbo et al.~\cite{dibbo2023sok}
& Attack only
& Image/Text
& --
& --
& Taxonomy and open challenges \\

Song et al.~\cite{song2022survey}
& Attack only
& Image
& --
& --
& Python Implementation analysis \\

Jegorova et al.~\cite{jegorova2022survey}
& --
& Image
& --
& --
& -- \\

Zhou et al.~\cite{zhou2024model}
& \checkmark
& Image/Text/Graph
& --
& --
& \makecell{Thorough summary of evaluation \\ metrics and datasets} \\

\textbf{Ours}
& \checkmark (fine-grained)
& Image/Text/Graph
& \checkmark
& \checkmark
& \makecell{Rich mathematical formulations\\
and original illustrative figures} \\

\bottomrule
\end{tabular}
}
\end{table*}

\textit{Image Data Recovery (Sec. \ref{sec:mi_ml} and \ref{sec:mi_main}).} Sensitive image data like human faces or medical diagnostic images are also in the scope of MI attacks. Compared with tabular data, an image contains an intractable large set of features to reconstruct, which makes the feature-picking strategy impractical. To address this, \cite{fredrikson2015model} proposes an approach against face classifiers that formulates the model inversion as an optimization problem. Subsequent studies \cite{zhang2020secret, wang2021variational, yuan2023pseudo} generally follow this basic paradigm and further employ numerous techniques (e.g., generative models) to promote the reconstruction of higher-dimensional and higher-resolution images. Another roadmap \cite{yang2019neural,liu2024prediction} is to construct a set of input-output pairs by repeatedly querying the victim model and train another inversion network to map the output predictions back to the input images. Besides, \cite{carlini2023extracting, chen2024extracting} consider directly inducing the target diffusion models to generate their private training data.

\textit{Text Data Recovery (Sec. \ref{sec:mi_text}).} Examples of text data leakage are analogous to tabular ones. 
One difference is that in certain scenarios, the user input \cite{li2023sentence} or prompts preceding the input sentences \cite{morris2023language} are also pivotal privacy components targeted by MI attacks.
Moreover, the distinctive nature of text data and language models motivates rich forms of attack algorithms, which can be broadly categorized into \textit{Embedding Optimization} \cite{parikh2022canary, zhang2023ethicist}, \textit{Inverse Mapping} \cite{pan2020privacy, song2020information}, \textit{Token search} \cite{carlini2021extracting, elmahdy2022privacy}, and \textit{Malicious Prompt Design} \cite{huang2022large, nasr2023scalable}.

\textit{Graph Data Recovery (Sec. \ref{sec:mi_graph}).} Graphs containing rich personal information and relational data (e.g., friendship or intimacy) between users are also vulnerable to MI attacks. Researchers delve into discrete structures of graph data and propose powerful techniques to achieve successful reconstruction. We further categorize them into: \emph{Adjacency Optimization} \cite{zhang2021graphmi, zhou2023on}, \emph{Inverse Mapping} \cite{duddu2020quantifying, zhang2022inference} and \emph{Relationship Reasoning} \cite{he2021stealing, wu2022LINKTELLER}.

\subsection{Learning Tasks}
Given that these attacks are also task-specific, we undertake a concise review of these methods from the perspective of learning tasks. Figure \ref{fig:attack_taxonomy} demonstrates that most MI studies focus on classification tasks, which typically involve minimizing the classification loss specific to the target label to update the data using diverse optimization algorithms. For generation tasks, attackers usually conduct a heuristic search for the private data based on a pre-defined metric \cite{carlini2021extracting, elmahdy2022privacy} or directly induce the generative models to output their training data \cite{carlini2023extracting, chen2024extracting}. 
Attacks against representation learning \cite{carlini2021extracting, li2023sentence} tasks generally target text data, and their algorithms are highly determined by the specific pre-training tasks of the victim model. %In addition, \cite{fredrikson2014privacy, hidano2017model} extra considers the regression task and recovers sensitive tabular data by selecting the most probable values for a linear regression model.

\textit{Relationship with Data Modality.} The victim models trained on image and graph data are primarily used for classification tasks, whereas MI attacks on text data generally target language models for generation or representation learning. Correspondingly, the varying learning tasks of victim models result in disparate attack strategies.

\subsection{Comparison with Existing Related Surveys}
The vulnerabilities of DNNs have attracted increasing research attention \cite{fang2023gifd, fang2025one, yu2025gi, qiu2024closer}. \textit{E.g.}, \cite{le2024comprehensive} discusses backdoor attacks and defenses in face recognition systems. In this work, we focus on model inversion, a fundamental and critical privacy threat in machine learning systems.

Several recent surveys have explored this topic from different perspectives. A comparison between our survey and existing surveys across several key aspects is provided in Table~\ref{tab:survey_comparison}. \cite{dibbo2023sok} focuses on early MI studies largely restricted to tabular and image data, and emphasizes taxonomies and open challenges. \cite{song2022survey} analyzes Python-based implementations of several attack methods. Meanwhile, \cite{jegorova2022survey} briefly discusses MI as part of a broader overview of privacy inference attacks.
In comparison, this work provides a more comprehensive and in-depth treatment of both advanced MI attack and defense across diverse data modalities and learning tasks. 
% We systematically analyze techniques and present a holistic view of the field.

A concurrent study \cite{zhou2024model} also presents an MI survey across multiple domains.
% A follow-up study \cite{zhou2024model} also presents an MI survey across multiple domains.
However, our work makes several distinct contributions. First, we present a more fine-grained taxonomy of attack methods by organizing key methodological dimensions in Table \ref{table:cv_attack}, enabling more systematic comparisons across attacks. We also propose a dedicated taxonomy of defense strategies in Figure \ref{fig:defense_taxonomy} for understanding defense design principles.

In contrast to a largely enumerative presentation, we provide a more integrated analysis of existing methods with explicit comparisons, better clarifying their relationships and distinctions. Our work is further supported by rich mathematical formulations and well-designed original illustrative figures. 
Finally, we contribute an open-source toolkit for MI attacks, which has been widely adopted by the community. Using this toolkit, we reproduce representative attacks and present visualized results in Figure \ref{fig:cv_reconstruct}, providing intuitive comparisons of their effectiveness.

\section{MI Attacks on ML Scenarios}
\label{sec:mi_ml}
% \todo{(FFH) Introduce early attacks on traditional ML attacks. Yixiang and Hongyao, try to divide these attacks into different categories. And remember to write the formulas of early image MI attacks,} 
This section provides a brief review of early MI attacks and defenses on traditional ML models with simple structured data. 
% \subsection{Previous Exploration Stage}

The pioneering study by Fredrickson et al. \cite{fredrikson2014privacy} proposes the first model inversion attack against tabular data in a case study of personalized warfarin dosing \cite{kamali2010pharmacogenetics}. They simply infer the target attribute $x_{t}^{*}$ ($i.e.$, genotype) from a linear regression model via \textit{maximum a posterior probability} (MAP) estimate as follows:

\begin{equation}
\begin{aligned}
    x_{t}^{*} = \mathop{\arg\max}\limits_{x_{t}}\Sigma_{\mathbf{x}\in\mathbf{\hat{X}}:\mathbf{x}_{t}=x_{t}}\Pi_{1\leq i \leq m} p_{i}(\mathbf{x}_{i}),
     % x_{t}^{*} = \mathop{\arg\max}\limits_{x_{t}}\sum_{\mathbf{x}\in\mathbf{\hat{X}}:\mathbf{x}_{t}=x_{t}}\prod_{1\leq i \leq m} p_{i}(\mathbf{x}_{i}),
\end{aligned}
\label{eq:mape}
\end{equation}

\noindent where $\mathbf{x}_{i}$ is the $i$-th attribute of the sample $\mathbf{x}$, $p_i(\cdot)$ denotes the marginal distribution of $\mathbf{x}_{i}$, and $m$ is the number of attributes. 
Given the access to the $k$ insensitive attributes $\mathbf{z}=(x_1, \dots, x_k)$ and target label $y$, the candidates set $\mathbf{\hat{X}}$ includes samples that match these non-private attributes and are predicted into $y$, i.e., $\mathbf{\hat{X}}=\{\mathbf{\hat{x}}: \mathbf{\hat{x}_{K}}=\mathbf{z}, f(\mathbf{\hat{x}})=y\}$, where $\mathbf{K}$ denotes the corresponding indices. To mitigate the stated privacy threat, \cite{wang2015regression} proposes the first MI defense based on differential privacy (DP) \cite{zhang2012functional}. 
They leverage an adaptive strategy by adding more noise for sensitive attributes and less noise for non-sensitive attributes to improve MI robustness and retain the utility of target models. 
Subsequently, more in-depth MI attack methods emerge. 
\cite{fredrikson2015model} explores attacks on decision trees and facial recognition models. They develop novel estimators for decision trees trained on tabular data while applying gradient descent directly on pixels of grayscale images when attacking shallow facial recognition models. 
Besides, \cite{fredrikson2015model} presents the first definition of \textit{black-box} and \textit{white-box} MI attacks according to the knowledge about the target model.
To build a universal MI framework, \cite{wu2016methodology} summarizes previous methods and proposes a more rigorous formulation of MI attacks. Considering attacks in the Federated Learning (FL) system, \cite{hitaj2017deep} introduces a simple GAN model to enhance the reconstruction. Regarding the shared global model in FL as the discriminator, the attacker trains a generator to learn the same distribution as the target category. Once the training finishes, the generator can directly produce samples similar to private training data. Apart from image recovery, MI attacks were first extended to text modality by \cite{song2017machine}, which explores attacks from the lens of a service provider for programming code. In the white-box scenario, they manipulate the training codes to encode the user's private information into model parameters using a proposed encoding strategy. After the model is released, the privacy can be extracted by carefully analyzing the model parameters. 
For black-box attacks, model parameters become inaccessible. Hence, they synthesize extra malicious training data whose labels carry the encoded private information. The attacker queries the released model with these synthetic data and then analyzes the returned labels to infer the hidden private information.
To relax the assumption about access to some non-sensitive attributes \cite{fredrikson2014privacy}, \cite{hidano2018model} designs a general black-box MI framework against the linear regression model that does not require any knowledge about target samples.
% against linear regression models.
 % without any knowledge about the target sample
By considering the data poisoning mechanism, they divide the attack into three stages: \textit{Setup}, \textit{Poisoning}, and \textit{Model Inversion}. 
In the initial \textit{Setup} stage, the attacker estimates the regression coefficients and necessary hyperparameters of the victim model. During the \textit{Poisoning}, the estimated parameters are utilized to generate poisoned training data that aims to reduce the correlation between non-sensitive attributes and model parameters.
Finally, the target model becomes highly correlated with the sensitive attributes and leaks considerable private information in the \textit{Model Inversion} stage.
Recently, \cite{dibbo2025improving} has explored a novel defense strategy against MI attacks on traditional linear models. Specifically, they introduce sparse coding layers into the classifier, which compute the sparse representation of their inputs by Locally Competitive Algorithm (LCA) \cite{rozell2008sparse}. With fewer active neurons, the sparse coding layers can jettison unnecessary private information while maintaining the utility of the classifier, leading to more robustness against MI attacks.

\textit{Inversion Model-Based Image Recovery.}
% \todo{(FFH) Briefly introduce early MI attacks on DNNs but with gray images. e.g., encoder/decoder-based attacks. And tell that more powerful attacks that can handle RGB images are in the following subsection.}
This category of MI attack constitutes a series of black-box methods against relatively shallow networks mostly trained on grayscale images. 
With the assumption that the predicted vectors of private images are known to the attacker, these approaches treat the target model as a fixed encoder and construct an inversion model as the corresponding decoder, which maps these private confidence vectors of the target model back into the reconstructed images. 
Formally, the inversion model $\mathcal{M}(\cdot)$ is trained by minimizing:

\begin{equation}
\begin{aligned}
    \mathcal{M}^{*} = \mathop{\arg\min}\limits_{\mathcal{M}}\mathbb{E}_{x\sim X_{pub}}\mathcal{L}(\mathcal{M}(f_{\theta}(x)),x),
\end{aligned}
\label{eq:inversion}
\end{equation}

\noindent where $\mathcal{L}(\cdot, \cdot)$ is the distance loss for reconstructed images, $X_{pub}$ represents the auxiliary public dataset and $f_{\theta}(\cdot)$ denotes the target model. Once the inversion model is trained, the adversary can reconstruct the target private images with the given corresponding prediction vectors.
\cite{yang2019neural} first proposes this inversion paradigm in scenarios where only corrupted predictions of private images are available (i.e., only partial confidence scores are accessible). Specifically, the attacker collects public data as an auxiliary training dataset for the inversion model. To simulate the corruption in the confidence vectors, the adversary truncates the predictions of public images with top-$k$ selection, which is then fed into the inversion model $\mathcal{M}(\cdot)$ to perform optimization via Eq. (\ref{eq:inversion}). 
When the training is completed, the target images are reconstructed directly from the given prediction through the inversion model. \cite{zhao2021exploiting} explores the privacy risk in Explainable Artificial Intelligence (XAI) and presents a novel inversion model structure to enhance MI attacks. To better leverage the explanations (e.g., Gradients \cite{simonyan2013deep} and LRP \cite{bach2015pixel}) from the target model, they design a multi-modal inversion model to receive the processed explanations as a second input.
Besides, an additional network (e.g., Flatten and U-Net \cite{ronneberger2015u}) is introduced in front of the inversion model to preprocess the explanations. With the rich information contained in the explanations, they first conquer the challenge of reconstructing RGB images in the context of XAI learning scenarios. 
To improve the inversion model training, \cite{zhou2023boosting} generates adversarial samples with the untargeted attack SimBA \cite{guo2019simple} as data augmentation to expand the auxiliary training set. In addition to the reconstruction loss in Eq. (\ref{eq:inversion}), a novel semantic loss is introduced to ensure the semantic information in the reconstructed samples. A surrogate model is first trained under the guidance of the target model using knowledge distillation techniques. The semantic loss is then computed and serves as the classification loss of the target label provided by the surrogate model. 

Previous methods usually attach great importance to the confidence of the target class while neglecting the information from other classes' confidences \cite{zhang2023analysis} that measure the differences between each predicted vector of the same target class. To utilize these differences to strengthen sample-wise reconstruction, they insert a \textit{nonlinear amplification layer} ahead of the inversion model to enlarge minor entries within the prediction vector, amplifying the differences between predicted vectors with the same class and further promoting the inversion model to achieve better sample-wise recovery. 
Recent work \cite{zhang2024aligning} considers the domain divergence issue between the auxiliary and the private dataset. 
% and proposes a different insight to overcome this challenge. 
Inspired by the principle of generative adversarial learning, they regard a pre-trained auxiliary classifier as the \textit{generator}, while jointly adversarially training a domain \textit{discriminator} to distinguish the generated and the real private confidence vectors. Under the guidance of the domain \textit{discriminator}, the attacker fine-tunes the \textit{generator} to produce prediction vectors whose domain aligns with the private ones,
Correspondingly, the inversion model $\mathcal{M}(\cdot)$ is trained simultaneously with the generated auxiliary prediction vectors as input. Moreover, they also adopt the additional \textit{nonlinear amplification layer} \cite{zhang2023analysis} and achieve the recovery of the RGB images on shallow networks for classification tasks. 
In addition, they \cite{zhang2024aligning} further explore the scenario where attackers have access to private features extracted by the target model. The attacker then removes the \textit{nonlinear amplification layer} and adopts a specially designed feature alignment model as the \textit{generator} to utilize the extracted features better.

Although these early MI attacks achieve high efficiency and low computational overhead on traditional ML models, they are limited by the shortcomings of insufficient image generation capability and optimization algorithms. As a result, they perform well solely on shallow models and grayscale images, and fail to conduct valid attacks in more practical scenarios for DNN models or high-dimensional RGB images.

\label{threat_model}
% BY FH, YWB 

\section{MI Attacks on DNNs}
\label{sec:mi_main}
%
% --------------------------------------------------------
% BY YHY 
% Model inversion attacks are aimed at reconstructing a private dataset from a release target model.In this section, we will provide a detailed description of model inversion attacks and defenses for visual modality.

% MODIFIED BY QYX
In this section, we provide detailed introductions to the core features and differences of current mainstream MI attacks on DNNs. 
% We summarize the major characteristics of representative attacks in Table \ref{table:cv_attack} and present a taxonomy of defense strategies in Section \ref{defense}.

% --------------------------------------------------------

% --------------------------------------------------------
% BY YHY
% --------------------------------------------------------

\def\cvresultwidth{0.095\linewidth}
\def\cvresultinnerwidth{0.98\linewidth}

\begin{figure*}[htbp]

\centering
\setlength{\tabcolsep}{1pt}
  \normalsize
% \begin{subfigure}{0.99\linewidth}
    \begin{minipage}[t]{\cvresultwidth}
    \centering
    \includegraphics[width=\cvresultinnerwidth]{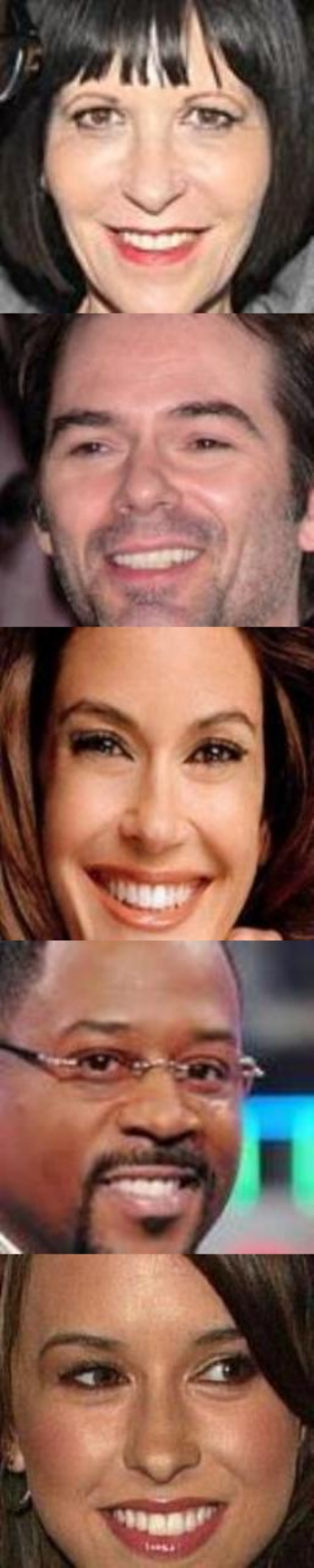}
    \centering
    \caption*{\textbf{\footnotesize{Ground Truth}}}
    \end{minipage}%
% 
    % \begin{minipage}[t]{\cvresultwidth}
    % \centering
    % \includegraphics[width=\cvresultinnerwidth]{gmi.pdf}
    % \centering
    % \caption*{\textbf{\footnotesize{GMI \cite{zhang2020secret}}}}
    % \end{minipage}%
% 
    % \begin{minipage}[t]{\cvresultwidth}
    % \centering
    % \includegraphics[width=\cvresultinnerwidth]{ked.pdf}
    % \centering
    % \caption*{\textbf{\footnotesize{KEDMI \cite{chen2021knowledge}}}}
    % \end{minipage}%
% 
    % \begin{minipage}[t]{\cvresultwidth}
    % \centering
    % \includegraphics[width=\cvresultinnerwidth]{mirror.pdf}
    % \centering
    % \caption*{\textbf{\footnotesize{Mirror (white) \cite{an2022mirror}}}}
    % \end{minipage}%
% 
    \begin{minipage}[t]{\cvresultwidth}
    \centering
    \includegraphics[width=\cvresultinnerwidth]{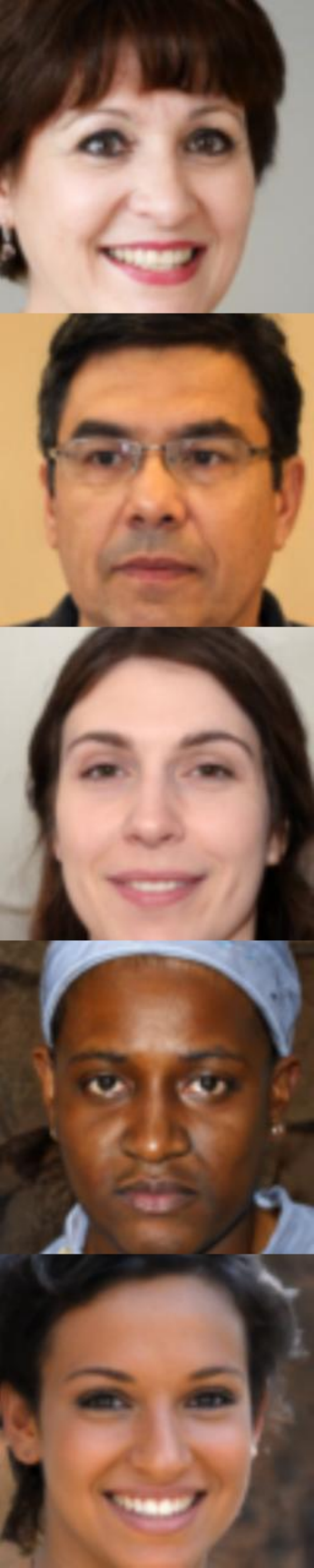}
    \centering
    \caption*{\textbf{\footnotesize{PPA \cite{struppek2022plug}}}}
    \end{minipage}%
    \begin{minipage}[t]{\cvresultwidth}
    \centering
    \includegraphics[width=\cvresultinnerwidth]{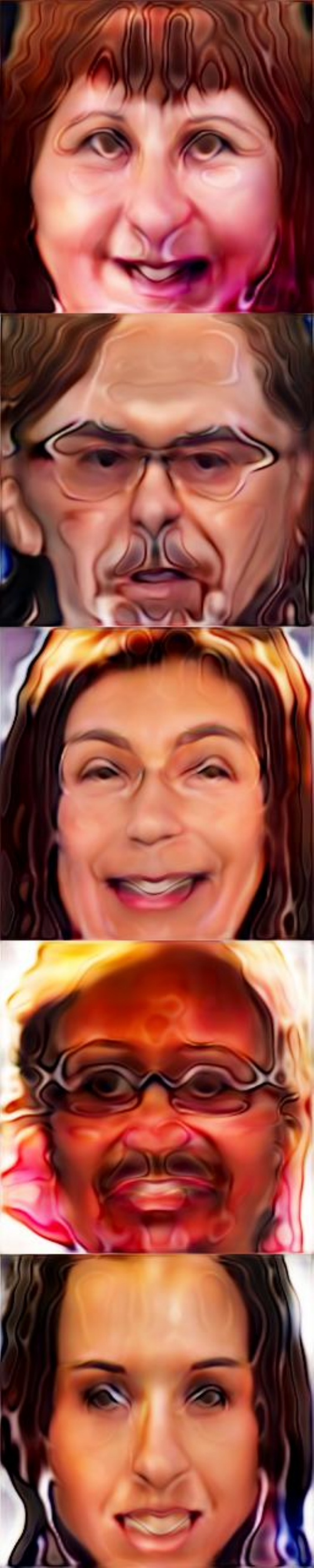}
    \centering
    \caption*{\textbf{\footnotesize{PLGMI \cite{yuan2023pseudo}}}}
    \end{minipage}%
    \begin{minipage}[t]{\cvresultwidth}
    \centering
    \includegraphics[width=\cvresultinnerwidth]{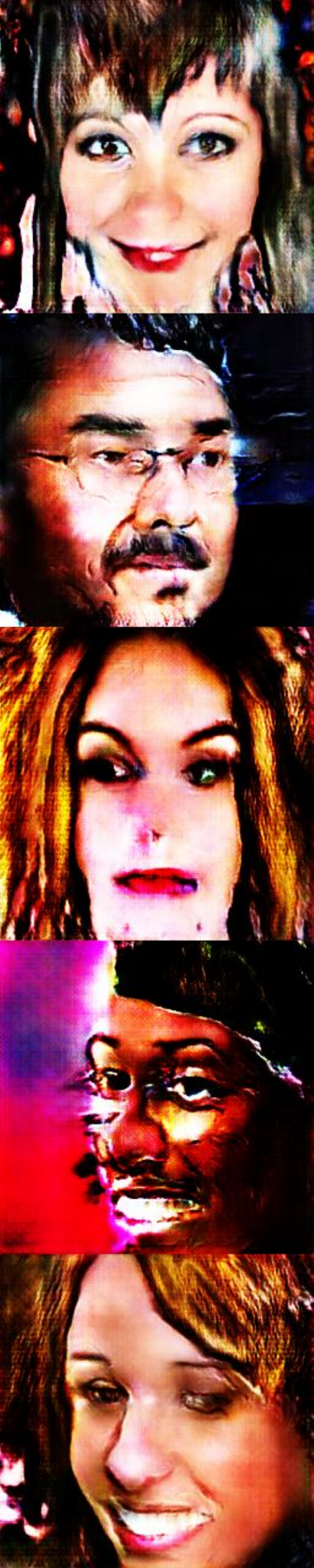}
    \centering
    \caption*{\textbf{\footnotesize{LOMMA +GMI \cite{nguyen2023re}}}}
    \end{minipage}%
    \begin{minipage}[t]{\cvresultwidth}
    \centering
    \includegraphics[width=\cvresultinnerwidth]{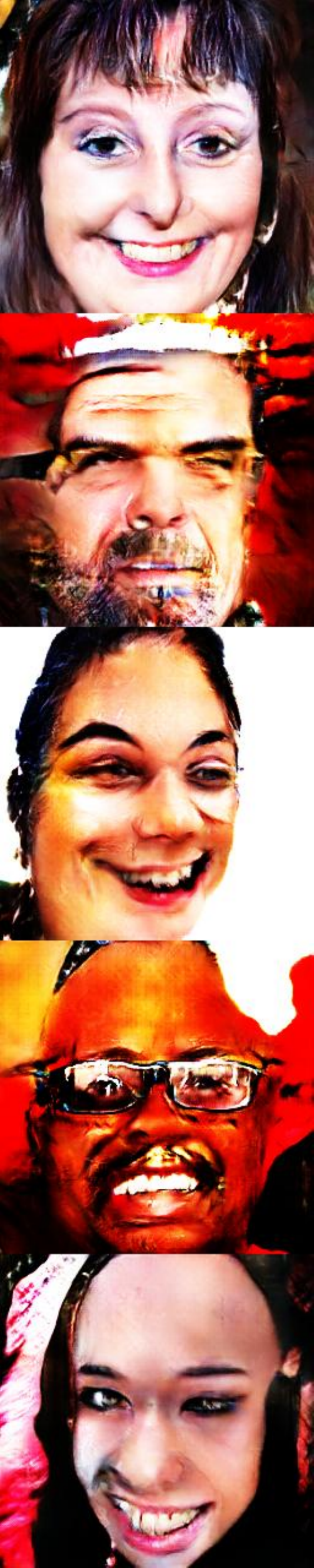}
    \centering
    \caption*{\textbf{\footnotesize{LOMMA +KED \cite{nguyen2023re}}}}
    \end{minipage}%
    \begin{minipage}[t]{\cvresultwidth}
    \centering
    \includegraphics[width=\cvresultinnerwidth]{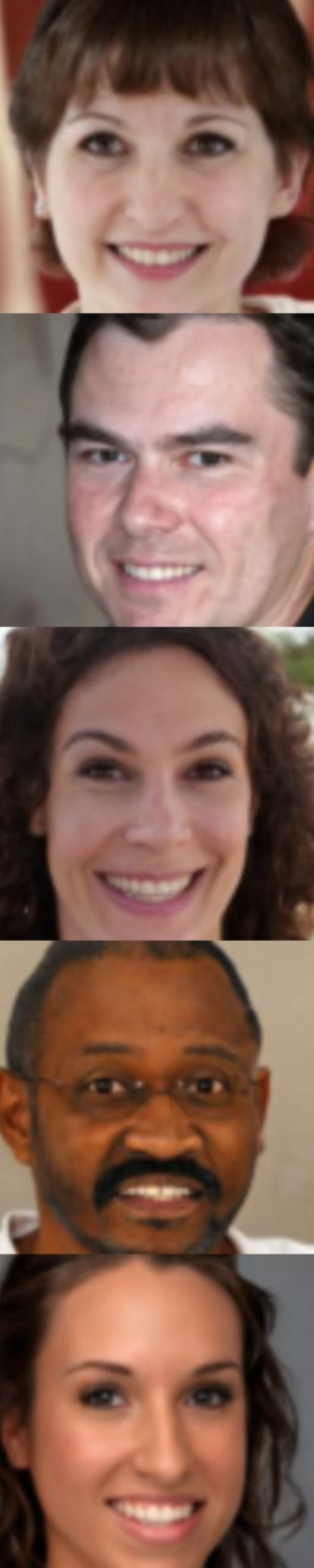}
    \centering
    \caption*{\textbf{\footnotesize{IF-GMI \cite{qiu2024closer}}}}
    \end{minipage}%
    \begin{minipage}[t]{\cvresultwidth}
    \centering
    \includegraphics[width=\cvresultinnerwidth]{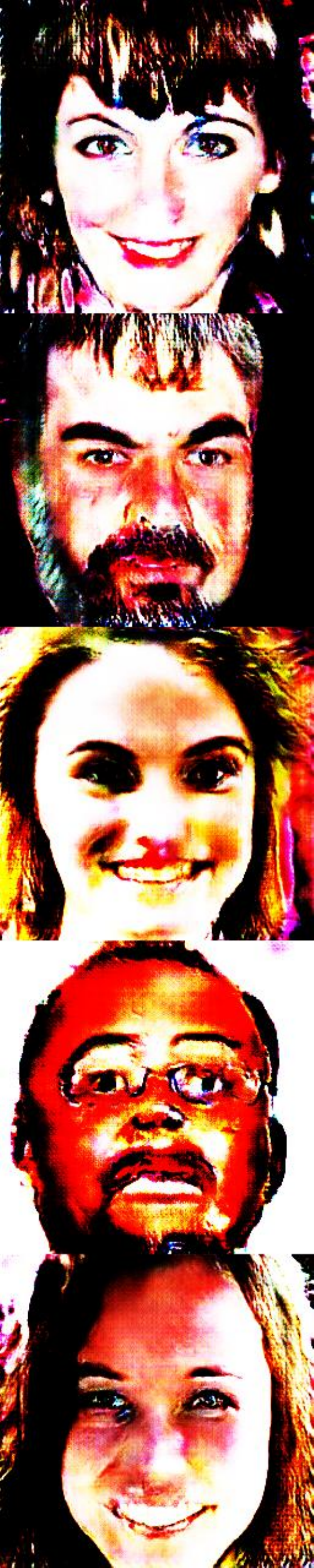}
    \centering
    \caption*{\textbf{\footnotesize{BREPMI \cite{kahla2022label}}}}
    \end{minipage}%
% 
    % \begin{minipage}[t]{\cvresultwidth}
    % \centering
    % \includegraphics[width=\cvresultinnerwidth]{mirror_black.pdf}
    % \centering
    % \caption*{\textbf{\footnotesize{Mirror (black) \cite{an2022mirror}}}}
    % \end{minipage}%
% 
%
    \begin{minipage}[t]{\cvresultwidth}
    \centering
    \includegraphics[width=\cvresultinnerwidth]{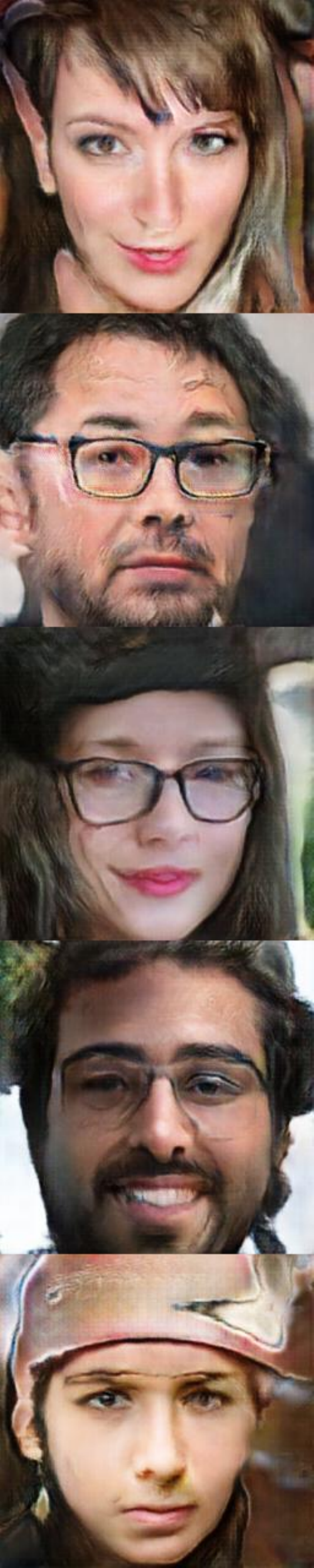}
    \centering
    \caption*{\textbf{\footnotesize{RLBMI \cite{han2023reinforcement}}}}
    \end{minipage}%
    \begin{minipage}[t]{\cvresultwidth}
    \centering
    \includegraphics[width=\cvresultinnerwidth]{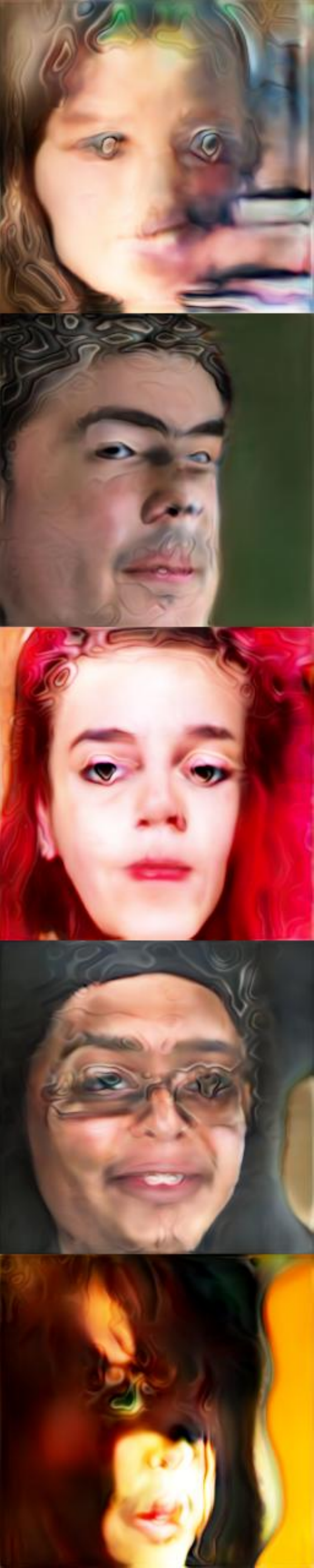}
    \centering
    \caption*{\textbf{\footnotesize{LOKT \cite{nguyen2023labelonly}}}}
    \end{minipage}%
    \begin{minipage}[t]{\cvresultwidth}
    \centering
    \includegraphics[width=\cvresultinnerwidth]{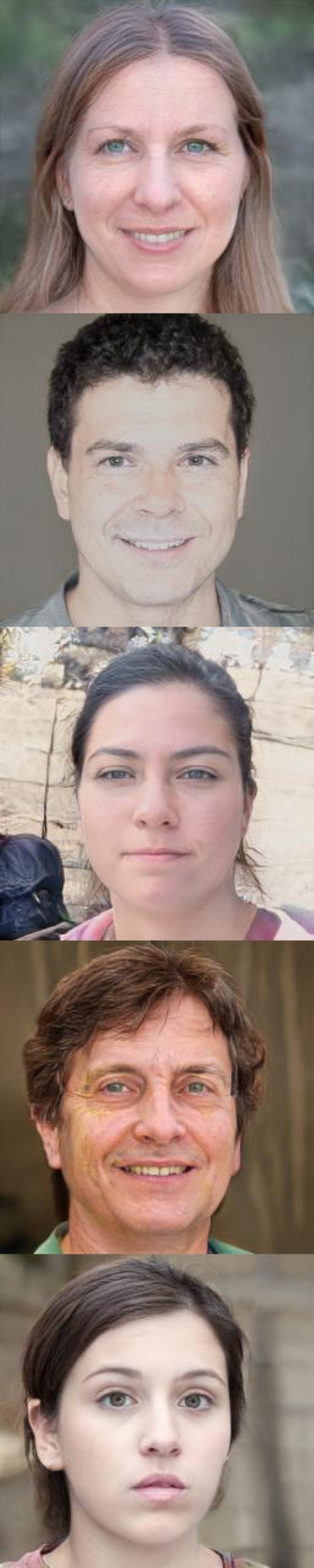}
    \centering
    \caption*{\textbf{\footnotesize{C2FMI \cite{ye2023c2fmi}}}}
    \end{minipage}
% \end{subfigure}
\caption{Visual comparison of representative SOTA attacks against image classification. It can be observed that these attacks (e.g., IF-GMI \cite{qiu2024closer}) cause serious privacy leakage about the semantic features in human faces. Note that the last four columns display black-box attacks.}
\label{fig:cv_reconstruct}
\end{figure*}

\subsection{Attacks for Image Classification}
% \subsection{Private Image Recovery}
% --------------------------------------------------------
% BY YHY
% We outline the workflow and types of model inversion attacks in this section. We begin by detailing the attack pipeline and various attack models. Next, we delve into the optimization process for both white-box and black-box scenarios. Finally, we explore advanced techniques in recent attack methods. A comprehensive comparison of these methods, highlighting their differences and similarities, is presented in Table

% \todo{(FFH) Emphasize that these attacks are the mainstream method of recent MI study, with excellent effect and great threat.}

% Model inversion attacks against image classification is the 
% In recent years, attacks against image classier are a major direction of model inversion attacks, especially against human face recognition classifier. By reconstructing the training data for a specific target class, the attacker can get the private face features of the specific humans.
Attacks on image classifiers, particularly those for human face recognition, have become a significant research focus of model inversion attacks. Figure \ref{fig:cv_reconstruct} shows the visual results reconstructed by several representative state-of-the-art (SOTA) attacks. By reconstructing training data for a given target class, an attacker is able to access an individual's private facial features, which poses a serious privacy breach threat.
% Some attack examples of different methods are presented in Figure \ref{cvvisual}. 

We begin by elucidating the general threat model of MI on DNNs. Then, to better clarify and compare these MI approaches, we disassemble the optimization process and summarize the major characteristics of current attacks in Table \ref{table:cv_attack}, based on which we systematically analyze the attack paradigm across four perspectives: 
\textit{Generative Models}, \textit{Data Initialization}, \textit{Attack Process} and \textit{Additional Generic Techniques}. Finally, we review multiple commonly used metrics to evaluate the effectiveness of MI attacks.  

% --------------------------------------------------------
\begin{figure}[t]
\centerline{\includegraphics[width=\columnwidth]{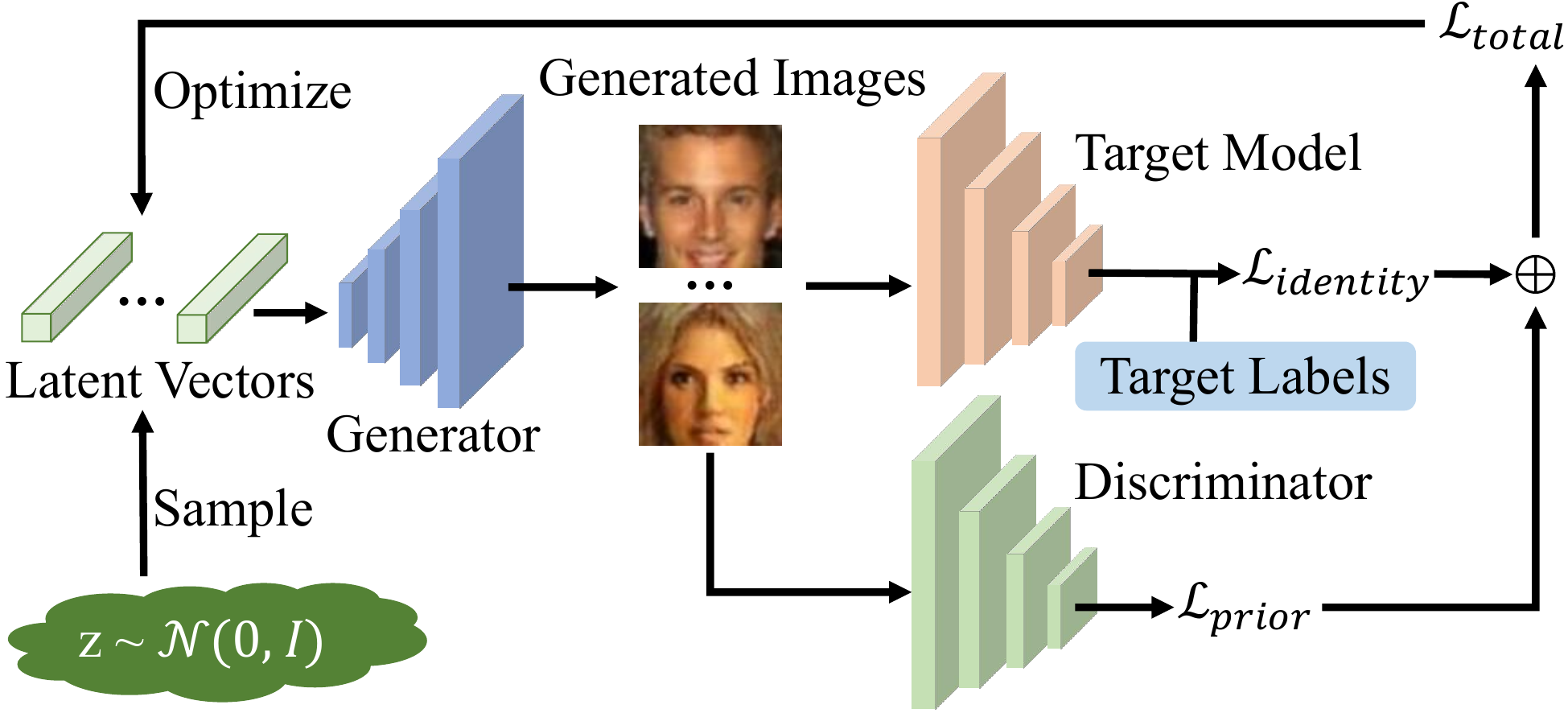}}
\caption{Threat model of generative model inversion attacks against image classification tasks.}
\label{basic_generative_pipeline_for_CV}
\end{figure}
% TODO: strengthen, establish, delve into, validate, examine

\subsubsection{Threat Model}
Due to the remarkable improvement facilitated by generative image priors, generative models have been incorporated into the basic paradigm for subsequent research of MI on DNNs \cite{zhang2020secret,wang2021variational,yuan2023pseudo}.  
Figure \ref{basic_generative_pipeline_for_CV} outlines the threat model for a typical GAN-based model inversion attack. Given the target image classifier $f_{\theta}(\cdot)$ parameterized with weights $\theta$ and the trained generator $G(\cdot)$, attackers attempt to recover private images $\mathbf{x}^*$ from the generator's latent vectors $\mathbf{z}$ that are initialized by the Gaussian distribution:
\begin{equation}
\begin{aligned}
    \mathbf{z}^{*} = \mathop{\arg\min}\limits_{\mathbf{z}} \mathcal{L}_{iden}(f_\theta(G(\mathbf{z})),c) + \lambda\mathcal{L}_{prior}(\mathbf{z};G),
\end{aligned}
\label{eq:attack_optim}
\end{equation}
\noindent where $c$ is the target class, $\mathcal{L}_{iden}(\cdot,\cdot)$ typically denotes the identity loss (i.e., the classification loss of the target label), $\lambda$ is a hyperparameter, and $\mathcal{L}_{prior}(\cdot)$ is the prior knowledge regularization, e.g., the discriminator's classification loss used to improve the reality of $G(\mathbf{z})$. Intuitively, the first term attempts to drive the generated images $G(\mathbf{z})$ to be classified as the desired category by the victim classifier, which essentially endeavors to reduce the generations' deviation from the private data based on the perspective of the target model $f_{\theta}(\cdot)$. The latter term serves as image priors to enhance the quality of the reconstructed samples. Before launching an attack, the adversary needs to acquire a generative model $G(\cdot)$ as prior information about the target images. This can be achieved through either manual training of a specialized generator \cite{zhang2020secret, yuan2023pseudo} or direct utilization of a pre-trained generator \cite{struppek2022plug}. Subsequently, the generator is leveraged to generate dummy images that are fed into the target classifier. 
During the recovery, the attacker performs iterative updates of the latent vectors $\mathbf{z}$ by minimizing the loss function in Eq. (\ref{eq:attack_optim}). Once obtaining the optimal $\mathbf{z}^{*}$, the reconstructed images that closely align with the private images can be generated by $\mathbf{x}^{*}=G(\mathbf{z}^{*})$. 
 
% --------------------------------------------------------
 \renewcommand{\dblfloatpagefraction}{.9}
% MODIFIED BY YWB
\begin{table*}[htbp]
  \setlength{\tabcolsep}{1pt}
  \normalsize
  \centering
  % \belowrulesep=0pt
  % \aboverulesep=0pt
  % MODIFIED BY QYX, YHY
  % yhy: 'that can attack DNNs' 删掉，有点冗余
  % conditional GAN放在生成模型章节里面讲了
  \begin{threeparttable} 

    \resizebox{\linewidth}{!}{
    % \begin{tabular}{cccccccp{14.785em}} \toprule
    \begin{tabular}{lccccccc} \toprule
    \multicolumn{1}{c}{\multirow{2}[0]{*}{Method}} & \multicolumn{1}{c}{\multirow{2}[0]{*}{Generative Models}} & \multicolumn{2}{c}{Data Initialization} & \multicolumn{3}{c}{Optimization} & \multicolumn{1}{c}{\multirow{2}[0]{*}{Extra Features}} \\
    \cmidrule{3-4}\cmidrule(lr){5-7}
          &       & Resolution & Preprocess & Optim-algorithm & Loss-fn  & Search Space &  \multicolumn{1}{c}{} \\
    \midrule
    \multicolumn{8}{c}{\textit{\textbf{\fontsize{12.0pt}{\baselineskip}\selectfont White-Box Attacks}}} \\
    \midrule
    % DeepInversion \cite{yin2020dreaming} & -     & $224\times 224$ & \XSolidBrush & Adam &  $\mathcal{L}_{CE}+\mathcal{R}_{prior}$ &  Pixel space &  \multicolumn{1}{c}{-} \\
    % \cmidrule{1-8}
    GMI \cite{zhang2020secret}   & \multicolumn{1}{p{11.785em}}{\makecell[c]{WGAN-GP}} & $64\times 64$ & \XSolidBrush & Momentum SGD & $\mathcal{L}_{CE}+\lambda\mathcal{L}_{Dis}$ & $\mathcal{Z}$ space & \multicolumn{1}{c}{-} \\
    \cmidrule{1-8}
    KEDMI \cite{chen2021knowledge} & Inversion-specific GAN$^{\dag}$ & $64\times 64$ & \XSolidBrush & Adam & $\mathcal{L}_{CE}+\lambda\mathcal{L}_{Dis}$ & $\mathcal{Z}^*$ space  & \makecell[c]{Distributional recovery} \\
    \cmidrule{1-8}
    % VMI   & \multicolumn{1}{p{11.785em}}{\makecell[c]{DCGAN \\ StyleGan2-Ada}} & \multicolumn{1}{p{7.5em}}{64× 64,\newline{}128× 128} & \multicolumn{1}{p{11.785em}}{\XSolidBrush} & $\mathcal{Z}$ space & Variational loss & Adam  & \multicolumn{1}{c}{Variational inference} \\
    VMI \cite{wang2021variational}   & \multicolumn{1}{p{11.785em}}{\makecell[c]{DCGAN \\ StyleGAN2-Ada}} & \multicolumn{1}{c}{\makecell[c] {$64\times 64$\\ $128\times 128$}} & \XSolidBrush & Adam & $\mathcal{L}_{CE}+\lambda\mathcal{L}_{KL}$ & $\mathcal{Z}$ space  & \multicolumn{1}{c}{Variational inference} \\
    \cmidrule{1-8}
    $\alpha$-GAN-MI \cite{alphaganmi}& $\alpha$-GAN & $64\times 64$ & \CheckmarkBold & - & $\mathcal{L}_{CE}$ & $\mathcal{Z}$ space & \multicolumn{1}{c}{-} \\
    \cmidrule{1-8}SecretGen \cite{yuan2022secretgen}& WGAN-GP & $64\times 64$ & \CheckmarkBold & Momentum SGD & $\mathcal{L}_{CE}+\lambda\mathcal{L}_{Dis}$ & $\mathcal{Z}$ space & \multicolumn{1}{c}{-} \\
    \cmidrule{1-8}
    Mirror \cite{an2022mirror}& StyleGAN & $224\times 224$ & \CheckmarkBold & Adam & $\mathcal{L}_{CE}$ & $\mathcal{P}$ space & \multicolumn{1}{c}{$\mathbf{p}$ vector Clipping} \\
    \cmidrule{1-8}
    PPA \cite{struppek2022plug}& \makecell[c]{BigGAN \\ StyleGAN2-Ada} & $224\times 224$ & \CheckmarkBold & Adam & $\mathcal{L}_{Poincar\acute{e}}$ & $\mathcal{W}$ space & \multicolumn{1}{c}{Results selection} \\
    \cmidrule{1-8}
    PLGMI \cite{yuan2023pseudo}& Conditional GAN & $64\times 64$ & \XSolidBrush & Adam & $\mathcal{L}_{MM}$ & $\mathcal{Z}$ space & \makecell[c]{Pseudo labels guidance} \\
    \cmidrule{1-8}
    LOMMA \cite{nguyen2023re}& $\sim$ & $\sim$ & $\sim$ & $\sim$ & $\mathcal{L}_{Logit} + \lambda\mathcal{L}_{Feature}$ & $\sim$ & \multicolumn{1}{c}{Model augmentation} \\ 
    % \multicolumn{1}{p{11.785em}}{\makecell[c] {WGAN \\ Inversion-specific GAN$^{\dag}$}}
    % \multicolumn{1}{p{10.145em}}{\makecell[c]{Momentum SGD \\ Adam}}
    \cmidrule{1-8}
    DMMIA \cite{qi2023model} &  StyleGAN2-Ada & $224\times 224$ &  \XSolidBrush & Adam & $\mathcal{L}_{CE} + \mathcal{L}_{Memory}$ & - & \makecell[c]{Dynamic memory bank} \\
    \cmidrule{1-8}
    % Leverage historically\\ learned knowledge
    IF-GMI \cite{qiu2024closer} & StyleGAN2-Ada & $224\times 224$ & \CheckmarkBold & Adam & $\mathcal{L}_{Poincar\acute{e}}$ & $\mathcal{W}$ space & \multicolumn{1}{c}{Intermediate feature optimization} \\
    \cmidrule{1-8}
    PPDG-MI \cite{ppdg} & $\sim$ & $\sim$ & $\sim$ & $\sim$ & $\sim$ & $\sim$ & \multicolumn{1}{c}{Generator fine-tuning} \\
    \cmidrule{1-8}
    AlignMI \cite{alignmi} & $\sim$ & $\sim$ & $\sim$ & $\sim$ & $\mathcal{L}_{avg}$ & $\sim$ & \multicolumn{1}{c}{Training-Free alignment promotion} \\
    \midrule
    \multicolumn{8}{c}{\textit{\textbf{\fontsize{12.0pt}{\baselineskip}\selectfont Black-Box Attacks}}} \\
    \midrule
    SecretGen \cite{yuan2022secretgen}& WGAN-GP & $64\times 64$ & \CheckmarkBold & Momentum SGD & $\mathcal{L}_{Dis}$ & $\mathcal{Z}$ space & \multicolumn{1}{c}{-} \\
    \cmidrule{1-8}
    BREPMI \cite{kahla2022label}& WGAN-GP & $64\times 64$ & \CheckmarkBold & Gradient estimation & -  & $\mathcal{Z}$ space & \multicolumn{1}{c}{-} \\
    \cmidrule{1-8}
    Mirror \cite{an2022mirror}& StyleGAN & $224\times 224$ & \CheckmarkBold & Genetic algorithm & -   &  $\mathcal{P}$ space & \multicolumn{1}{c}{$\mathbf{p}$ vector Clipping} \\
    \cmidrule{1-8}
    BMI-S \cite{xu2023sparse}& StyleGAN2-Ada & $224\times 224$ & \XSolidBrush & Gradient estimation   & - & $\mathcal{W}$ space  & \multicolumn{1}{c}{Style vectors selection} \\
    \cmidrule{1-8}
    C2FMI \cite{ye2023c2fmi}& StyleGAN2 & $128\times 128$ & \CheckmarkBold & Genetic algorithm  & -     & $\mathcal{W}$ space & \multicolumn{1}{c}{Initial coarse optimization} \\
    \cmidrule{1-8}
    % 改名了？
    RLBMI \cite{han2023reinforcement}& WGAN-GP & $64\times 64$ & \XSolidBrush & \multicolumn{1}{c}{SAC}  & - & $\mathcal{Z}$ space     & Reinforcement learning \\
    \cmidrule{1-8}
    LOKT \cite{nguyen2023labelonly} & T-ACGAN &  $128\times 128$ & \XSolidBrush & Adam & $\mathcal{L}_{Max-margin}$ & $\mathcal{Z}$ space & Invert surrogate models \\ \cmidrule{1-8}
    DiffusionMI \cite{liu2023diffusion}& Conditional Diffusion & $64\times 64$ & \XSolidBrush & \makecell[c]{Optimization-free}     & -     & -     & Results selection  \\ \cmidrule{1-8}
    P2I-MI \cite{liu2024prediction}& StyleGAN2-Ada & $64\times 64$ & \XSolidBrush & \makecell[c]{Optimization-free}     & -     & -     &  Prediction alignment  \\ \cmidrule{1-8}
    SMILE \cite{smile} & StyleGAN & $224\times 224$ & \CheckmarkBold & MetaModelOnePlusOne \cite{beyer2002evolution} & -   &  $\mathcal{P}$ space & \multicolumn{1}{c}{Long-tailed learning} \\
    \bottomrule
    \end{tabular}}
    \begin{tablenotes}[flushleft] %添加此处
	\item {\fontsize{7.5pt}{8pt}\selectfont  $^*$The latent vectors are sampled from a learnable Gaussian Distribution. }
        \item {\fontsize{7.5pt}{8pt}\selectfont $^{\dag}$This customized GAN structure is first proposed by KEDMI \cite{chen2021knowledge}.}
        \item {\fontsize{7.5pt}{8pt}\selectfont $^{\sim}$ LOMMA \cite{nguyen2023re}, PPDG-MI \cite{ppdg} and AlignMI \cite{alignmi} are plug-and-play techniques that can seamlessly adapt to generative attacks.}
     \end{tablenotes} %添加此处
    \end{threeparttable} %添加此处
    % \vspace{-0.5em}
    \caption{Summary of existing MI attacks for visual tasks. Note that SecretGen and Mirror consider both white-box and black-box scenarios.}
    % Note that $\sim$ represents that LOMMA is a plug-and-play technique that can perfectly combine with existing model inversion attacks.}
    % \vspace{-1em}
  \label{table:cv_attack}%
\end{table*}%

%--------------------------------------------------------
% \subsubsection{Generative Models}
% % 原图，GAN， diffuison   
% BY YHY
\subsubsection {Generative Models}
% no model
% Initial approaches in MI attacks ignore the usage of auxiliary models, opting instead to directly optimize on random images with Gauss distribution \cite{yin2020dreaming}.
% However, this method of direct optimization in high-dimensional space often leads to the overfitting of the target model and a scarcity of semantic content, resulting in poor generative outcomes.
% Generating images by directly optimizing in high-dimensional image space often leads to the overfitting of the target model and a scarcity of semantic content \cite{zhang2020secret}, resulting in poor generative outcomes.
% 

% MODIFIED BY QYX
% Early studies directly searched image pixels to perform MI attacks, which were limited to shallow networks and gray images. 
As introduced in Sec. \ref{sec:mi_ml}, most previous studies on traditional ML scenarios directly optimize the random noises and neglect the benefits of powerful generative priors, resulting in poor reconstruction outcomes when faced with high-dimensional images \cite{zhang2020secret}. 
% Therefore, these methods are limited to shallow networks and gray images. 
% improve
% To overcome this, GANs and diffusion models are employed to gain prior knowledge about the target data for more effective reconstructed image generation. 
To overcome this difficulty, GMI \cite{zhang2020secret} proposes to employ GAN models to guide image generation and achieve outstanding improvements on deeper CNNs with RGB images. They manually train a GAN model on a large public dataset $X_{pub}$, which contains images with similar structures to the private data. Subsequent studies follow this GAN training paradigm \cite{kahla2022label,han2023reinforcement} and design special loss functions \cite{zhang2020secret, yuan2022secretgen} for effective GAN learning or utilize various GAN architectures tailored to the MI scenario, as summarized in Table \ref{table:cv_attack}. 
% VAE-GAN
% Based on 
% \cite{alphaganmi}
% Condition
Moreover, some methods further exploit information from victim models to strengthen the generator's capability. A widely used technique involves feeding the GAN's training images into the victim classifier to obtain pseudo labels, which then serve as class-condition inputs to the generator \cite{yuan2023pseudo,nguyen2023labelonly} and the discriminator \cite{chen2021knowledge,yuan2023pseudo,nguyen2023labelonly} in the generative adversarial training.
% 垚，引用加全了吗？怎么感觉还是漏了不少
% 
% 
% \cite{zhang2020secret}.
% 
% 
% BY YHY
% Nonetheless, training a GAN specifically for attack purposes takes a long time, and the quality of images generated by the GAN itself is not high enough, thus confining the generated images to a resolution of $64\times 64$.

% MODIFIED BY QYX
However, constrained by the limited data volume and poor quality of auxiliary datasets for GAN training, the trained generator is only capable of recovering low-resolution images at $64\times64$. Additionally, performing generative adversarial learning incurs expensive computational overheads and consumes a significant amount of time.
% style gan
% In response to these limitations, \cite{an2022mirror,struppek2022plug} use pre-trained StyleGANs to significantly enhance the resolution of generated images up to $256\times 256$ or even $1024\times 1024$.This approach not only improves image quality but also reduces the resource demands and time associated with conducting the attack.
% MODIFIED BY QYX
One viable resolution is to utilize publicly available pre-trained GANs that have been well-trained on large image datasets \cite{wang2021variational,an2022mirror,ye2023c2fmi,liu2024prediction}. Compared with GANs trained manually with low-quality auxiliary datasets, these pre-trained generators typically have more intricate structures and hold richer semantic information, enabling higher dimensional reconstruction up to $224\times 224$ resolution \cite{struppek2022plug, qi2023model, qiu2024closer,xu2023sparse}.

% √ :)
% diffuresosion
% fh: 在LPIPS和KNN Dist这种感知上的距离有更好的表现, 得写全面，说它主实验在xx指标上更差，但是在感知指标上更好。然后label-only的情况下又xxx  简单描述下它的思路, 是去训了一个
% \cite{liu2023diffusion} introduces the diffusion method, using categories as conditions to guide the image generation. The research achieves a higher attack accuracy under label-only conditions.
Furthermore, the recent work DiffusionMI \cite{liu2023diffusion} substitutes GANs with the prevalent conditional diffusion model \cite{ho2022classifier}. Specifically, they leverage pseudo-labels classified by the target model as input conditions to train a denoising model with better guidance. This approach demonstrates notable improvements in generation quality over former GAN-based methods, particularly in terms of human perception distance (e.g., LPIPS \cite{zhang2018unreasonable}).

Overall, the evolution trend of generative priors in current MI attacks is characterized by continuously enhancing image generation capability while reducing the deployment overhead for attackers, thereby making the attack algorithms more applicable to real-world scenarios.

\subsubsection{Data Initialization}
We then introduce the features of data initialization in MI.
% BY QYX
\textbf{Resolution.}
% yhy: 表格中的分辨率是目标模型的分辨率。反演模型的分辨率可能更大
% The resolution shown in table \ref{table:cv_attack} is the resolution of the target model
We summarize the image resolutions employed by different algorithms in Table \ref{table:cv_attack}. Given that more image pixels imply more parameters to optimize, the difficulty of MI attacks is highly determined by the resolution of recovered images. Early MI methods resort to reconstructing low-resolution images for improved performance, \textit{e.g}., GMI \cite{zhang2020secret} and KEDMI \cite{chen2021knowledge}.
To enable attacks on higher-resolution images, subsequent studies introduce stronger pre-trained GAN models (e.g., StyleGAN \cite{karras2019style}) and manage to recover images with a resolution of $224\times 224$.

% 224×224 -> $224\times 224$
% reconstructed images 最大是1024，224只是分类器的resolution
% 这里的重建图像就是为了区别于generated image，生成图像是1024，而重建图像就是最终输入到目标模型那个图像
% ok, 那其它地方也改一下，然后要不要再别的地方提一下这两个的区别
% 不必了，谈到分辨率一律表示重建图片的分辨率
% --------------------------------------------------------
% BY YHY
% initial vector selection
% Due to the complexities arising from entanglement in the search space, some initial latent vectors pose challenges in being optimized effectively, limiting the quality of reconstructed images. 
% % 
% \cite{an2022mirror,struppek2022plug} solve the problem by sampling a large set of hidden vectors from a Gaussian distribution and selecting the best few as initial vectors. In conjunction with this, \cite{an2022mirror} use $l_1$ ball constraint to limit the search space.

% MODIFIED BY QYX
% yhy: a large number ? 普通的只取batch size个vector，不像mirror这些取2000, 200000个这么多
% PPA也取得是2000、5000这样的
%  PPA不是也选择了吗
% 第一句的意思应该是早期的他不select，直接上来就优化，这些肯定是比较少的
% ok，我再看看怎么润色一下

% 感觉现在不错
% 彳亍，那这里就算完成了
\textbf{Preprocess.}
\label{para:preprocess}
While \cite{zhang2020secret,chen2021knowledge} randomly sample initial latent vectors from certain data distributions, \cite{yuan2022secretgen,struppek2022plug} highlight that a batch of appropriately initialized latent vectors tends to yield better reconstruction results. 
A common and effective approach is to employ the classification confidence to assist the selection process \cite{an2022mirror,yuan2022secretgen}.
To obtain those latent vectors, the attacker first samples a large number of latent vectors and generates their corresponding images. These images then undergo a series of data augmentation transformations and are fed into the target classifier to obtain their corresponding confidence scores. The latent vectors with the highest confidence scores as the target class are therefore selected for the later optimization process \cite{struppek2022plug,qiu2024closer}.
In addition, C2FMI \cite{ye2023c2fmi} further proposes a coarse inversion strategy that additionally trains an inverse network $\mathcal{M}(\cdot)$ to improve the initialization process.
The attacker first feeds the images $\mathbf{x}$ into the target model $f_{\theta}(\cdot)$ and a pre-trained feature extractor $\xi(\cdot)$ to acquire predicted confidences and extracted features, respectively. The obtained confidence-feature pairs are leveraged to train the inverse network $\mathcal{M}(\cdot)$ that maps the confidence scores back to the feature space of $\xi(\cdot)$, which then serves as a surrogate model for attackers to perform white-box MI attacks toward the target label. Finally, these optimized latent vectors that can synthesize images close to the target category are subsequently used as the initial latent vectors.
% The expected output of the target model, a confidence vector with a high value for the target label, is then mapped to the expected output of the pre-trained feature extractor. 
% The latent vectors are optimized to minimize the mean squared error (MSE) loss between the expected output and the output of the feature extractor for images generated by these latent vectors. 
% 
\cite{alphaganmi} conducts this process from a different perspective, which initializes the latent vectors based on real facial features with the image encoder of the $\alpha$-GAN \cite{rosca2017variational}.
In practice, the attacker averages several randomly sampled public images and then inputs the averaged image into the $\alpha$-GAN \cite{rosca2017variational}'s image encoder to obtain the initial latent vector. 

To obtain extra knowledge gains, GMI \cite{zhang2020secret} and SecretGen \cite{yuan2022secretgen} additionally consider the scenario where attackers can access the blurred or masked versions of private images and fully leverage them as auxiliary information to enhance the initialization of latent vectors.
Specifically, the extracted features of corrupted images are concatenated with randomly initialized noise as input latent vectors to the generator, converting the image generation task into an image restoration task to enhance their underlying semantic information. 

\subsubsection{Attack Process}

% --------------------------------------------------------
% BY YHY
% The scenarios of MI attacks can be divided into white-box and black-box. In the white-box scenario, attackers have complete access to the structure and parameters of the target model. 
% % 
% In contrast, the black-box scenario restricts them to only the model's predictions, such as confidence vectors or labels, without revealing internal details like gradients.
% To better analyze and compare different reconstruction processes, we categorize these methods into white-box and black-box attacks based on the attacker's capability and knowledge. Specifically, a white-box scenario implies that attackers have full access to the weights and outputs of the target model. Conversely, in black-box settings, only the predicted confidence probabilities or hard labels are available. 
% We then introduce the optimization algorithms and loss functions of different methods. 

% --------------------------------------------------------

% --------------------------------------------------------
% BY YHY
\textbf{White-Box Attacks.}
% CE Loss
% White-box attacks leverage gradients to optimize latent vectors, primarily using cross-entropy loss for this purpose.
In the white-box scenario, the attackers have full access to the weights and outputs of the target model, where they primarily conduct the inversion by employing gradient optimization with Momentum SGD or Adam optimizer to minimize the loss function in Eq. (\ref{eq:attack_optim}).
% As summarized in Table \ref{table:cv_attack}, white-box attacks generally apply Momentum SGD or Adam optimizer to conduct the inversion. 
This involves computing the identity loss $L_{iden}$ and the prior loss $L_{prior}$. 
As summarized in Table \ref{table:cv_attack}, the majority of white-box methods adopt cross-entropy (CE) loss as the metric to compute the identity loss.
% Most of them update the latent codes using the gradients computed from the Cross-Entropy (CE) loss. 
% 梯度消失
% However, this loss function tends to suffer from vanishing gradients, which hinders the optimization process. To overcome this, \cite{struppek2022plug} move the optimization from Euclidean space to hyperbolic space, employing Poinc\'are loss, while \cite{yuan2023pseudo} adopt max-margin loss to address this issue.
% However, this loss function is susceptible to the problem of vanishing gradients, which impedes the optimization process. To mitigate this, the adoption of Poinc\'are loss \cite{struppek2022plug} and max-margin loss \cite{yuan2023pseudo} are propose as effective alternatives.
% Considering that CE loss suffers from the gradient vanishing problem, researchers propose to employ Poincar\'e loss \cite{struppek2022plug} or max-margin loss \cite{yuan2023pseudo} to mitigate this issue.
However, PPA \cite{struppek2022plug} proves that the cross-entropy loss is susceptible to gradient vanishing issues, which makes it challenging to obtain the optimal latent vectors during the reconstruction. To solve this problem, PPA moves the optimization to hyperbolic spaces with constant negative curvature by employing the Poincaré loss function, which can guarantee sufficient gradient information to the iteration process. 
Specifically, the $l_1$ normalized output confidences $\hat{y}'$ and the target one-hot vector $y$ are viewed as two points in a Poincaré ball, and the loss is formulated as the hyperbolic distance between two points:
% as shown in Eq. \ref{eq:poincare distance}.
\begin{equation}
\label{eq:poincare_distance}
\mathcal{L}_{\text{Poincar\'e}} = \text{arccosh}(1+\frac{2||\hat{y}'-y||_2^2}{
(1-||\hat{y}'||_2^2)(1-||y||_2^2)
}).
% \mathcal{L}_{\text{Poincar\'e}} = \text{arccosh}(1+\frac{2||\hat{\mathbf{y}}-{\mathbf{y}}||_2^2}{
% (1-||\hat{\mathbf{y}}||_2^2)(1-||{\mathbf{y}}||_2^2)
% }).
\end{equation}
PPA adopts an edited version of the typical Poincaré loss by replacing $1$ with $0.9999$ for enhanced attack performance.
Another effective solution proposed by PLGMI \cite{yuan2023pseudo} is to substitute CE loss with the max-margin (MM) loss function $\mathcal{L}_{MM}$, which maximizes the output logit of the target class while minimizing that of an additional category with the largest logit:
\begin{equation}
\label{eq:max_min}
\mathcal{L}_{MM} = -l_c(x)+\max\limits_{j\neq c}l_j(x),
\end{equation}
\noindent where $c$ is the target class and $l_{i}$ represents the output logit of the class $i$.
The resultant derivative of max-margin loss concerning the logits contains constant values that can help avoid the gradient vanishing problem. 
Moreover, PLGMI points out that the max-margin loss facilitates searching for samples that closely resemble the target class while distinctly different from other classes, which aligns with the basic goal of MI attacks. 
Nevertheless, LOMMA \cite{nguyen2023re} rethinks the optimization objective and proposes a different viewpoint. They emphasize that the fundamental goal of MI attacks is to reconstruct images that highly resemble those in the target class, rather than deviating from non-target classes. Since CE loss inherently combines both objectives, they suggest bypassing the softmax function in the cross-entropy loss and directly maximizing the logit of the target class. 
Recent work AlignMI \cite{alignmi} has deciphered the core mechanism of MIAs through the geometric lens of the manifold hypothesis. They reveal that better alignment between loss gradients and the generator manifold leads to more serious vulnerability. Building on this insight, they propose to sample multiple variants within the local neighborhood of synthetic samples during optimization and compute the average of the corresponding loss gradients. This operation suppresses off-manifold noise and enhances gradient-manifold alignment, achieving strong attack in a training-free manner.

% 去掉无关优化
% Moreover, \cite{nguyen2023re} point out that the objective should be to approach the target class rather than to distance from non-target classes, but cross-entropy loss combines both. They propose bypassing the softmax function in cross-entropy loss for direct optimization towards the target class's score.

 % 

 As mentioned in Section \ref{threat_model}, various regularization terms are introduced as prior knowledge to ensure the quality and fidelity of the generated images. One representative attempt is the use of the discrimination classification loss $\mathcal{L}_{Dis}$, i.e., the realism penalization from GAN's discriminator \cite{zhang2020secret, yuan2022secretgen}. 
 However, \cite{struppek2022plug,an2022mirror} note that the GAN is trained on a public dataset and thus the $\mathcal{L}_{Dis}$ might impair the inversion performance by matching the generated images with the distributions of the public data rather than the private ones. Consequently, subsequent MI methods no longer employ this term to avoid the distribution shift.
 %
 % TODO: Yixiang, rectify the following content to make it more clear!!!
 % TODO: Yixiang, add an introduction to the DMMIA!
% \in Q_{\mathbf{x}}$, where $Q_{\mathbf{x}}$ is a variational family.
% 
% To ensure in-distribution optimization, \cite{an2022mirror} clip the Gaussian-like $\mathcal{P}$-space with the variance of large random samples. 
Apart from the constraint on the confidence scores, LOMMA \cite{nguyen2023re} analyzes that successful MI attacks are expected to require the feature similarity between the reconstructed and private images given by the target model. Therefore, an $\mathcal{L}_{Feature}$ term is utilized to minimize the distance between the penultimate layer representations of reconstructed samples and public images, further boosting the performance of MI attacks. 

% 要修改，从motivation开始
Unlike the above methods, DMMIA \cite{qi2023model} focuses on the issue in the GAN model itself, i.e., the catastrophic forgetting problem where the generated images from early reconstruction contain more diverse characteristics, but some disappear as the optimization goes further, decreasing the diversity of inverted images. 
To solve this issue, DMMIA uses the generated samples to build two types of memory banks, which are the learnable intra-class multicentric representation (IMR) and the non-parametric inter-class discriminative representation (IDR), respectively. IMR learns multiple distinctive features of intra-class images to represent the target class while IDR stores the historical knowledge into prototypes for each class. 
Based on the two dynamically maintained memory banks, they propose a novel regularizer $\mathcal{L}_{Memory}=\lambda _{1}\mathcal{L}_{imr}+\lambda _{2}\mathcal{L}_{idr}$, where $\mathcal{L}_{imr}$ leverages IMR to prevent the optimization from overfitting to specific image features and $\mathcal{L}_{idr}$ encourages the reconstruction of images with more class-distinguishable characteristics. 

Instead of instance-level data reconstruction, KEDMI \cite{chen2021knowledge} aims to directly recover the private data distribution for a given label with a learnable Gaussian distribution $\mathcal{N}(\mu, \sigma^2)$. They adopt the reparameterization trick to make the loss differentiable and optimize the learnable parameters $\mu$ and $\sigma$. Once finished modeling the private data distribution, attackers can directly sample $z$ from the learned distribution to generate sensitive images. VMI \cite{wang2021variational} also discusses the distribution-level recovery of the training data. To establish theoretical understanding, VMI formulates the MI attack as a variational inference that attempts to estimate the target posterior distribution $p(\mathbf{x}|y)$ with a variational distribution $q(\mathbf{x})$. Based on the theoretical analysis, an extra KL-divergence regularizer is introduced to constrain the distance between a learnable distribution $q(\mathbf{z})$ estimated by deep flow models \cite{NEURIPS2018_d139db6a} and the latent distribution $p_{AUX}(\mathbf{z})$ of public auxiliary images in the GAN's latent space.
By optimizing the flow model using the proposed loss function, the attacker gradually optimizes the latent distribution $q(\mathbf{z})$ to reach the latent area of target data, further estimating the target data distribution $q(\mathbf{x})$ through the GAN model $G(\cdot)$, i.e., $q(\mathbf{z}) \stackrel{G}{\longrightarrow} q(\mathbf{x})$.

While the latest MI attacks have achieved outstanding performance in white-box settings, their reliance on white-box access to the target model restricts their applicability exclusively to offline-deployed or open-source models, making them inapplicable to real-world scenarios where service providers only expose black-box API interfaces.

% abandoned losses
% Some initial approaches use discriminator loss \cite{zhang2020secret,chen2021knowledge,yuan2022secretgen} 
% and various regularization terms \cite{yin2020dreaming}, such as Batch Normalization (BN) layer normalization, 
% to steer the optimization process. 

% 

% why dont use
% However, \cite{struppek2022plug} highlight a critical drawback of this strategy: it tends to align the generated images more closely with the distributions of public training datasets rather than those of private datasets. In light of this insight, contemporary attack methodologies have largely abandoned the use of these loss components.

% 
% 
% --------------------------------------------------------

% --------------------------------------------------------
% BY YHY

\textbf{Black-Box Attacks.}
% 
% Contrasting with white-box attacks, black-box attacks do not have access to any information inside the target model, precluding the direct optimization of hidden vectors via gradient methods.0.1
In contrast to white-box settings, this type of attack has no access to any information from the target model except for the output content. Consequently, the gradient information is unavailable for performing the back-propagation operation.
% ADDED BY QYX

% --------------------------------------------------------
% Convert to white-box.
% --------------------------------------------------------
% \todo{check title}
% \textit{Convert to white-box scenarios.} 
% 
SecretGen \cite{yuan2022secretgen} proposes a straightforward solution to the problem of gradient inaccessibility. The attacker first samples numerous latent vectors from random noise and selects the ones that produce images predicted as the correct labels. These vectors are then optimized exclusively with the discriminator loss to improve the reconstruction quality.
% 
% LOKT \cite{nguyen2023labelonly} labels each image in the public dataset with pseudo-labels (as shown in Section \ref{sec:pseudo label guidance}) and uses these image-label pairs to train an ACGAN \cite{odena2017conditional}. The attacker then uses the ACGAN to generate a large number of image-label pairs to train multiple surrogate models. Consequently, the attacker can execute white-box attack methods by replacing the target model with the surrogate model.
% LOKT \cite{nguyen2023labelonly} labels each image generated by the generator with pseudo-labels (as shown in Section \ref{sec:pseudo label guidance}) and uses these image-label pairs to train an target model-assisted ACGAN, \cite{odena2017conditional}.
% --------------------------------------------------------
% Reinforcement Learning
% --------------------------------------------------------
% 
RLBMI \cite{han2023reinforcement} introduces a reinforcement learning-based model inversion attack that leverages confidence scores to provide rewards and optimizes the agent model with the Soft Actor-Critic (SAC) \cite{haarnoja2018soft} algorithm.
This algorithm employs an actor-critic architecture where both states and actions are represented as the latent vectors of the generator, and the next state is computed as a weighted combination of the current state and the action. In this actor-critic framework, the actor network generates an action based on the current state, while the critic network estimates the corresponding reward. To achieve the goal of MI, RLBMI designs the reward with three components: state reward, action reward, and distinguishability reward. The state and action rewards are defined as the confidence scores of the target label with respect to the next state and the action respectively, and distinguishability reward measures the disparity between the confidence score of the target class and the highest confidence score of other classes. By updating the actor-critic networks, the actor network learns to produce actions that are able to yield higher rewards, which contributes to significantly enhancing the effectiveness of MI attacks. 
% During the attack process, the attacker incrementally generates actions, resulting in a large set of states. The states with the highest confidence scores for the target label are then selected to reconstruct the images.
% 添加优化对象和原理的关系

% --------------------------------------------------------
% gradient simulation
% --------------------------------------------------------
% \textit{Gradient simulation.}
% 
% 
\textit{Label-only scenarios.} BREPMI \cite{kahla2022label} proposes a boundary-repelling strategy to tackle label-only scenarios where only the predicted hard labels are available. Their intuition is that the farther image from the class decision boundary tends to be more representative of the class. Consequently, BREPMI utilizes zero-order optimization to urge the latent vector to gradually move away from the decision boundary. In light that images that are not predicted as the target class represent incorrect optimization directions, the attacker estimates the gradients $\hat{g}_\mathbf{z}$ of latent vectors $\mathbf{z}$ opposite to these misclassified images by randomly sampling $N$ unit vectors $\mathbf{u}$ and calculating the loss with sampled points on a sphere at radius $r$:
\begin{equation}
\label{eq:brep}
\hat{g}_\mathbf{z} = \frac{1}{N} \sum_{i=1}^N\Phi_c(\mathbf{z}+r\mathbf{u_i})\mathbf{u_i},
\end{equation}
\noindent where $\Phi_c(\cdot)$ denotes a function that equals zero if the generated image is classified as the target class $c$, otherwise $-1$. The attacker calculates the gradient as an average over those misclassified points and optimizes the latent vector $\mathbf{z}$ using gradient ascent to move them in the direction opposite to this unintended average. 
During the optimization, BREPMI increases the value of $r$ when all sampled points are classified into the target label to better estimate the gradients. A following study BMI-S \cite{xu2023sparse} adopts a similar pipeline with BREPMI \cite{kahla2022label} but assumes the soft labels are available. The research replaces the $\Phi_c(\cdot)$ with CE loss in Eq. \ref{eq:brep} and fixes the value of $r$ during the optimization. 
LOKT \cite{nguyen2023labelonly} also considers label-only scenarios and introduces a Target model-assisted ACGAN (T-ACGAN). During the T-ACGAN training, LOKT employs the target model to tag public auxiliary images and synthesized images generated by generator $G$ with pseudo labels as class supervision to improve the training of ACGAN's discriminator ${D}(\cdot)$ and classification head ${C}(\cdot)$ (where ${D}(\cdot)$ is used to penalize the image realism and ${C}\circ {D}(\cdot)$ is an auxiliary classifier to provide identity loss for better GAN learning). 
With the training of ACGAN finished, the attacker could directly use $C\circ D(\cdot)$ as the surrogate model or additionally train multiple surrogate models with sufficient image-label pairs generated by $G(\cdot)$ and labeled by the victim model $f_{\theta}(\cdot)$. 
Finally, the attacker executes SOTA white-box MI methods on the target model replaced by the obtained surrogate models, which achieves satisfying attack results.
% \cite{xu2023sparse} proposes to use the Natural Evolution Strategy (NES) to estimate gradients of the loss function. Given a latent vector, the attacker sample some near vectors and compute the losses of them to estimate the direction of gradient.

% --------------------------------------------------------
% gradient free
% --------------------------------------------------------

\textit{Gradient-free optimizer}. 
Orthogonal to the above methods, Mirror \cite{an2022mirror} and C2FMI \cite{ye2023c2fmi} firstly explore gradient-free optimization techniques, which utilize genetic algorithms instead of gradients for optimization. Specifically, the population of $k$ candidates in the genetic algorithm represents the latent vectors of GANs, and the fitness score is defined as the confidence score for the target label.
% initial selection
To obtain an adequate initial population, Mirror \cite{an2022mirror} employs a simple initial selection strategy while C2FMI \cite{ye2023c2fmi} proposes a more complicated coarse optimization method as detailed in Section \ref{para:preprocess}.
% Genetic 
In each iteration of the genetic algorithm, a subset of candidates is selected as parents and undergoes crossover to generate new candidates. Additionally, some candidates are mutated either by adding random noise (Mirror \cite{an2022mirror}) or by crossing over with the candidate that has the highest score (C2FMI \cite{ye2023c2fmi}) to create more high-quality candidates. The top $k$ candidates with the highest scores are retained for the next iteration.
This process repeats over multiple iterations and continually refines the candidates toward achieving higher confidence scores for the target label. Finally, results are generated from the candidates with the highest scores via the GAN model.

% --------------------------------------------------------
%  learning-based
% --------------------------------------------------------
% 
P2I-MI \cite{liu2024prediction} claims that optimization-based methods suffer from high cost and low efficiency. 
Thus, P2I-MI proposes an optimization-free training-based strategy to overcome this issue. The attacker trains a prediction alignment encoder (PAE) to map the output prediction to the style vector $\mathbf{w}$ in the $\mathcal{W}$ space of StyleGAN2-Ada. 
During the attack stage, the research proposes an ensemble scheme that samples a batch of public images and gets their corresponding predicted confidences to be mapped into style vectors with the learned PAE. 
The final style vector $\mathbf{w}$ is calculated as a weighted average of these style vectors, which serves as the condition input for StyleGAN2-Ada to generate the recovered image.
In addition, the prevalent denoising diffusion models are also introduced to MI attacks. DiffusionMI \cite{liu2023diffusion} uses an auxiliary dataset labeled with pseudo labels to train a class-conditional diffusion model, which is directly used to iteratively generate plenty of target images without any further optimization during the attack phase.

Recent study SMILE \cite{smile} argues that existing black-box MIAs suffer from two critical limitations: prohibitive query overhead and the extremely long-tailed distribution of samples generated by the generator. To address these limitations, SMILE proposes a novel black-box MIA method with a two-stage pipeline. In the first stage, a long-tailed surrogate model training scheme is adopted: the first n-1 layers of the pre-trained model are frozen to serve as the shared backbone, with only 3 ensemble classifiers being trained. The surrogate model is optimized by integrating a distillation loss, a diversity loss, and a top-k reweighting strategy. In the second stage, an automatic black-box optimization algorithm selector NGOpt is leveraged to identify the optimal gradient-free black-box optimization algorithm and the MetaModelOnePlusOne \cite{beyer2002evolution} algorithm is selected finally. Meanwhile, a P-space clipping strategy is applied during the optimization process to expand the search space, thereby significantly reducing the overall query overhead.

Overall, the critical challenge limiting the real-world application of current MI attacks in black-box settings lies in the increased complexity of the optimization process resulting from the absence of the gradient information.

% !!!!!!!!!!!!!!!!!!!!!!!!!!!!!!!!!!!!

% diffusion
% TODO: in      troduce `prevalent`
% However, GAN-based methods in black-box attacks usually fail to accurately determine the color characteristics of the target, such as skin tone or pupil color. Avoiding this, \cite{liu2023diffusion} emphasised the importance of generators, using an advanced diffusion model instead of GANs.
% MODIFIED BY QYX
% In recent years, diffusion models have demonstrated superior performance in image generation, especially in determining the color characteristics of the target, such as skin tone or pupil color.
% % 
% \cite{liu2023diffusion} first investigate the impact of a generator on MI attacks within a label-only black-box setting and use a diffusion model to substitute the usual GAN, achieving improvements on attack performance.
%
% --------------------------------------------------------

% --------------------------------------------------------
% BY YHY
\begin{figure*}[t]
\centerline{\includegraphics[width=\linewidth]{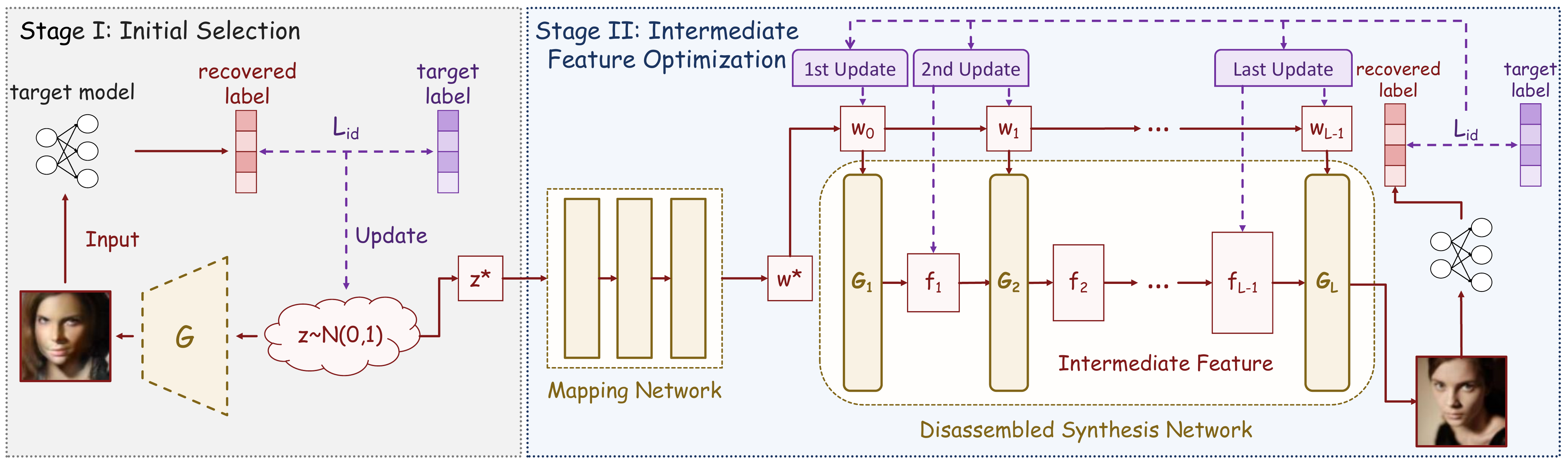}}
\caption{Illustration of the pipeline of IF-GMI attack procedure.}
\label{pipeline_of_ifgmi}
\end{figure*}
\textbf{Search Space.}
% 初始化要么是高斯分布，要么是stylegan的w-space，与model绑定，写在这一章节
Most previous studies focus on finding the optimal vectors in the GAN's latent space $\mathcal{Z}$.
However, \cite{an2022mirror} shows that as the resolution increases, optimization becomes under-constrained due to the large space sparsity. Feature entanglement in $\mathcal{Z}$ space is another challenge that further impedes the optimization process since images sharing similar features may not correspond to analogous latent vectors.
% style gan   w space
StyleGANs \cite{karras2019style} transform the latent space into the well-disentangled $\mathcal{W}$ space through a mapping network $G_{mapping}: \mathcal{Z}\rightarrow \mathcal{W}$.  Therefore, searching the $\mathcal{W}$ space alleviates the above issues, which has been adopted by a series of following studies \cite{struppek2022plug,ye2023c2fmi}. 
% \cite{an2022mirror} search them in $\mathcal{P}$ space, defined as the space before the final LeakyReLU function in the mapping network to simulate Gaussian distributions while remaining disentangled.
% 
% !!!!!!!!!!!!!!!!!!!!!!!!!!!!!!!!!!!!
% Taking into account the challenge of modeling $\mathcal{W}$ distribution, \cite{an2022mirror} propose the concept of $\mathcal{P}$ space, which represents the feature space located before the final LeakyReLU function in the mapping network of StyleGAN. This design can preserve the capabilities of controlling styles from the $W$ space while ensuring the controls adhere to Gaussian distributions.
% 

% --------------------------------------------------------

Previous methods perform the attack within a fixed search space. In contrast, IF-GMI \cite{qiu2024closer} innovatively operates optimization across multiple feature spaces to fully exploit the hidden information in GAN's intermediate layers, as shown in Fig. \ref{pipeline_of_ifgmi}. To perform the intermediate layer optimization, a StyleGAN2-Ada $G(\cdot)$ is divided into $L+1$ blocks as follows:
\begin{equation}
G(\cdot) = G_{L+1}\circ G_L\circ \dots \circ G_2 \circ G_1(\cdot).
\end{equation}
% $\mathbf{f}_i$ denotes the  output intermediate feature of $G_i$:
% \begin{equation}
% \mathbf{f}_i = G_i(\mathbf{f}_{i-1}, \mathbf{w}_i).
% \end{equation}
% Here, $\mathbf{f}_0$ is fixed as a constant input.

During the attack process, the attacker first optimizes the inputs of $G_1(\cdot)$ and then continues the optimization intermediate features layer by layer until reaching $G_L(\cdot)$. Moreover, IF-GMI \cite{qiu2024closer} employs a $l_1$ ball to limit the optimization space of each feature during the intermediate search. The latest advance PPDG \cite{ppdg} has taken a further step by directly fine-tuning the GAN prior during the optimization process. Specifically, pseudo-private data (\textit{i.e.}, surrogate samples containing the characteristic features of private training data) is first generated using the target model and the GAN prior. Next, a top-k strategy is adopted to select high-quality pseudo-private data based on the model’s prediction confidence. Finally, the screened pseudo-private data is used to fine-tune the GAN prior, thereby narrowing the gap between the GAN prior distribution and the private data distribution, thereby fundamentally boosting the performance of MIAs.

Besides, some methods propose to constrain the search space to avoid unrealistic image generation. 
% 
% Taking into account the difficulty of modeling $\mathcal{W}$ distribution, \cite{an2022mirror} propose the concept of $\mathcal{P}$ space, the feature space located before the final LeakyReLU function in the mapping network, to constraint the final $\mathbf{w}$ vectors within the target distribution. 
Mirror \cite{an2022mirror} and SMILE \cite{smile} suggest a $\mathcal{P}$-space clipping strategy to constrain the final $\mathbf{w}$ vectors within the target distribution. The $\mathcal{P}$ space is defined as the feature space located before the final LeakyReLU function in the mapping network.
Numerous vectors are first sampled from $\mathcal{Z}$-space to compute the mean $\mu$ and variance $\sigma$ of the activation values in $\mathcal{P}$-space. 
Given an optimized $\mathbf{w}$ in $\mathcal{W}$-space, attackers first obtain $\mathbf{p}$ by projecting $\mathbf{w}$ to $\mathcal{P}$-space. Then $\mathbf{p}$ is clipped within the range $[\mu - \sigma, \mu + \sigma ]$ and projected back to the $\mathcal{W}$-space. This operation preserves the style control capabilities of the $\mathcal{W}$ space while ensuring these controls are within the desired distribution. 

\subsubsection{Additional Generic Techniques}
% --------------------------------------------------------
% BY YHY
% Here we elaborate on the innovative techniques employed by contemporary attack methods. These techniques notably enhance the efficacy of attacks.
 Various innovative techniques have been explored and incorporated into numerous studies as auxiliary. Next, we supplement a detailed review of several generic mechanisms that are leveraged in multiple attack methods.
% --------------------------------------------------------

% --------------------------------------------------------
% BY YHY
% \textbf{Auxiliary Dataset Selection.}
% Dataset selection
% \cite{yuan2023pseudo,liu2023diffusion} employ an auxiliary dataset selection strategy to guide the training of the attack model. This strategy fundamentally involves utilizing a publicly available dataset as input. and assigning labels based on the target model's output, thereby extracting prior features characteristic of the private data. 

% MODIFIED BY QYX
\textbf{Pseudo Label Guidance.}
\label{sec:pseudo label guidance}
As introduced before, pseudo labels have been utilized to guide the training of the image generator \cite{yuan2023pseudo,nguyen2023labelonly,liu2023diffusion}, the discriminator \cite{chen2021knowledge,yuan2023pseudo,nguyen2023labelonly}, and surrogate models \cite{nguyen2023labelonly,nguyen2023re}. Specifically, the victim image classifier is leveraged to reclassify unlabeled images (\textit{e.g.}, the images generated from the GAN model or public images) with labels from the private dataset. Since the victim model is trained on the private dataset, the provided pseudo labels contribute to exploiting additional privacy information within the private training data, thus enhancing the capabilities of the generator or discriminator trained on these auxiliary labeled images.

% --------------------------------------------------------

% --------------------------------------------------------
% BY YHY
% % Augmentation
% Some augmentation methods are employed to mitigate the risk of overfitting the target model. 
% % 
% % Data Augment: PPA, diffusion
% \cite{struppek2022plug,liu2023diffusion} apply random transforms to generated images before input into the target model to enhance the robustness of the reconstructed images.
% % 
% % Model Augment: lomma 
% Furthermore, \cite{nguyen2023re} use model distillation to train auxiliary augmentation models, synergistically guiding the inversion process to reduce overfitting.

% MODIFIED BY QYX
% Augmentation
\textbf{Augmentation.}
Many studies have integrated various augmentation techniques into the MI workflow to improve the attack effects.
% These techniques can be categorized into data augmentation and model augmentation.  
% Data Augment: PPA, diffusion，SecretGen, PLGMI
% 举几个例子即可
% !!!!!!!!!!!!!!!!!!!!!!!!!!!!!!!!!!!!!!!!!!!!!!!!!!!!
\cite{yuan2022secretgen} employ sequential cutout as a data augmentation for images to improve the initial latent vector selection. \cite{struppek2022plug,liu2023diffusion} process the reconstructed images with image transformation and select results with higher confidence from the output of the target model. Furthermore, \cite{yuan2023pseudo} performs random augmentations on the generated images before feeding them to the target model to provide more stable convergence to realistic images during the GAN training. 
% Model Augment: lomma 
In addition to traditional data augmentations, \cite{nguyen2023re} presents a novel approach called model augmentation. This involves training several auxiliary models from the target model using model distillation techniques. During the MI process, the adversary utilizes an ensemble of the original target model and the trained auxiliary models to calculate the loss function. By augmenting with surrogate models, this strategy mitigates the issue of overfitting to the target model and encourages the reconstructed images to capture more distinctive features of the private data.
% --------------------------------------------------------
% !!!!!!!!!!!!!!!!!!!!!!!!!!!!!!!!!!!!!!!!!!!!!!!!!!!!

% result selection
% 这段是不是润色了PPA的原文？
% \cite{struppek2022plug} highlight that neural networks are usually not robust and overconfident in their predictions. Even the incorporation of an image prior does not effectively mitigate the production of misleading samples.
% % 
% Consequently, \cite{struppek2022plug,liu2023diffusion} implement transformations on generated images and select the samples with the highest scores to improve attack accuracy.

% MODIFIED BY QYX
\textbf{Results Selection.}
\cite{struppek2022plug} notes that DNNs often exhibit overconfidence in their predictions, leading to low transferability in attack results. More concretely, while some reconstructed images are labeled with high confidence by the target model, they receive low scores when evaluated by another model. To overcome this challenge, \cite{struppek2022plug,liu2023diffusion} apply data augmentations to the generated images before classifying them with the target classifier. By selecting results with the highest confidence score after the augmentation, this approach achieves recovery with enhanced attack accuracy and superior transferability.
% diffusion也进行了post selection
% 一样的方法嘛？
% 差不多，只是diffusion是label only，它是选一堆transform完以后仍然被认为是目标类的图
% 那就一样是数据增强了
% --------------------------------------------------------

\subsubsection{Evaluation Metrics}
% Some common evaluation metrics for model inversion attacks are as follows.
% MODIFIED BY QYX
Assessing the degree of privacy leakage raised by MI attacks is another pivotal open issue. 
Generally, researchers evaluate the performance based on the similarity between the reconstructed images and target class images. In addition to the common distance measurement (e.g., PSNR \cite{sheikh2006statistical}, LPIPS \cite{zhang2018unreasonable} or FID \cite{heusel2017gans}), novel metrics tailored for MI evaluation are introduced and widely adopted. We summarize these metrics in detail as follows.
\begin{itemize}
    % \item \textbf{Accuracy.}  Classification accuracy of the target model.
    \item \textbf{Accuracy.} This metric is calculated as the classification accuracy of an auxiliary evaluation classifier (trained on the same private dataset as the victim model but with a different architecture) when applied to these reconstructed samples. It serves as a pivotal criterion for how well the generated samples resemble the target class. The higher accuracy the synthesized samples achieve on the evaluation model, the more privacy in the training dataset is exposed \cite{zhang2020secret}.
    % \item \textbf{Feature Distance (Feat Dist).} The
    % $\mathcal{L}_{2}$ feature distance between the reconstructed image and the centroid of the target class \cite{zhang2020secret}.
    % \item \textbf{K-Nearest Neighbor Distance (KNN Dist).} The shortest feature distance between the attacked image and all training images with the same class.
    \item \textbf{Feature Distance.} The feature distance is defined as the $\mathcal{L}_2$ distance between the feature vector of the reconstructed image and the feature centroid of the private samples specific to the target class. Note that the feature vector is the output from the penultimate layer of the evaluation classifier. 
    \item \textbf{$K$-nearest neighbor (KNN) distance.} The KNN distance represents the minimum distance from the reconstructed image to the images within the target class \cite{zhang2020secret}. In practical terms, the $K$ is usually $1$, i.e., the nearest image within the target class to the synthesized sample is first identified, and their feature distance is computed as the KNN distance metric.
    \item \textbf{Sample Diversity.} \cite{wang2021variational} proposes this metric that focuses on the intra-class diversity of the reconstructed images. It first computes the improved Precision-Recall \cite{pr} and Density-Coverage \cite{dc} scores regarding the classification results of the reconstructed images on a given evaluation model. The average of the two quantities is the \textit{sample diversity}, whose higher values indicate greater intra-class diversity of the synthesized samples. 
    % \item \textbf{FID.} Fr\'echet inception distance\cite{heusel2017gans} is a commonly used metric to evaluate the quality and diversity of images synthesized by generative models. The typical MI setting adopts the Inception-v3 pre-trained on ImageNet to calculate the distance of penultimate features extracted from the real and the recovered images. A lower FID score indicates higher realism and diversity \cite{wang2021variational}.
    % \item \textbf{Peak Signal-to-Noise Ratio (PSNR).} The ratio of an image’s maximum squared pixel fluctuation over the mean squared error between the private image and the reconstructed image. Notably, PSNR attaches great importance to pixel-wise distance and thus sometimes can not accurately measure the disclosed privacy in MI, which actually emphasizes more on semantic-level similarity.
    % \item \textbf{Learned Perceptual Image Patch Similarity (LPIPS).} The human perceptual similarity \cite{zhang2018unreasonable} between the reconstructed and private images, quantified by the discrepancy of their feature representations from multiple layers of a given pre-trained network.
\end{itemize}

\subsection{Attacks against Image Generation}
While most existing algorithms are designed for image classification tasks, recent research also explores the potential for inverting the generative model to reconstruct their private training data. 

Previous research \cite{xu2019ganobfuscator} demonstrates that GAN models \cite{goodfellow2016nips, arjovsky2017wasserstein} are susceptible to model inversion attacks. By repeatedly sampling images from the generator's output distribution, it's highly likely to generate raw samples from the GAN's training set. Compared with GAN models, the denoising diffusion model \cite{croitoru2023diffusion} is able to produce images of higher quality and realism. However, \cite{carlini2023extracting} indicates that they are also more susceptible to being inverted to reconstruct private data, which leaks nearly twice as much training data as GAN models. 

Specifically, they propose a generation and filtering mechanism and target a range of prevalent text-conditioned diffusion models, such as Stable Diffusion \cite{rombach2022high} and Imagen \cite{saharia2022photorealistic}. 
The extraction process consists of two steps: (1) repeatedly query the diffusion model with the selected text prompt to obtain a large number of images, and (2) construct a graph using these generations by establishing an edge between the generated images if they are considered to be sufficiently similar based on a pre-defined distance metric. Ultimately, the largest clique in the graph, which comprises at least 10 nodes, is identified as a memorized image.
In essence, this serious data leakage issue is a production of diffusion's significant data \textit{memorization} phenomenon, which allows them to regenerate individual raw training samples under certain guidance. 

A subsequent study \cite{gu2023memorization} further examines the influencing factors on data memorization and confirms that the conditioning input of diffusion models greatly increases the data memorization and thus renders them more vulnerable to MI attacks.
Building upon this line, SIDE \cite{chen2024extracting} delves into the more challenging attacks on unconditional diffusion models.
To establish a universal theoretical analysis, SIDE first proposes a novel and effective memorization metric to provide theoretical support for enhanced attacks, which also corroborates that the input conditions can enhance the data memorization in diffusion models and further result in more information leakage. 
To address the difficulty of the missing input conditions in unconditional diffusion models, they propose an effective solution that additionally trains a time-dependent classifier and uses its classification information as surrogate conditions to guide the denoising process for better private data generation.

Existing attacks have been able to effectively induce memory leakage, but their practical applicability is still confined to untargeted data leakage, failing to achieve targeted privacy extraction.

% by kjw
\section{Data Modalities}
While existing MI attacks on DNNs mainly focus on image recovery, recent progress has extended the attacks to text and graph data, which also garners significant attention and concerns.

\label{sec:mi_more_modal}
\subsection{Text Data}
\label{sec:mi_text}
% In the field of NLP, models are trained to transform sentences into tokens for downstream tasks. Large 
DNN models that process natural-language text also suffer from MI attacks. 
% Due to the language models' \textit{unintended memorization} of training data \cite{carlini2019secret}, an adversary can invert the victim models and incur serious privacy leakage. 
This has been largely amplified by the rise of LLMs since these models are widely used and trained on massive text corpora that often contain considerable privacy-sensitive information. Formally, we divide these various attacks into \textit{Embedding Optimization}, \textit{Inverse Mapping}, \textit{Token Search}, and \textit{Malicious Prompt Design}.
\begin{figure}[tbp]
\centerline{\includegraphics[width=\columnwidth]{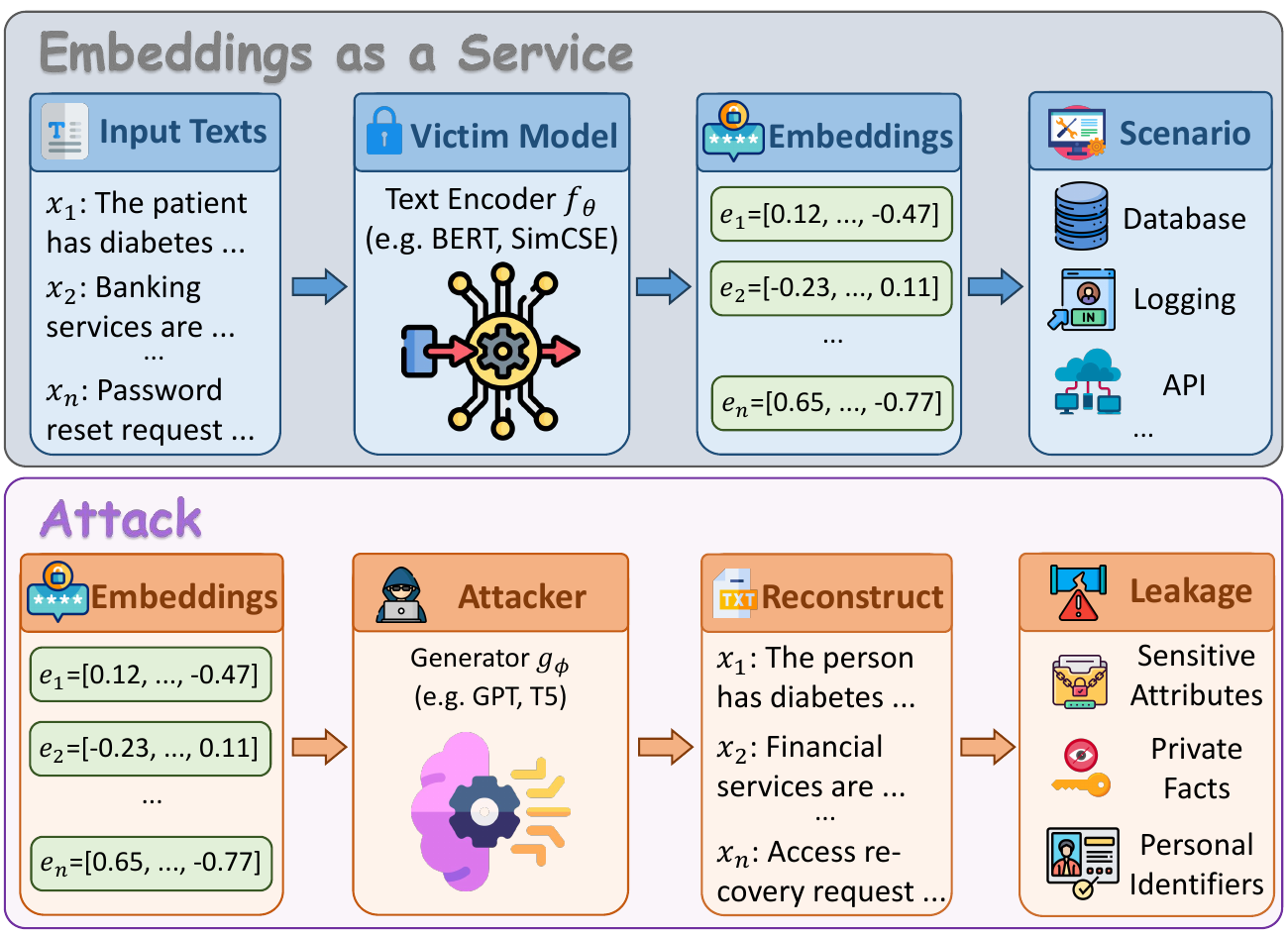}}
\caption{
Overview of embedding inversion attacks in Embeddings-as-a-Service. Leaked sentence embeddings can be exploited by an attacker to reconstruct the original texts, leading to exposure of sensitive attributes, private facts, and personal identifiers.
}
\label{fig:embedding_inversion}
\end{figure}

\textbf{Embedding Optimization.} These attacks mainly take advantage of the full access to the victim model by formulating the reconstruction as an optimization process that utilizes the back-propagated gradients. \cite{parikh2022canary} attacks the LSTM model $f_\theta(\cdot)$ by conducting a discrete optimization, where each token $x_i$ of the sentence $s$ is represented as a logit vector $z_i$ to get the embedding $E_{s}$ of the dummy sentence $s$. Given the target label $y$, the vectors $z_1, \ldots, z_n$ are repeatedly optimized with the gradients computed from the cross-entropy loss $\mathcal{L}(f_\theta(E_{s}), y)$. Subsequently, \cite{zhang2023ethicist} adopts the paradigm of prompt learning that freezes the parameter of the victim GPT-Neo and tunes the pre-sentence soft prompt embedding $S$ by forcing the model to output the target suffix $b$ for a given prefix $a$. The learned prompt $S$ elicits the \textit{memorization} of the target language model and induces it to respond with private training data. Additionally, it introduces a smoothing regularizer to make the loss distribution of the suffix sequence smoother.
Despite their effectiveness, these methods rely on strong white-box access and costly iterative optimization, which limits their practicality in real-world LLM services where only black-box APIs are usually available.

\textbf{Inverse Mapping.} The rapid development of \textit{general-purpose language models} (e.g., BERT \cite{devlin-etal-2019-bert}) makes them widely used feature extractors to encode sentences into dense embeddings.
In the privacy-sensitive learning with service providers, users may expect that it is a safe way to submit text features encoded using the language model. However, research has demonstrated that the submitted sentence embeddings can also incur the leakage of users' private data, as illustrated in Figure \ref{fig:embedding_inversion}.
 To be specific, \cite{pan2020privacy} constructs two types of attacks, namely pattern reconstruction attacks and keywords inference attacks.
It collects an external corpus $\mathcal{D}_{ext}=\{s_i\}_{i=1}^{N}$ to extract sensitive information $\{T(s_i)\}_{i=1}^{N}$ as labels and query the language model to obtain the sentence embedding $\{f_{\theta}(s_i)\}_{i=1}^{N}$. An inversion model consisting of a linear SVM and a 3-layer MLP with 80 hidden units is trained on the $(f_{\theta}(s_i), T(s_i))$ pairs to invert the sentence embedding to the sensitive information for attack purpose. However, \cite{song2020information} claims that the direct inversion from output to input with only one step is a highly non-convex optimization issue and limits the recovery accuracy, especially as the victim model grows deeper. Therefore, it divides the inversion into two stages: (1) learn a mapping function $\mathcal{M}(\cdot)$ to map the higher layer embedding $f_\theta(s_i)$ to the shallow-layer features (rather than the input words), and (2) infer the private words by matching their corresponding embedding with the obtained shallow layer representations $\mathcal{M}(f_\theta(s_i))$. 
Nevertheless, one common drawback of these two attacks is they only reconstruct orderless sets of words that lack meaningful semantic information. To address this problem, \cite{li2023sentence} reformulates the attack as a generation task, which trains a generative attacker model, \textit{i.e.,} GPT-2, to iteratively map the sentence embedding $f_{\theta}(s)$ with previous contexts $x_1, \ldots, x_{i-1}$ into the private word $x_i$.
Subsequently, \cite{chen2024text} further studies embedding inversion beyond monolingual English settings by investigating black-box attacks against multilingual sentence embeddings. They show that generative inversion models can recover text content across multiple languages, even without explicit prior knowledge of the target language.
However, these studies typically assume unlimited query access to the victim model. \cite{huang2024transferable} investigates a more realistic query-free black-box setting, where the attacker cannot access or query the victim encoder and only observes a small set of leaked text-embedding pairs. Their method learns a surrogate encoder to mimic the victim model, and then trains an embedding-to-text attacker on large-scale surrogate-generated embeddings. The resulting attacker exhibits strong transferability, which can generalize to embeddings produced by unseen victim encoders.
From a practical perspective, inverse-mapping attacks highlight a fundamental trade-off between attack realism and reconstruction quality. Query-based methods can obtain stronger reconstruction performance by repeatedly accessing the victim encoder, but this assumption may be unrealistic for deployed embedding services with strict rate limits or closed APIs. In contrast, query-free transferable attacks better match leakage scenarios in real-world systems, yet their effectiveness depends on the availability of representative leaked pairs and the quality of the surrogate encoder. 

The capability of LLMs can be greatly enhanced through well-crafted prompts, which has promoted the continued design of high-quality instructional prompts by service providers. These prompts have also become a valuable asset within their intellectual property rights (IPR). However, malicious users can cause a serious infringement of their IPR by recovering the invisible prompt via the model output. \cite{morris2023language} trains an inversion model $\mathcal{M}(\cdot)$ with an encoder-decoder backbone, to map the model's next-token probabilities $\mathbf{v}$ back to initial tokens $x_{1:N}$. During the recovery, the output probability vector of the next token is first unrolled into a sequence of pseudo-embeddings, which are then encoded to provide conditions for the decoder to recover the target prompt. 
Since LLMs are usually available via API, the output confidence vectors may be inaccessible. Given the typical LLM setting that provides the argmax of the confidences and allows users to add a logit bias to adjust the output distribution, the probability of each token can be estimated based on its difference from the most probable word. This is calculated as the smallest logit bias required to make that word most likely. 
PILS~\cite{nazir2026better} exploits the redundancy of language models' next-token distributions by replacing full vocabulary-sized logprob vectors with compact representations over selected tokens. It further incorporates multi-step logprob sequences into an encoder-decoder inverter, thereby improving hidden prompt recovery while reducing logprob access costs.
Instead of estimating the confidence distribution, Output2prompt \cite{zhang2024extracting} directly uses LLM output sequences as model input to train the inversion model $\mathcal{M}(\cdot)$. Besides, a sparse encoder is employed to reduce the memory costs of training $\mathcal{M}(\cdot)$.
These methods provide a systematic route to recover hidden prompts from model outputs, but their deployment practicality depends on the availability of informative output signals and the high cost of training an additional inversion model, which may fail to satisfy in black-box scenarios.

% \textbf{Black-Box Attacks}
% \textbf{Black-Box Attacks.} Given that only the model output is accessible, the gradient back-propagation becomes impractical. Instead, studies choose \textit{search} or \textit{query-based} strategies.

% Since gradient back-propagation becomes impractical under black-box scenarios, researchers have developed novel algorithms, namely the \textit{Token Search} and \textit{Malicious Prompt Design}, to extract private sentences.

% This section presents the 
% search-based methods recover training data in a token-level generate-then-rank manner.
\textbf{Token Search.} In the black-box scenario, only model outputs are accessible, and the gradient back-propagation becomes impractical. Researchers choose optimization-free \textit{search-based} strategies to handle this challenge. Given the sentence's prefix tokens $a=x_1, x_2, \ldots, x_i$, attackers attempt to search the most probable pathway for generating the remaining portion of the sentence, i.e., $b=x_{i+1}, x_{i+2}, \ldots, x_n$.
% multiple candidates for the next token $x_{i+1}$. They rank and select based on specific criteria. 
Based on the search objective for the subsequent tokens, these methods are further divided into three parts as follows.

%\textbf{Perplexity Objective Search.} 
\textbf{Perplexity-metric Search.} Carlini et al. \cite{carlini2019secret} use \textit{perplexity} to measure the probability of data sequences. Specifically, given a sequence $s=x_1\ldots x_n$ and the target generative sequence model $f_\theta$ (i.e., LSTM, qRNN), the perplexity $\mathrm{P}_{\theta}(s)$ can be expressed as:

\begin{equation}
    \mathrm{P}_{\theta}(s) = \exp\left(-\frac{1}{n} \sum_{i=1}^n \log f_\theta \left(x_i|x_1, \ldots, x_{i-1}\right)\right),
\end{equation}

\noindent where $\mathrm{P}_{\theta}(s)$ quantifies the “surprise” of a given sequence, and a lower perplexity value indicates a higher probability that the language model has seen the sequence.
In this manner, the problem is transformed into finding the sequence with minimal perplexity. However, directly enumerating all possible sequences and computing their perplexity leads to an exponentially growing search space. A series of mechanisms has been proposed to improve search efficiency. 
% Taking the prefix as the start, the end of the sentence as the termination, and the negative log-likelihood assigned by the language model to the token $t$ following the sequence $a$ as the cost between two nodes, 
\cite{carlini2019secret} uses a modification of the Dijkstra algorithm, which efficiently reduces the search range by avoiding visiting unnecessary sequence paths. 
The subsequent method \cite{carlini2021extracting} adopts a greedy-based algorithm that only retains tokens with top-$k$ confidence during each iteration of generating the next token. Moreover, they provide several techniques for improved generated sentence diversity and enhanced attack accuracy.
% e.g., feeds the model like GPT-2 with a set of prefixes crawled from the Internet.
% To , they suggest several additional strategies such as calculating the zlib entropy for the generated text as an auxiliary metric.
 \cite{yu2023bag} adopts a look-ahead mechanism to improve the quality of generated tokens. Instead of only depending on the next token's probability for the top-$k$ selection, they use the posterior probability of the subsequent sequence to inform the generation of the next token $x_t$. 

%\textbf{Target Label-based Search.}
% Classifiers may also suffer from attacks in a token-by-token generation paradigm.
\textbf{Confidence-metric Search.} Considering the sentiment classification task, \cite{elmahdy2022privacy} searches for the next token that maximizes the confidence score of target class $y$ as predicted by a fine-tuned BERT classifier. To counteract the target model's bias towards its high-frequency tokens, an auxiliary regularizer is introduced to penalize the frequent occurrence of certain tokens. 

However, this method also faces the challenge of an exponentially growing search space. To tackle this, \cite{elmahdy2024deconstructing} first uses BERT with the original generation head to generate candidate tokens with top-$k$ confidence, which are further selected by maximizing the probability of label $y$ with the classification head.

\textbf{Difference-metric Search.} The pre-train and fine-tune paradigm has gained growing popularity, where models are first trained on a large, public dataset and then fine-tuned on a small, private dataset. 
Accordingly, several attacks appear to invert the fine-tuning dataset by exploiting the difference between the pre-trained model $f_\theta$ and fine-tuned model $f_{\theta^{\prime}}$. Considering the token sequence $s={x_1 \ldots x_n}$, \cite{zanella2020analyzing} defines the difference score $DS_{f_\theta}^{f_{\theta^{\prime}}}=\sum_{i=1}^{n} {f_{\theta^\prime}}(x_i|x_{<i}) - {f_\theta}(x_i|x_{<i})$, which measures the difference in the output probabilities between $f_\theta$ and $f_{\theta^{\prime}}$. Intuitively, a larger difference score value indicates a higher probability that the fine-tuned model $f_{\theta^{\prime}}$ has seen sequence $s$ while the original $f_\theta$ does not, i.e., $s$ belongs to the fine-tuning dataset. By performing a beam search, they select the next tokens with the highest differential scores and achieve a remarkable attack success rate. 
% This token selection operation is repeated several times to obtain multiple reconstructed sentences. 
Subsequently, \cite{panchendrarajan2021dataset} defines a novel metric $score_{r}$, which additionally considers the distribution-level difference between the target sentence and the current set of reconstructed sentences to more accurately quantify the difference. 
They first query the fine-tuned GPT-2 model $f_{\theta^\prime}$ with an empty prompt. Then a complete sentence $S_f$ is generated by iteratively querying $f_{\theta^\prime}$ until it outputs $EOS$ token. After iteratively querying $f_\theta$ with the first half of $S_f$ as the initial prompt, they obtain $S_g[l/2:]$ generated by the initial pre-trained model. As the larger difference between $S_g[l/2:]$ and $S_f[l/2:]$ indicates the greater possibility that $S_f$ comes from the fine-tuning dataset, the attacker calculates the $score_{r}$ based on $S_g[l/2:]$ and $S_f[l/2:]$ to determine whether the generated sentence belongs to the private fine-tuning dataset, thus achieving the aim of data extraction.

Overall, these search-based attacks provide a practical gradient-free route for text recovery, especially when only generated outputs or confidence scores are available. Nevertheless, they trade off accessibility for efficiency: the discrete search space grows exponentially with sequence length, and the attack performance heavily depends on the quality of the search heuristic and the availability of sufficient model queries.

\begin{figure}[tbp]
\centerline{\includegraphics[width=0.85\columnwidth]{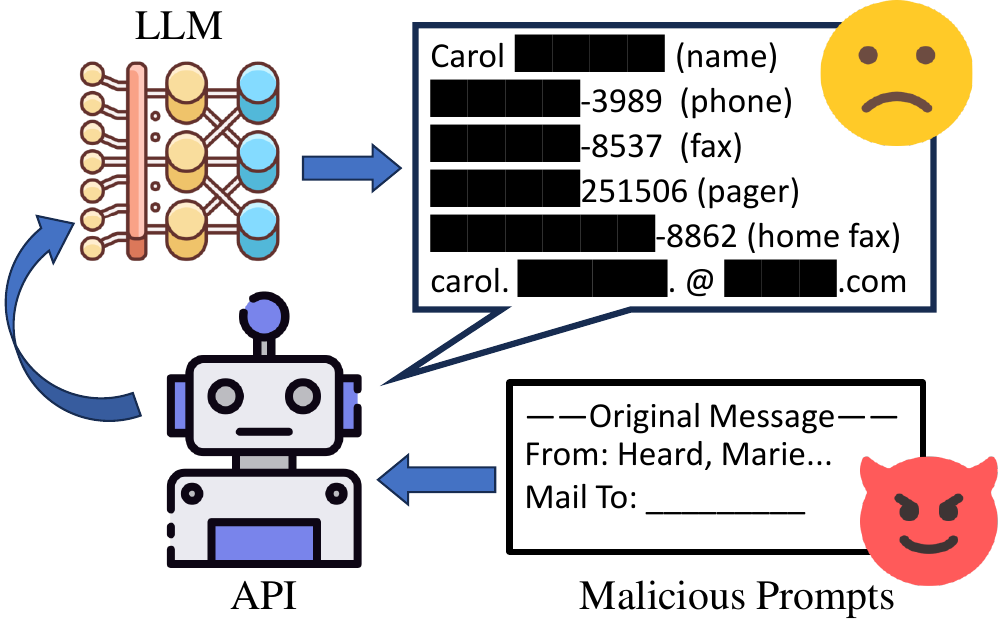}}
\caption{An example of privacy leakage on LLMs.}
\label{GPT_leak}
\end{figure}

\textbf{Malicious Prompt Design.} The conversational LLMs have demonstrated powerful capabilities and great potential. However, several studies have shown that some elaborately designed prompts can invert the model to output sensitive training data. As Figure \ref{GPT_leak} depicts, feeding malicious prompts into language models such as GPT-Neo can incur serious privacy information leakage \cite{huang2022large} in both zero-shot and few-shot settings (i.e., providing several demonstrations as prior knowledge to guide LLM's predictions).
More surprisingly, \cite{nasr2023scalable} suggests that data can also be extracted by simply asking ChatGPT to repeat a word multiple times. One possible reason is that this instruction causes the model to “escape” its aligning training and revert to its original language modeling objective.
Beyond training-data extraction, malicious prompts may also induce conversational LLMs to disclose their hidden system instructions. \cite{zhangeffective} formulates this threat as a prompt extraction attack, where the adversary queries a conversational API with carefully designed instructions that encourage the model to repeat or summarize its prior context. By aggregating the returned generations, the attacker can identify the most plausible reconstruction of the secret prompt. This attack shows that malicious prompt design can expose not only memorized training content but also confidential instructions that govern LLM-based services.

Malicious prompt design offers a lightweight and practical attack surface because it only requires standard API access. Nevertheless, its effectiveness depends heavily on the model's instruction-following capability and deployed safety filters, making it less controllable than attacks based on explicit confidence scores, gradients, or learned inversion models.

\subsection{Graph Data}
\label{sec:mi_graph}

\begin{figure}[tbp!]
\centerline{\includegraphics[width=\columnwidth]{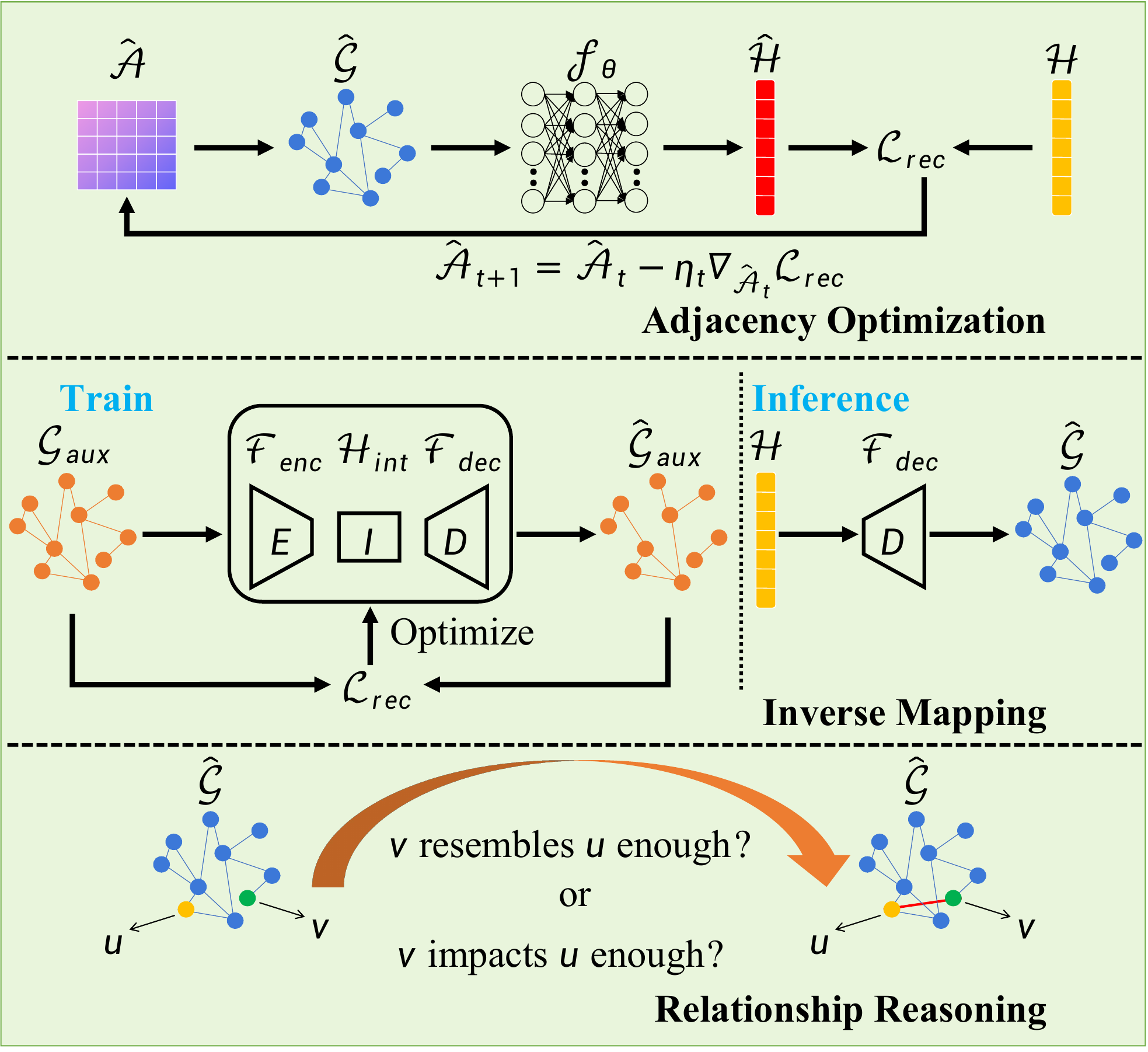}}
\caption{Different types of MI attacks on graph data.}
\label{attack_methods_for_graph}
\end{figure}

Graph data \cite{xia2021graph, rong2020deep, qiao2018data, chami2022machine, aggarwal2010graph} illustrates the linkage situation between any pair of nodes and indicates their mutual interaction. Graph Neural Networks (GNNs) \cite{wu2020comprehensive, dai2024comprehensive, scarselli2008graph, gupta2021graph, xie2022self} exhibit remarkable efficacy in processing complex graph-structured data and have achieved superior performance in non-Euclidean domains. The wide application of GNNs has brought prominent privacy leakage risks. For instance, the gradient inversion attack GraphLeak \cite{zhang2024graphleak} proposes to invert patients' electronic health record graphs from shared gradients. 
% targets medical federated learning systems, 
In the area of MI attacks, given the target GNN $f_{\theta}(\cdot)$ pre-trained on the private graph $\mathcal{G}$, the adversary aims to infer the adjacency matrix $\hat{\mathcal{A}}$, which is then converted into graph $\hat{\mathcal{G}}$ to reconstruct the private training dataset.  As shown in Figure \ref{attack_methods_for_graph}, MI attacks on this modality can be characterized into \emph{Adjacency Optimization}, \emph{Inverse Mapping}, and \emph{Relationship Reasoning}.

% leverage open access to GNNs to reconstruct the topology of the private graph data. 
% \add{These methods primarily target trained GNN models, leveraging their predictive outputs for inversion. Meanwhile, in distributed training frameworks such as federated learning, the shared model gradient updates (which aggregate information from multiple parties' private data) also become a significant channel for attackers to reconstruct the training data. For instance, the GraphLeak \cite{zhang2024graphleak} targets medical federated learning systems, proposing to invert patients' electronic health record graphs from shared gradients. This indicates that incorporating domain-specific structured priors is a promising direction for enhancing inversion efficacy (whether based on model outputs or gradients) on complex data modalities like graphs.}

\textbf{Adjacency Optimization.}
These attacks reconstruct the private graph by directly optimizing the adjacency matrix. Since back-propagated gradients are needed for updating, they generally require white-box settings. Denote the prediction vector or the embeddings output by the target GNN model as $\mathcal{H}$, i.e., $\mathcal{H} = f_{\theta}(\mathcal{G})$. An adjacency matrix $\hat{\mathcal{A}}_0$ is first initialized to obtain the dummy graph $\hat{\mathcal{G}}_0$ and the dummy output $\hat{\mathcal{H}}_0 = f_{\theta}(\hat{\mathcal{G}}_0)$. By minimizing the distance $\mathcal{L}_{rec}$ between $\hat{\mathcal{H}}_0$ and $\mathcal{H}$ with gradient descent, an attacker iteratively updates $\hat{\mathcal{A}}$ to find the optimal solution:

\begin{equation}
\hat{\mathcal{A}}_{t+1} = \hat{\mathcal{A}}_{t} - \eta_{t}\nabla_{\hat{\mathcal{A}}_{t}}\mathcal{L}_{rec},
\label{eq:graph_basic}
\end{equation}

\noindent where $t$ is the number of iterations, $\hat{\mathcal{A}}_{t}$ is the adjusted adjacency matrix after each iteration, and $\eta_{t}$ is the corresponding learning rate. Based on this, \cite{zhang2021graphmi} proposes GraphMI that uses a projected gradient module to tackle the discreteness of graph data while introducing the feature smoothness term $\mathcal{L}_s$ and the F-norm term $||\hat{\mathcal{A}}||_F$ for the regularization of feature smoothness and sparsity. RL-GraphMI \cite{zhang2023model} extends GraphMI to the more challenging hard-label black-box settings by adopting gradient estimation and reinforcement learning methods \cite{kaelbling1996reinforcement, li2017deep, wiering2012reinforcement, arulkumaran2017deep, ernst2024introduction}. From the perspective of information theory, \cite{zhou2023on} proposes to regard $f_{\theta}(\cdot)$ as a Markov chain and iteratively optimizes $\hat{\mathcal{A}}$ via a flexible chain approximation, which further achieves excellent attack results. Specifically, they adapt the original iterative optimization problem of (\ref{eq:graph_basic}) into a chain-based formulation as:

\begin{equation}
\begin{split}
\mathcal{\hat{A}}^* = \arg \max_{\mathcal{\hat{A}}}
\underbrace{\lambda_{p} \mathcal{I}(\mathcal{H}_{\mathcal{A}} ; \mathcal{H}^{i}_{\mathcal{\hat{A}}})}_{\text{propagation approximation}}
- \underbrace{\lambda_{c} \mathcal{S}(\mathcal{\hat{A}})}_{\text{complexity}} \\
+ \underbrace{\lambda_{o} \mathcal{I}(\bm{Y}_\mathcal{A} ; \bm{Y}_{\mathcal{\hat{A}}})
+ \lambda_{s} \mathcal{I}(Y ; \bm{Y}_{\mathcal{\hat{A}}})}_{\text{outputs approximation}},
\end{split}
\label{eq:graph_enhanced}
\end{equation}

\noindent where $\mathcal{I}(\cdot;\cdot)$ is the mutual information, $\mathcal{S}(\cdot)$ is the information entropy, $\mathcal{H}_{\mathcal{A}}$ denotes the embeddings of the original adjacency matrix $\mathcal{A}$, $\mathcal{H}^{i}_{\mathcal{\hat{A}}}$ denotes the representations of the dummy adjacency matrix $\hat{\mathcal{A}}$ output by the $i$-th layer ($1\leq i \leq L$) from the $L$-layer GNN target model $f_{\theta}(\cdot)$, $\bm{Y}_\mathcal{A}$ denotes the classification outputs transformed from $\mathcal{H}_{\mathcal{A}}$ by the downstream linear layer with activation, $\bm{Y}_\mathcal{\hat{A}}$ denotes the classification outputs transformed from $\mathcal{H}_{\mathcal{\hat{A}}}^{L}$ by the downstream linear layer with activation, $Y$ denotes the labels for the classification task, and $\lambda_{p},\lambda_{c},\lambda_{o},\lambda_{s}$ are the weight factors. By optimizing $\hat{\mathcal{A}}$ based on (\ref{eq:graph_enhanced}), a series of intermediate representations are also taken into consideration in addition to the final model output. Thus, the attackers can maximize the approximation of encoding and decoding processes of the target model to enhance the correlation between the ground-truth $\mathcal{A}$ and the estimated $\hat{\mathcal{A}}$, and minimize the complexity to avoid non-optimal solutions by reducing the graph density.

\textbf{Inverse Mapping.}
Different from the adjacency optimization methods that iteratively optimize $\hat{\mathcal{A}}$, these attacks attempt to construct an inverse mapping that reverses the original input-to-output mapping of the target model $f_{\theta}(\cdot)$ to recover the private graph. Most of them are intended for black-box settings where only the GNN output $\mathcal{H}$ is available. Given the difficulty of reconstruction with limited information, some additional auxiliary knowledge is introduced. Specifically, they first train an autoencoder $\mathcal{F}_{auto}(\cdot)$ on an auxiliary graph dataset $\mathcal{G}_{aux}$ by minimizing the reconstruction loss between $\hat{\mathcal{G}}_{aux}$ and $\mathcal{G}_{aux}$, so that mutual conversion between graph data and embeddings can be adequately learned. Once $\mathcal{F}_{auto}(\cdot)$ has been fully trained, they disassemble the encoder $\mathcal{F}_{enc}(\cdot)$ and decoder $\mathcal{F}_{dec}(\cdot)$ from $\mathcal{F}_{auto}(\cdot)$,
% \begin{equation}
% \mathcal{F}_{auto}(\cdot) = \mathcal{F}_{dec} \circ \mathcal{F}_{enc}(\cdot),
% \end{equation}
and utilize $\mathcal{F}_{dec}(\cdot)$ to map $\mathcal{H}$ back to $\hat{\mathcal{G}}$, i.e., $\hat{\mathcal{G}} = \mathcal{F}_{dec}(\mathcal{H})$. \cite{duddu2020quantifying} leverages such inverse mapping methods and reconstructs $\hat{\mathcal{G}}$ with high accuracy to quantify privacy leakage in graph embeddings. However, the distribution of latent features $\mathcal{H}_{int}$ within the autoencoder $\mathcal{F}_{auto}(\cdot)$ may differ from the original distribution of $\mathcal{H}$. To alleviate this issue, \cite{zhang2022inference} proposes to query $f_{\theta}(\cdot)$ with $\mathcal{G}_{aux}$ to acquire $\mathcal{H}_{aux}$, i.e., $\mathcal{H}_{aux}=f_{\theta}(\mathcal{G}_{aux})$, and fine-tune $\mathcal{F}_{dec}(\cdot)$ with the constructed $(\mathcal{H}_{aux}, \mathcal{G}_{aux})$ pair for enhanced reconstruction performance.

\textbf{Relationship Reasoning.}
These attacks are also designed for black-box scenarios. To successfully conduct the more challenging black-box attacks, they generally require some extra prior knowledge in addition to the GNN output $\mathcal{H}$, such as node attributes in the training data. Unlike the above two types of methods that constantly update $\hat{\mathcal{A}}$ or disassemble $\mathcal{F}_{dec}(\cdot)$ from $\mathcal{F}_{auto}(\cdot)$, they reconstruct $\hat{\mathcal{G}}$ by reasoning about the relationships among pairs of nodes using the introduced auxiliary knowledge. \cite{he2021stealing} considers three dimensions of auxiliary knowledge (node attributes, partial graph, or an auxiliary dataset), and evaluates $2^3=8$ settings based on whether each of the three dimensions is available. 
\cite{he2021stealing} presumes that two nodes $u$ and $v$ are linked if they share more similar attributes or predictions. They add an edge between $u$ and $v$ in the reconstructed graph when the distance of feature vectors between them is below a given threshold, and various metrics tailored for each of the $8$ auxiliary knowledge settings are designed to calculate such distances. Contrary to the above settings, \cite{wu2022LINKTELLER} assumes that the attackers only have access to node attributes in the training data. They suggest that if there is an edge between $u$ and $v$, the information of $u$ would be propagated to $v$ during training. Therefore, they assume that $u$ and $v$ are linked if changing the feature vector of $u$ can impact the prediction of $v$ to a certain extent.

Beyond algorithmic innovation, the practical deployment of graph MI attacks reveals critical trade‑offs that merit further discussion. For instance, while adjacency optimization methods such as \cite{zhang2021graphmi} achieve high-fidelity reconstruction in white‑box settings, their reliance on gradient back‑propagation limits applicability to offline or open‑source models, rendering them ineffective against real‑world online APIs; inverse mapping and relationship reasoning approaches \cite{duddu2020quantifying, he2021stealing} operate in black‑box scenarios but often depend on auxiliary graph datasets or node attributes that may be unavailable in privacy‑sensitive domains such as healthcare. One potential solution is to develop hybrid graph MI attack methods that combine the strengths of these methods, leveraging transfer learning \cite{TL} to reduce the dependence on auxiliary graph datasets and attributes.

% Moreover, as highlighted by recent work \cite{zhang2022inference}, defenses based on differential privacy or Laplacian noise effectively mitigate inversion risks but inevitably degrade the utility of graph embeddings, illustrating a fundamental tension between privacy preservation and model performance that must be carefully navigated in real‑world deployments.

% By YWB
% Delete by FFH.
% MODIFIED BY YWB: 修正Purifier处的引用无法换行的问题
% \begin{figure*}[tbp]\centerline{\includegraphics[width=\linewidth]{figure/cv_result/Defense_taxonomy_v7.pdf}}
% \caption{Taxonomy of defense strategies against MI attacks.}
% \label{fig:defense_taxonomy}
% \end{figure*}

\begin{figure*}[tbp]
    \centering
    \resizebox{\textwidth}{!}{
        \begin{forest}
            for tree={
                grow'=east,
                reversed=true,
                anchor=center,
                parent anchor=east,
                child anchor=west,
                base=left,
                font=\large,
                rectangle,
                draw=black,
                rounded corners,
                align=left,
                minimum width=4em,
                edge={darkgray, line width=1.5pt},
                edge path={
                    \noexpand\path[\forestoption{edge}]
                    (!u.parent anchor) -- ++(5pt,0) |- (.child anchor)\forestoption{edge label};
                },
                s sep=3pt,
                inner xsep=2pt,
                inner ysep=3pt,
                line width=0.8pt,
            },
            % 保持和原代码一致的层级宽度设置
            where level=0{
                text width=7em,
                rotate=90,
                anchor=center,
                parent anchor=south,
                child anchor=north,
            }{},
            where level=1{
                text width=8em,
                rotate=90,
                anchor=center,
                parent anchor=south,
                child anchor=north,
            }{},
            where level=2{text width=15em}{},
            where level=3{text width=35em}{},
            % 完全对应截图的树形结构
            [\,\,MI Defenses, fill=root-blue
                [\,\,Model Output \\ \,\,\,\,\,\,Processing, fill=model-robust-grey
                    [\,\,\,\,\,\,\,\,\,\,\,\,\,\,\,\,\,\,\,\,\,Additive Noise, fill=additive-purification-detection-orange
                        [{ Adversarial \cite{wen2021defending}, GNN-Leaks \cite{zhang2022inference}, LINKTELLER \cite{wu2022LINKTELLER}, DP \cite{ye2022one}, \\ SplitNoPeek \cite{titcombe2021practical},   Explanation Perturbation \cite{olatunji2023private}}, fill=final-grey]
                    ]
                    [\,\,\,\,\,\,\,\,\,\,\,\,\,\,\,\,\,\,\,\,\,\,\,\,Purification, fill=additive-purification-detection-orange 
                        [{Purification \cite{yang2020defending}, Link-Stealing \cite{he2021stealing}, Purifier \cite{yang2023purifier},  \\ SSD \cite{SSD}, Crafter \cite{wang2024crafter}}, fill=final-grey]
                    ]
                    [\,\,\,\,\,\,\,Detection and Filtering, fill=additive-purification-detection-orange
                        [{CtxLeak \cite{huang2022large}}, fill=final-grey]
                    ]
                ]
                [\,\,Robust Model \\ \,\,\,\,\,\,\,\,Learning, fill=model-robust-grey
                    [\,\,Secure Training Mechanism, fill=secure-robust-input-green
                        [{GraphMI \cite{zhang2021graphmi}, MID \cite{wang2021improving}, BiDO \cite{peng2022bilateral}, MC-GRA \cite{zhou2023on}, BiDO+ \cite{bidoplus},  \\ LS \cite{szegedy2016rethinking}, ResSFL \cite{li2022ressfl}, AT-AT \cite{ATAT}}, fill=final-grey]
                    ]
                    [\,\,\,\,\,\,\,\,\,\,\,\,Robust Fine-tuning, fill=secure-robust-input-green
                        [{TL \cite{TL}, GAN-based \cite{gong2023gan}, DP-BERT \cite{anil2022large}, Regularization \cite{ishihara2023training}}, fill=final-grey]
                    ]
                    [\,\,Robust Model Architecture, fill=secure-robust-input-green
                        [{RoLSS \cite{koh2024vulnerability}, LoFt \cite{calor}}, fill=final-grey]
                    ]
                    [\,\,\,\,Input Data Preprocessing, fill=secure-robust-input-green
                        [{RL-GraphMI \cite{zhang2023model}, Trap-MID \cite{trapmid}, De-duplication \cite{kandpal2022deduplicating}}, fill=final-grey]
                    ]
                ]
            ]
        \end{forest}
    }
    \caption{Taxonomy of defense strategies against MI attacks.}
    \label{fig:defense_taxonomy}
\end{figure*}
% \cite{wen2021defending,ye2022one,titcombe2021practical,zhang2022inference,wu2022LINKTELLER,olatunji2023private}
% % \cite{yang2020defending,yang2023purifier,wang2024crafter,he2021stealing,SSD}

\begin{figure}[tbp]
\centerline{\includegraphics[width=\columnwidth]{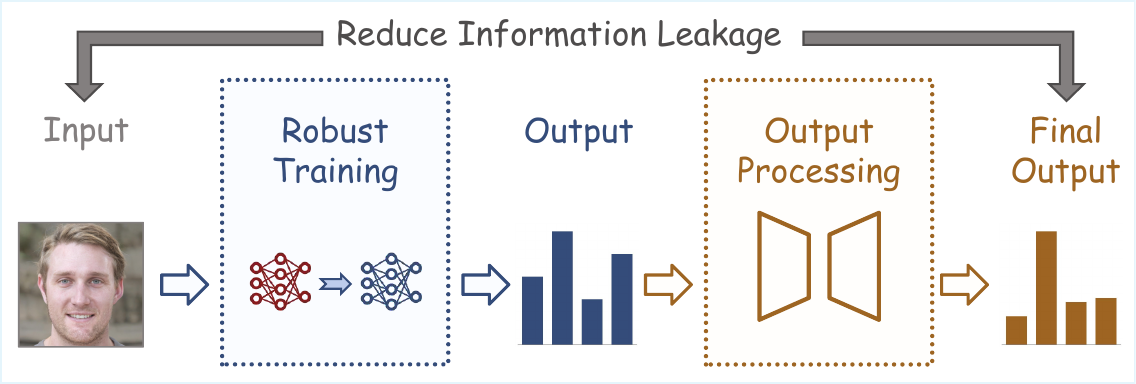}}
\caption{
General overview of the defense category.
}
\label{fig:defense objective}
\end{figure}
\section{Defenses against MI Attacks}

\label{sec:defense}

A rich quantity of defense strategies have been proposed to alleviate MI threats and achieved impressive resistance effects. 
Basically, these methods focus on improving MI robustness by eliminating the valid information stemming from the pre-trained model that an MI adversary might maliciously utilize. As presented in Figure \ref{fig:defense_taxonomy}, we primarily divide them into two categories: \textit{Model Output Processing} and \textit{Robust Model Learning}, as shown in Figure \ref{fig:defense objective}. Next, we illustrate these methods in accordance with the proposed taxonomy.
\subsection{Model Output Processing} 

Since MI attacks fundamentally exploit the redundant information contained in the victim model's output \cite{wang2021improving}, a viable defense mechanism is to reduce this redundancy by obscuring the model output with various techniques.

% -----------------------------------
%  加噪声
% -----------------------------------
\textbf{Additive noise.}
Intuitively, \cite{wen2021defending} proposes perturbing the confidence scores with well-designed noise crafted through adversarial learning with an inversion model.
The inversion model serves as an MI adversary, which aims to recover the input data from the confidence vector. Meanwhile, the adversarial noise is updated to resist the inversion model by maximizing the MSE between the original images and images reconstructed from the inversion model. To maintain the main task accuracy, they impose a constraint on the noise that assures the predicted label unchanged.
Considering the Differential Privacy (DP) \cite{dwork2006differential, dwork2008differential, abadi2016deep, ji2014differential, ha2019differential} guarantee, \cite{ye2022one} designs an approach that divides the score vector into multiple sub-ranges and applies an exponential mechanism to replace and normalize the values in each sub-range. This mechanism assures the target model's MI robustness with rigorous theoretical proof of differential privacy. 

\cite{titcombe2021practical} further examines the scenario of split learning, where a global model is split into several components and distributed to disparate clients. During the training, each client calculates the features output from its partial model and then uploads them to the central server for aggregation. Considering the uploaded features also face the threat of being converted into private images by an attack model, \cite{titcombe2021practical} suggests a simple yet efficient resolution that incorporates Laplacian noise into the intermediate representations before transmitting them to the computing server, thereby obscuring the potential MI attacker.

To protect graph data, \cite{zhang2022inference} also applies Laplacian noise to the model output $\mathcal{H}$. They release a noisy but usable version of output embeddings $\widetilde{\mathcal{H}} = \mathcal{H} + \Delta$, where $\Delta$ denotes the Laplacian noise. \cite{wu2022LINKTELLER} also develops a differentially private GNN model and adds noise to the adjacency matrix to guarantee DP. \cite{olatunji2023private} perturbs feature explanations using a randomized response mechanism and reduces the attack to a random guess. Since applying noise may result in performance degradation, it is essential to choose a decent level of noise and achieve a trade-off between utility and privacy.

 \renewcommand{\dblfloatpagefraction}{.9}
% MODIFIED BY YWB
\begin{table*}[htbp]
  \setlength{\tabcolsep}{1.3pt}
  \normalsize
  \centering
  \begin{threeparttable} 
    \resizebox{\linewidth}{!}{
    % \begin{tabular}{cccccccp{14.785em}} \toprule
    \begin{tabular}{lcccc} \toprule
    Defense &           % name
    Publication &              % Output Processing / Robust Learning
    Process Type &             % Training / Inference / Architecture
    Core Feature &       % 其它特色
    % Empty           % Black-box / White-box
    \\ \midrule
    SplitNoPeek \cite{titcombe2021practical} & ICLR Workshop'21  & Additive noise & Apply Laplacian noise  \\
    Adversarial \cite{wen2021defending} & IEEE CSR'21 &  Additive noise & Adversarial sample optimization \\
    DP \cite{ye2022one} & IEEE TIFS'22 & Additive noise & Differential privacy technique \\
    Purification \cite{yang2020defending} & arXiv'20 & Purification & Autoencoder-based purifier \\
    Purifier \cite{yang2023purifier} & AAAI'23 & Purification & Conditional VAE-based purifier \\
    Crafter \cite{wang2024crafter} & NDSS'24 & Purification & Optimize to near public samples \\
    SSD \cite{SSD} & ICLR'25 & Purification & Reduce conditional mutual information \\
     % Robust Learning
    MID \cite{wang2021improving} & AAAI'21 & Secure training mechanism & Minimize mutual information \\
    BiDO \cite{peng2022bilateral} &  KDD'22 &  Secure training mechanism  & Bilateral dependency optimization \\
    ResSFL \cite{li2022ressfl} & CVPR'22 &  Secure training mechanisms  & Adversarial training with attacks \\
    LS \cite{szegedy2016rethinking} & ICLR'24 &  Secure training mechanism  & Negative label smoothing \\
    BiDO+ \cite{bidoplus} & IEEE TPAMI'25 &  Secure training mechanism  & Outlier exposure regularizer \\
    AT-AT \cite{ATAT} & ICLR'26 & Secure training mechanism & Adversarially non-robust feature \\
    GAN-based \cite{gong2023gan} & IEEE TIFS'23 & Robust Fine-tuning  & Adversarial training with attacks \\
    TL \cite{TL} & CVPR'24 & Robust Fine-tuning  & Freeze first few layers  \\
    RoLSS \cite{koh2024vulnerability} & ECCV'24 & Robust model architecture  & Remove skip connections \\
    LoFt \cite{calor}  &  KDD'26 & Robust model architecture  & Low-rank feature filtering \\
    Trap-MID \cite{trapmid} &   NeurIPS'24 & Input data preprocessing  & Image trigger injection \\
    \bottomrule
    \end{tabular}}
    \end{threeparttable} %添加此处
    % \vspace{-0.5em}
    \caption{Summary of existing MI defenses for visual tasks.}
    
    % \vspace{-1em}
  \label{table:defense}%
\end{table*}%

% -----------------------------------
%  纯化
% -----------------------------------
\textbf{Purification.}
This type of defense aims to purify the confidence scores by reducing the redundant information associated with privacy leakage, which contributes to significantly increasing the difficulty of MI attacks.

In light that a major mechanism of MI attacks is that the outputs are distinguishable for diverse input images, \cite{yang2020defending} seeks to defend by rendering model predictions less distinguishable and less sensitive to input data changes. 
Concretely, they propose an autoencoder-based purifier ${P}(\cdot)$ that encourages different input images to be mapped into the same or similar output by reducing the dispersion of the confidence scores from target classifier $f_{\theta}(\cdot)$.   
% and the purified confidences $P(f_{\theta}(\mathbf{x}))$ are returned to the querying user as final model outputs.  the pioneering research 
% This approach encourages different input images to be mapped into the same or similar model output. 
%
To enhance the purifier ${P}(\cdot)$, they employ an adversarial learning technique, which involves jointly learning an adversarial inversion model $\mathcal{M}(\cdot)$ via minimizing the inversion loss:
\begin{equation}
\label{eq:L_inv}
\begin{aligned}
\mathcal{L}_{inv} = ||\mathcal{M}({P}(f_{\theta}(\mathbf{x}))) - \mathbf{x}||_2^2, 
\end{aligned}
\end{equation}
where $\mathbf{x}$ denotes the input images. To train the purifier $\mathcal{P}(\cdot)$, they additionally introduce the confidence reconstruction loss $\mathcal{L}_{rec}$ and the classification loss $\mathcal{L}_{iden}$ as:  
\begin{equation} 
\label{eq:purifier}
\begin{aligned}
    \mathcal{L}_{rec} &= ||{P}(f_{\theta}(\mathbf{x})) - f_\theta(\mathbf{x})||_2^2, \\
    \mathcal{L}_{iden} &= \mathbb{CE}(P(f_{\theta}(\mathbf{x})), 
\mathop{\arg\max} f_\theta(\mathbf{x})).
\end{aligned}
\end{equation}
Finally, the purifier's optimization objective can be formulated as $\mathcal{L}_{total}= \mathcal{L}_{rec}+\lambda_1\mathcal{L}_{iden} - \lambda_2\mathcal{L}_{inv}$, where $\lambda_1$ and $\lambda_2$ are hyperparameters.
% \todo{make it clear}
\cite{yang2023purifier} follows this pipeline and substitutes the autoencoder with a variational autoencoder with the predicted label as conditioning input, which enhances the output indistinguishability and thus ensures the model's MI resiliency. A novel defense framework is to intentionally mislead the attackers into reconstructing public samples \cite{wang2024crafter}.
They consider the edge computing scenario with similar settings to the split learning \cite{titcombe2021practical}, where a user uploads the intermediate features $Enc(x)$ encoded from its private image $x$ to the central service provider. 
To protect the $Enc(x)$ from the threat of an MI attacker, the edge device carefully optimizes these vulnerable features to be closer to the features of the public sample, hence misleading the attacker into recovering insensitive public images. Besides, an Euclidean distance is adopted to restrict the deviation from the original features to further preserve the utility of perturbed features. 
% Moreover, the study proposes a label swapper to achieve label indistinguishability. 
% For each class, the output label is randomly replaced with the label having the second highest confidence score (computed on the private dataset) to further confuse potential attackers. \todo{check}
% The defender randomly selects a subset of the private training set and obtains each confidence vector outputted by the CVAE and the corresponding labels of the training data in the subset. Then, each label is randomly replaced with the label having the second highest confidence score. During inference, the defender uses a $k$-nearest neighbor (KNN) strategy to obtain the nearest confidence score to vote for the final output.

SSD \cite{SSD} proposes a novel defense strategy specifically designed for black-box scenarios. They propose to postprocess output logits to minimize the conditional mutual information between inputs and outputs. This objective is formulated as a convex optimization problem and solved via a water-filling algorithm.

\cite{he2021stealing} considers guarding graph data by restricting the GNN model to output only the $k$ largest posterior probabilities, which purifies the confidences to diminish unnecessary information and thus decreases the potential risk under MI threats.

\textbf{Detection and Filtering.} A distinctive attribute of language models is that the model's output contents possess meaningful semantics and can be further analyzed for security concerns. Therefore, \cite{huang2022large} proposes an effective defense approach that detects and filters the privacy-sensitive information within the prediction. Specifically, this method includes a detection module to examine whether the output text contains sensitive information and a filtering module to take appropriate measures based on the detection module's action like refusing to answer.  

Overall, the main challenge of output processing defenses is to reduce privacy leakage without severely harming model utility; moreover, since such methods operate by perturbing or purifying returned outputs, they are mainly applicable to black-box API settings and are generally unsuitable for open-source or offline-deployed models where attackers can directly access model parameters and internal information.

% --------------------------------------------------------
% \textbf{Improvement of Training Model}
% Modified BY YHY
% \textbf{Robust Model Training}
% --------------------------------------------------------
% BY CBL                 minor modification by YHY: 数学公式后加标点, MIR -> MID
% minor modified by QYX: 期望用空心符号,损失函数用\mathcal
% \textbf{Robust Model Training.} MID \cite{wang2021improving} emphasizes that MI attacks leverage the correlation between the input and output of the target model, thereby proposing to penalize the mutual information between input $X$ and output $\hat{Y}$ during the training. 
% The training loss can be expressed as:
% \begin{equation}
% \min_{\theta} \mathbb{E}_{(x,y) \sim p_{X,Y}(x,y)} \left[ \mathcal{L}(f_{\theta}(x), y) \right] + \lambda \mathcal{I}(X, \hat{Y}),
% \end{equation}
% where $f_{\theta}(\cdot)$ is the target model, $\mathcal{L}_{\theta}(f(x), y)$ denotes the loss for the main task,  \(\mathcal{I}(X, \hat{Y})\) is the approximated mutual information, and $\lambda$ is a coefficient balancing privacy and utility.
% 
\subsection{Robust Model Learning} 

% Instead of post-processing the output, some defense strategies are proposed to enhance the robustness of the victim model itself. 
% In addition to post-processing the model outputs, several methods have been proposed to enhance the robustness of the victim model itself. These methods involve introducing regularization terms and misleading inversion process.
Despite the considerable success of these model output processing methods, they are constrained to black-box scenarios, where an attacker is merely capable of making queries and obtaining the corresponding output. In contrast, in white-box scenarios, attackers gain complete access to the actual model output before the post-processing. To address this challenge, researchers seek to enhance the model's inherent robustness.

% \textbf{Introducing regularization terms}

\textbf{Secure Training Mechanism.}
This series of algorithms modifies the model training process to boost the MI robustness.
% -----------------------------------------------------
%                     MID
% -----------------------------------------------------
The pioneering research MID \cite{wang2021improving} is a classic approach to exploring robust training and analyzes that the inversion attack greatly depends on the correlation between the input $X$ and output $\hat{Y}$ of the target model $f_{\theta}$, thereby proposing to penalize the mutual information between $X$ and $Y$ during model training. 
The training objective can be expressed as:
\begin{equation}
\min_{\theta} \mathbb{E}_{({x},y) \sim p_{X,Y}({x},y)} \mathcal{L}(f_{\theta}({x}), y) + \lambda \mathcal{I}(X, \hat{Y}),
\end{equation}
where $y$ is the ground truth label, $\mathcal{L}(\cdot, \cdot)$ denotes the loss for the main task, \(\mathcal{I}(\cdot, \cdot)\) denotes the mutual information, and $\lambda$ is a coefficient balancing privacy and utility. However, it is impracticable to directly compute the $\mathcal{I}(X, \hat{Y})$  due to the difficulty of modeling the joint distribution of model input and output. Inspired by the work on information bottleneck, they substitute it with its upper bound $\mathcal{I}(X, Z)$ \cite{kolchinsky2019nonlinear}, where $Z$ represents a stochastic encoding of the input $X$ at some intermediate layers. Then a variational method is employed to approximate the $\mathcal{I}(X, \hat{Y})$ term.

% -----------------------------------------------------
%                     BiDO
% -----------------------------------------------------
Nonetheless, MID meets the dilemma between data privacy and model utility as the regularizer conflicts with the main task loss. \cite{peng2022bilateral} proposes a bilateral dependency optimization (BiDO) to solve this problem. Instead of directly diminishing the correlation between inputs and outputs, BiDO minimizes the dependency $d(X, Z)$ between inputs $X$ and latent representations $Z$ while maximizing the dependency $d(Z, \hat{Y})$ between latent representations $Z$ and outputs $\hat{Y}$. The first term limits the propagation of redundant information from inputs to the latent representations, preventing misuse by the adversary. The latter term facilitates the latent layers to learn discriminative representations, ensuring model utility for classification. The overall training loss can be formulated as:
\begin{equation}
\label{math:bido}
\begin{split}
    \mathcal{L}_{BiDO} = \mathcal{L}_{CE}(\hat{Y}, Y) + \lambda_x \sum_{i=1}^m d(X, Z_i) \\
    - \lambda_y \sum_{i=1}^m d(Z_i, \hat Y),
\end{split}
\end{equation}
where $\lambda_x$ and $\lambda_y$ denote the hyperparameters controlling the intensity of loss items, and $Z_i, i=1, \dots, m$ denotes the $i$-th layer's latent representations specific to the input $X$. Additionally, the study evaluates both Constrained Covariance \cite{coco} and the Hilbert-Schmidt Independence Criterion (HSIC) \cite{hsic} as distance metrics to quantify dependence and finds that HSIC yields better defense performance. 
BiDO+ \cite{bidoplus} further augments BiDO with an outlier exposure (OE) regularizer to mitigate its insufficient OOD awareness, which further enhances the defense effectiveness. 
% The OE loss is formulated as
% \begin{equation}
% \mathcal{L}_{OE}=\mathbb{E}_{x\sim D_{in}^{train},x'\sim D_{out}^{aux}}
% \left[\ell_{OE}\big(F(x;\theta),F(x';\theta)\big)\right].
% \end{equation}
% which explicitly leverages auxiliary OOD samples during training. It  enlarges the response discrepancy between ID and OOD data.
% By encouraging more separable responses between ID and auxiliary OOD samples, this term suppresses overconfident predictions on unknown inputs and improves OOD detection. At the same time, it further enhances robustness against model inversion attacks by making non-training-like samples harder to exploit for privacy reconstruction.

% \cite{gong2023gan} explore an alternative method to mislead the attackers. They incorporate GAN-based fake samples into the training of the victim model. They fine-tune the victim model by minimize the loss of public samples and maximize that of private ones.
% -----------------------------------------------------
%                     LS
% -----------------------------------------------------
Label smoothing \cite{szegedy2016rethinking}, initially proposed to improve model generalization or calibration, is investigated by \cite{labelsmoothing} to defend model inversion attacks.
Given the initial one-hot label $\mathbf{y}$, the class number $C$, and the smoothing factor $\alpha$, the smoothed label is defined as:
\begin{equation}
\mathbf{y}^{LS}=(1-\alpha)\cdot\mathbf{y} + \frac{\alpha}{C}.
\end{equation}
Correspondingly, the training loss incorporating the label smoothing technique is presented as:
% where alpha represents the smoothing coefficient. 

\begin{equation}
\label{math:labelsmoothing}
    \mathcal{L}_{LS} = (1-\alpha) \mathcal{L}_{CE}(\hat{\mathbf{y}}, \mathbf{y}) + 
    \frac{\alpha}{C}\sum_{k=1}^C \mathcal{L}_{CE}(\hat{\mathbf{y}}, \mathbbm{1}),
\end{equation}
where $\hat{\mathbf{y}}$ and $\mathbf{y}$ represent the predicted confidences and the ground truth, and $\mathbbm{1}$ denotes an all one vector with $C$ entries. LS conducts sufficient experiments to analyze the influence of the smoothing factor $\alpha$ and reveals that larger values of $\alpha$ lead to greater vulnerability to MI attacks. In contrast to the standard training which adopts $\alpha$ greater than or equal to zero, LS encourages negative values of $\alpha$ to reduce the privacy threat.
% Using an increasing batch size schedule, it scales up the batch size to millions to improve the utility of the DP-SGD step for BERT. 

The latest research AT-AT \cite{ATAT} challenges the prevailing view in the field of MI defenses, which holds that HSIC between private inputs and intermediate features renders models more vulnerable to training data privacy leakage. This work posits that adversarially non-robust features (\textit{i.e.}, subtle discriminative signals imperceptible to the human eye that can flip classification predictions with minute pixel perturbations) constitute the root cause of privacy leakage in deep learning models. Existing defense methods achieve privacy protection only through unintentionally driving models to rely on non-robust features for classification, at the cost of a drastic degradation in model accuracy when exposed to minor adversarial perturbations. Accordingly, AT-AT proposes to deliberately suppress the model’s reliance on robust features during training. Specifically, it appends an anti-adversarial loss as a regularization term on top of the standard clean-image classification loss, enabling a more favorable trade-off between classification accuracy and privacy preservation. The training objective can be expressed as:
\begin{equation}
\mathcal{L}_{AT-AT} = \mathcal{L}(f_{\theta}({x}), y) + \lambda \mathcal{L}(f_{\theta}({x+\delta}), y^*),
\end{equation}
where $y$ is the ground truth label, $\mathcal{L}(\cdot, \cdot)$ denotes the main loss, $\delta$ is a targeted adversarial perturbation for a random wrong label $y^*$.

% -----------------------------------------------------
%                 Ressfl
% -----------------------------------------------------
 ResSFL \cite{li2022ressfl} considers the split federated learning (SFL), where a central server might misuse the clients' uploaded intermediate features to reconstruct private data. A novel two-step framework is developed to deal with the MI threat.
 The first step adopts a publicly available dataset to pre-train a model with both high accuracy and strong MI resiliency. Similar to \cite{yang2020defending}, this stage jointly trains a powerful inversion model as an adversary to achieve the MI robustness of the pre-trained model. Besides, ResSFL analyzes that higher dimensions of uploaded features indicate better MI vulnerability. Hence, they modify the client model structure to reduce the dimension of output features.
 In the second step, the pre-trained model serves as the initial model for subsequent split learning. In addition to the regular main task loss, a weaker inversion model is incorporated to calculate the inversion loss that maximizes the distance between the private images and the images reconstructed by the inversion model. This allows for the optimization of the client-side model to achieve high accuracy while preserving MI resistance.

For language models, \cite{anil2022large} reveals that language models trained with variants of the differential privacy-based optimizer (i.e, scaling up the batch size to millions to improve the utility of the DP-SGD step) exhibit more robust MI resistance while encountering the expense of performance degradation and increased computation. Moreover, \cite{ishihara2023training} demonstrates the \textit{memorization} of language models essentially has a positive connection to the model overfitting. Therefore, adequate regularization and early stopping are also served as reliable alternative defense strategies.

In graph learning, \cite{zhang2021graphmi} considers differential privacy (DP) into the model training process by adding Gaussian noise to the clipped gradients in each training iteration for enhanced MI robustness. 
Similar to \cite{wang2021improving, peng2022bilateral}, \cite{zhou2023on} proposes to diminish the correlation between the GNN's inputs and outputs by injecting stochasticity into adjacent matrix $\mathcal{A}$ and regularizing the mutual dependency among graph representations, adjacency, and predicted labels during the model training.

\textbf{Robust Fine-tuning.}
As noted in \cite{TL}, the above approaches to directly modify model training inevitably result in either a degradation in model utility or great efforts in hyperparameter tuning, given the model's sensitivity to minor alterations in hyperparameters. To solve this challenge, another research thread proposes to fine-tune the trained model instead of directly operating the model training phase.

% -----------------------------------------------------
%                     GAN-based misleading
% -----------------------------------------------------
A GAN-based method \cite{gong2023gan} discusses the potential of fine-tuning a trained model to intentionally fool an attacker into inverting insensitive public samples.
Apart from the classifier $f_{\theta}(\cdot)$, the defender trains an additional public classifier $f_{p}(\cdot)$ and a GAN model $G_p(\cdot)$ using a publicly available dataset. 
Then, the defender uses the obtained GAN $G_p(\cdot)$ as image priors and conducts MI attacks on both the victim model $f_{\theta}(\cdot)$ and the public $f_{p}(\cdot)$ to reconstruct private and public images respectively.
Afterwards, the defender fine-tunes the victim model with the generated samples, i.e., minimizing the classification loss of insensitive public samples while maximizing that of private ones. This operation injects misleading information into the victim model’s predictions, hence deceiving the attacker into inverting samples that are significantly different from the private ones that should be protected.

% -----------------------------------------------------
%    Confidence Adaptaion
% -----------------------------------------------------
% \cite{calor} proposes to introduce a bias between the attacker's optimization goal and the actual attack objective. They fine-tune the pre-trained model with a confidence adaptation loss
% \begin{equation}
%     \mathcal{L}_{CA} = \lambda_1 \hat{y}_c^{\lambda_2} \log \hat{y}_c,
% \end{equation}
% where $\lambda_1$ and $\lambda_2$ are hyperparameters, and $\hat{y}_c$ is the predicted confidence on the label $c$. 
% % 
% With full optimization, the predicted confidence converges to $\exp{-1/\lambda_2} < 1$, while the target optimization goal remains at $1$. Consequently, the more the attacker optimizes, the greater the deviation between the reconstructed image and the original private image becomes.

% 
% -----------------------------------------------------
%                     TL
% -----------------------------------------------------
Subsequently, TL \cite{TL} presents a straightforward and feasible strategy based on the transfer learning technique \cite{pan2010survey}. 
 % limit the number of layers encoding sensitive information from private data.
The Fisher information is first adopted to measure the influence of each intermediate layer on the ultimate MI attack performance and the clean classification accuracy. These findings indicate that the initial layers of a model exhibit a notable correlation with MI attacks, whereas the last several layers contribute more to the main task accuracy.
Therefore, TL introduces a transfer learning mechanism since it typically fine-tunes the last few layers of a target model that are not relevant to MI attacks. Specifically, the algorithm involves pre-training the victim model on public datasets and then fine-tuning the last few layers on a private dataset, yielding excellent defense effects. 

\textbf{Robust Model Architecture.}
This type of defense strategy concentrates on designing a more robust model architecture to resist MI attacks. RoLSS \cite{koh2024vulnerability} proposes pioneering work to analyze how DNN architectures affect MI attacks. They first demonstrate that skip connections facilitate MI attacks due to their positive effect on the gradient flow during the inversion process. Nevertheless, they also reveal that simply adopting a previous study, RepVGG \cite{ding2021repvgg}, to convert a training-time multi-branch block (with skip connection) into an inference-time plain convolutional layer (without skip connection) performs poorly in resisting the MI attacks despite the removal of skip connections in the inference-time model architecture. 
Based on these findings and empirical study, they design novel model architecture techniques to improve MI robustness, including the removal of Last Stage Skip-Connection (RoLSS), Skip-Connection Scaling Factor (SSF), and Two-stage Training Scheme (TTS), respectively. Concretely, SSF introduces a scale factor $k\in[0,1]$ to the skip connections of the last stage, controlling the range of gradients as follows:
\begin{equation}
\label{math:ssf}
    z_{i+1}=block_{i}(z_{i})+k \cdot z_{i},
\end{equation}
where $z_{i}$ and $z_{i+1}$ represent the input and output, $block_{i}$ indicates the intermediate block of the classifier (e.g., ResNet block \cite{he2016deep}). In particular, the SSF method degrades to the RoLSS method when $k=0$. TTS method first trains the model with full skip connections to obtain reasonable convergence of parameters and then deploys RoLSS to the training process to improve robustness.  

LoFt \cite{calor} proposes a structural modification to the model by implementing a low-rank feature filtering strategy in the linear classification header. This approach compresses the input feature of the classification header into a lower-dimensional space, thereby reducing the information contained. After compression, the features are passed through a nonlinear activation function $Tanh$, which can introduce gradient vanishing issues. The final classification then relies solely on this compressed feature for decision-making. In summary, the low-rank compression strategy limits the amount of output information, and the gradient vanishing issue further hinders the attackers' optimization process, making it significantly harder to reconstruct private samples.

\textbf{Input Data Preprocessing.}
These methods elaborately modify training data to enhance robust model learning.
Trap-MID \cite{trapmid} defends against model inversion attacks by deliberately misleading the inversion process. Specifically, it injects a trigger into private training images, such that the reconstructed images are biased toward the trigger pattern instead of the genuine private features, thereby reducing privacy leakage.

% Text:
% Several techniques are used to enhance the MI robustness of the language models
% Since MI attacks seek to invert language models for extracting private information, an intuitive defense strategy is to purify the training data by removing the text containing personal information \cite{ishihara2023training}.
 \cite{kandpal2022deduplicating} demonstrates that the data duplication strengthens the \textit{memorization} of the victim language model, thereby increasing its risk of MI attacks. Therefore, training data de-duplication is an efficient approach to reduce the privacy breaches brought by MI attacks. 
% GNN:
In graph learning, \cite{zhang2023model} proves that ensuring DP cannot effectively prevent MI attacks and proposes an input processing technique, which conceals real edges by pre-processing the training graph, such as randomly flipping and rewriting. This operation tricks the attacker into reconstructing dummy data that significantly differs from the ground truth.  

The main challenge of robust model learning lies in high deployment cost, since they typically require fine-tuning or retraining the target model with additional objectives, regularizers, or architectural modifications. Meanwhile, these interventions often interfere with the original optimization target, making them prone to substantial utility degradation and unstable performance in practice.

\section{Social Impact}
\label{sec:social}
Given their effectiveness and practicality, MI attacks have become a significant privacy threat within the AI community. Case studies \cite{carlini2023extracting, nasr2023scalable} on production models (e.g., ChatGPT and Stable Diffusion) have intensified concerns about personal privacy and the urgent necessity for AI governance. Besides, the privacy risk is greatly amplified by the explosive development of AI generative content (AIGC), where numerous large models are trained on massive datasets that lack ownership verification.
Currently, individuals may not be aware of which models have been trained on their personal data and cannot confirm whether model owners have adequately protected their privacy, which could be considered a continuous infringement of their rights. 
Therefore, model-level privacy protection and its interplay with official lawmaking are becoming a heated topic.

The study by \cite{veale2018algorithms} presents an interesting viewpoint: if a model is vulnerable to MI attacks, the model itself should also be regarded as a form of personal data, and correspondingly trigger a series of rights and obligations as defined by existing data protection laws (e.g., the EU General Data Protection Regulation). Nevertheless, it is a considerably complicated task to quantify a model's vulnerability to MI attacks, and directly categorizing models as personal data could disrupt the power relationship between model holders and individual data owners, which could have a detrimental impact on technical development.
A more cautious approach for data users is to adopt robust and reliable defense strategies or establish access permissions to build responsible AI systems, thereby alleviating potential contradictions with individual data subjects. Besides, enacting new privacy legislation specific to model-level protection is indispensable, and introducing more regulatory policies is also strongly encouraged.

\section{Conclusion and Research Directions}
\label{sec:fur_con}

This survey presents an exhaustive review of the powerful MI attacks and defenses. First, we comprehensively categorize existing methods into several macroscopic taxonomies and present a systematic overview. We then briefly introduce early MI methods on traditional ML scenarios. Next, we characterize the mainstream attacks on DNNs from multiple perspectives, based on which we deeply illustrate their features and distinctions. Moreover, we investigate MI studies on more data modalities and provide detailed taxonomies. 
To facilitate the development of this field, we present a summary of several unsolved challenges and suggest promising directions for future research.

% \textbf{Automatic Malicious Prompts for LLMs.}
% % -------------------------------------------------------
% %  BY QYX
% %  MODIFIED BY YWB: 补充自动生成prompt的含义，并强调原有方法的handcrafted特性。此处应当写出与NLP领域的MI攻击的区别，即强调LLM恶意prompt生成的自动性，而不仅是点出LLM恶意prompt
% Previous methods \cite{wang2023decodingtrust} tend to manually generate handcrafted malicious prompts for LLMs. To further improve the attack performance, we suggest applying the paradigm of GAN-based MI attacks. Specifically, the attackers first train a prompt generator to efficiently obtain high-quality prompts by leveraging an open-source LLM, such as GPT-2 []. Then, they can transfer the pre-trained generator to black-box LLMs and automatically produce malicious prompts to induce leakage of sensitive privacy, which is more efficient and less time-consuming than manually designing prompts.
% -------------------------------------------------------

\textbf{Stronger Prior for MI attacks.}
In visual tasks, previous approaches leveraged various GANs as image priors for reconstruction guidance. A notable contribution by \cite{liu2023diffusion} introduced diffusion models into MI attacks. They manually trained a conditional diffusion model \cite{ho2022classifier} to generate $64\times 64$ resolution images from the target distribution. To conquer the harder scenario of higher-resolution image recovery, future research could explore pre-trained diffusion models with rich image priors or alternative techniques to enhance generative model utilization. 
% This paradigm extension is also applicable to other modalities.

\textbf{Defense From a Data Perspective.} 
Existing defenses for image data primarily operate on the model \cite{wang2021improving, peng2022bilateral} or its outputs \cite{wang2021improving}, while little attention is paid to the training data. Therefore, one potential defense strategy is specially processing the data by applying well-designed masks or perturbations to create confusing reconstruction results \cite{tan2024defending}. 
Additionally, data synthesis \cite{lei2023comprehensive} techniques have recently garnered significant attention. Since the synthesized images usually disguise the sensitive information, employing them in the training process emerges as a promising defensive approach against MI attacks.

\textbf{Certified Robustness for MI Defenses.} 
Despite existing studies on MI robustness, current defense approaches lack certified robustness guarantees. Certifiably robust approaches for DNNs have been developed against adversarial attacks \cite{adversarial2019certified} and backdoor attacks \cite{backdoorcertifying}. It is also imperative to establish a robust theoretical framework that can provide certified guarantees in the context of MI attacks.  Additionally, exploring ways to integrate user feedback and preferences into the certification defense pipeline can enhance the practicality of ensuring certified robustness in conversational AI models like ChatGPT.

%----------------------------------------------------
% BY QYX
% Target classifiers with different structures possess different predictive power. 
%\cite{zhang2020secret} conduct experiments on image data to explore the connection between the model structures and the vulnerability to MI attacks. They measure the attack performance on DNNs with different structures and investigate the effects of certain intermediate layers of the target model, such as the dropout and batch normalization layer. Their experimental results indicate that the attack performance has a significant correlation to the design of model structures. 
%However, this study is limited to several simple models and there is a lack of analysis about how DNN architecture designs affect model inversion attacks. 
% and reveal that a model is trained to obtain higher , which inspires us there is a likely correlation between the structure of the target model and its robustness to MI attacks. 
%Since seeking a robust model architecture facilitates defenses against MI and helps researchers to better understand the underlying attack mechanism, conducting in-depth studies on how to design robust model architectures is a promising direction for future research.

% --------------------------------------------------------
% BY YHY
% Recently, newly emerged multimodal models have gained increasing attention due to their complex data handling capabilities.
\textbf{Multi-Modal MI Attacks and Defenses.}
Recently, advanced multi-modal models trained on cross-modal data pairs have shown sophisticated capabilities in handling complex and diverse data. 
% 
% Despite these developments, there is a lack of research on MI attacks targeting multimodal models. 
% Consequently, in-depth investigations into how these multimodal models process and integrate various modalities, and explorations of potential vulnerabilities during the multimodal fusion process, represents a promising and necessary direction for model inversion attack and defense.
While current MI methods have made great progress on unimodal models, there is a lack of research on multi-modal foundation models, e.g., the CLIP \cite{radford2021learning} and SAM \cite{sam}. Exploring potential vulnerabilities during the cross-modal inference process can be a promising and necessary direction for model inversion attacks and defenses.

\textbf{Model Inversion for Good.}
Given the ability to recover diverse data from the target model's training set distribution, MI attacks can naturally be applied to various data-free learning scenarios \cite{yin2020dreaming, fang2021contrastive}. 
For instance, data reconstructed from a white-box teacher model can serve as a reliable dataset for training student models in data-free knowledge transfer and continual learning scenarios. Since advanced MI methodologies can generate data of higher quality, they are anticipated to enhance the efficacy and versatility of their application. In addition, future research could consider utilizing the synthesized data for a broader range of learning tasks and further improve their application performance.
 % The majority of DD techniques lie in the single-label classification task where its label space is an integer set. As for other challenging tasks such as detection [97] or segmentation [98], their label spaces locate in much higher dimensional spaces. Unlike DD with singlelabel classification the specific number of synthetic images (IPC) are pre-assigned for each class, enumerated alignment is impractical when the label space is highly-dimensional. Therefore, to represent the rich label space high-dimensional labels of synthetic data require cautious adjusting, which significantly hinders the DD efficiency especially when jointly optimized with the input data.

 % Multimodal dataset distillation. Whereas DD algorithms have achieved remarkable progress in terms of image[11, 24], graph [66], text [68], tabular [94], and recommen-dation system data [95], their applications on other datamodalities are still underexplored.Especially for the modali-ties such as video that rich spatial and temporal informationare tangled with each other in a single data [96], it is stillchallenging for DD algorithms to simultaneously distill thespatial and temporal features into a small volume of data.
% --------------------------------------------------------

\section*{Declarations}

\textbf{Conflict of Interest.} The authors declare that they have no conflict of interest. All authors confirm that there are no financial or non-financial interests, relationships, or affiliations (including but not limited to employment, consultancies, stock ownership, honoraria, grants, or other forms of compensation) with any organization or entity that could inappropriately influence or bias the work reported in this paper.

\textbf{Funding Declaration.} This work is supported in part by the National Natural Science Foundation of China under grant 62301189, 62576122, 62571298, and the Guangdong Basic and Applied Basic Research Foundation under grant 2026A1515011139.

\textbf{Data Availability.} The data supporting the visual comparison in Fig. \ref{fig:cv_reconstruct} is from an openly available dataset FaceScrub \cite{facescrub}, which can be obtained from its \href{https://vintage.winklerbros.net/facescrub.html}{official website}. Moreover, our open-source repository includes the code to generate the images shown in Fig. \ref{fig:cv_reconstruct}.

\section*{Compliance with Ethical Standards}

\textbf{Ethical Approval.} Not applicable. This study did not involve any experiments on human or animal participants, nor did it collect primary data requiring ethical review. Therefore, no ethical approval was required for this research. All data involved in this article were obtained from publicly available sources and were anonymized, ensuring no breach of privacy or confidentiality.

\textbf{Consent to Participate statement.} Not applicable. This review article is based exclusively on publicly available datasets and previously published literature. No human participants were involved in the conduct of this study, and therefore informed consent was not required.



\end{document}